\documentclass[11pt]{article}

\usepackage{acl}

\usepackage{times}
\usepackage{latexsym}
\usepackage[T1]{fontenc}
\usepackage[utf8]{inputenc}
\usepackage{microtype}
\usepackage{inconsolata}
\usepackage{graphicx}

\usepackage{amsmath}
\usepackage{amssymb}
\usepackage{bbm}
\usepackage{subcaption}
\usepackage{float}
\usepackage{booktabs}
\usepackage{tabularx}
\usepackage{array}
\usepackage{xcolor}
\usepackage{multirow}
\usepackage{colortbl}
\usepackage{algorithm}
\usepackage{algpseudocode}
\usepackage{enumitem}
\usepackage{tikz}
\usetikzlibrary{shapes,arrows,positioning}
\usepackage{pifont}

\usepackage{tcolorbox}
\tcbuselibrary{skins,breakable}
\usepackage{mdframed}
\usepackage{multicol}
\usepackage{soul}
\usepackage{wrapfig}
\usepackage{placeins}   % provides \FloatBarrier

\newcommand{\modeltier}[1]{%
  \vspace{0.8em}%
  \noindent{\large\textbf{\textbullet\ #1}}\par\nobreak\vspace{0.3em}}

\definecolor{lightblue}{RGB}{230,240,255}
\definecolor{lightgreen}{RGB}{230,255,230}
\definecolor{lightyellow}{RGB}{255,255,230}

\definecolor{red}{RGB}{220,40,30}
\definecolor{BrightOrange}{RGB}{255,140,0}
\definecolor{Gold}{RGB}{255,200,0}
\definecolor{Emerald}{RGB}{0,160,80}
\definecolor{Cobalt}{RGB}{0,120,215}
\definecolor{Sapphire}{RGB}{90,60,170}
\definecolor{Amethyst}{RGB}{160,80,200}
\definecolor{framegray}{RGB}{170,170,170}

\definecolor{csProblemBg}{RGB}{255,250,240}
\definecolor{csGuidedBg}{RGB}{237,244,255}
\definecolor{csAvalancheBg}{RGB}{255,243,240}
\definecolor{csRecoverBg}{RGB}{237,252,245}
\definecolor{csVanillaBg}{RGB}{235,240,255}
\definecolor{csExpBg}{RGB}{235,250,235}
\definecolor{csVanillaBlue}{RGB}{40,80,180}
\definecolor{csExpGreen}{RGB}{30,130,60}
\definecolor{csTurnPass}{RGB}{234,248,234}
\definecolor{csTurnFail}{RGB}{255,232,228}
\definecolor{csTurnNeutral}{RGB}{250,250,252}

\usepackage{listings}
\lstdefinestyle{cscode}{%
  basicstyle=\linespread{0.9}\fontsize{6}{7.5}\ttfamily,
  breaklines=true, breakatwhitespace=false,
  keepspaces=true, showspaces=false, showstringspaces=false,
  tabsize=2, frame=none, backgroundcolor=\color{gray!8},
  xleftmargin=2pt, xrightmargin=2pt,
}

\newcommand{\graphturn}{\textsc{CypherTurn}}
\newcommand{\cmark}{\ding{51}}
\newcommand{\xmark}{\ding{55}}
\newcommand{\cmarkc}{\textcolor{Emerald}{\ding{51}}}
\newcommand{\xmarkc}{\textcolor{red}{\ding{55}}}

\usetikzlibrary{shapes.geometric,shapes.callouts,arrows.meta,calc,backgrounds,fit}

\title{\graphturn: A Multi-Turn Benchmark for Conversational Text-to-Cypher Evaluation and the Autonomy Divergence}

\author{
  \textbf{Yuzhe Zhang}$^{1,3}$ \quad
  \textbf{Weijie Zhu}$^{2}$ \quad
  \textbf{Haolin Yang}$^{1}$ \quad
  \textbf{Ziyun Zhang}$^{1,3}$ \quad
  \textbf{Xianwei Xue}$^{2}$ \\
  \textbf{Mengke Chen}$^{2}$ \quad
  \textbf{Qiutong Pan}$^{2}$ \quad
  \textbf{Huaqian Cai}$^{1,3}$\thanks{Corresponding author: \texttt{caihuaqian@pku.edu.cn}} \\[0.5em]
  $^{1}$Peking University \quad
  $^{2}$Baidu Inc. \\
  $^{3}$National Key Lab of Data Space Technology and System
}

\begin{document}
\maketitle

% Float placement parameters (prevent all figures from drifting to the end)
\renewcommand{\topfraction}{0.9}
\renewcommand{\dbltopfraction}{0.9}
\renewcommand{\bottomfraction}{0.5}
\renewcommand{\textfraction}{0.05}
\renewcommand{\floatpagefraction}{0.8}
\renewcommand{\dblfloatpagefraction}{0.8}
\setcounter{topnumber}{4}
\setcounter{dbltopnumber}{4}
\setcounter{totalnumber}{8}

\begin{abstract}
Graph databases are increasingly queried through natural language, yet every existing benchmark evaluates isolated single-turn queries rather than the multi-turn sessions through which analysts actually work.
We introduce \textbf{\graphturn}, the first benchmark for conversational Text-to-Cypher evaluation, comprising 721 sessions and 5,927 turns across 7 knowledge graphs and 13 conversational phenomena.
We evaluate 15 models under a guided oracle protocol and a fully autonomous agentic protocol, yielding four findings.
First, the best model reaches only 64.7\% execution accuracy, and session-level correctness remains below 5\%.
Second, despite strong overall rank correlation, frontier models exhibit a consequential reordering of the top of the leaderboard under autonomous operation, a phenomenon we term the \textbf{Autonomy Divergence}, which reveals error-management as a partially independent capability from raw generation skill.
Third, scaling action budgets from $\times$3 to $\times$10 fails to close the autonomy gap, as the strongest frontier models self-limit to approximately two actions per turn regardless of available budget.
Fourth, single-turn Cypher fine-tuning degrades multi-turn instruction following, while architecture-appropriate specialization outperforms several frontier models.
These results establish \graphturn{} as an open challenge for conversational graph database reasoning. Code and data are available at \url{https://github.com/BarryQ/CypherTurn}.
\end{abstract}

\section{Introduction}
\label{sec:intro}

\begin{figure}[htbp]
  \centering
  \includegraphics[width=\columnwidth]{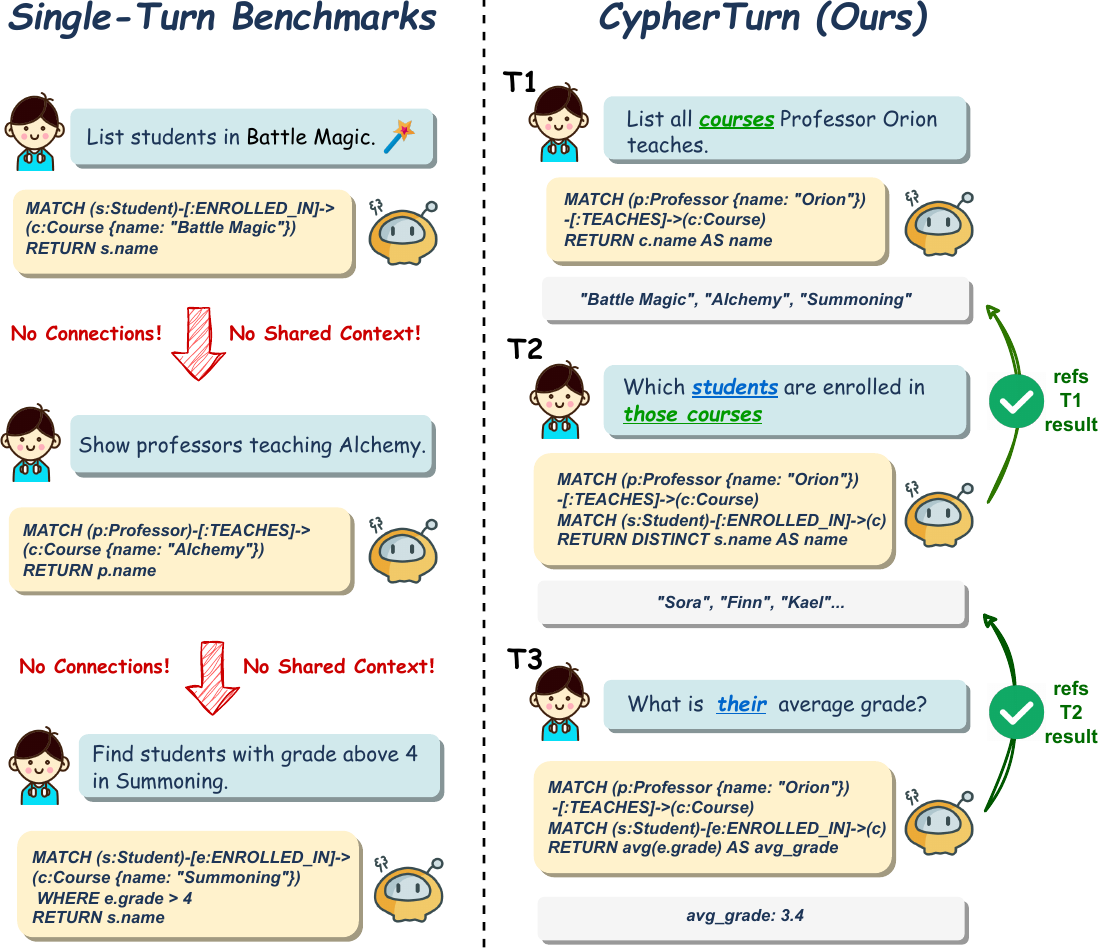}
  \caption{Existing single-turn benchmarks evaluate each query in isolation (left); \graphturn{} evaluates chain-dependent turns in sequence, where each query must be grounded in its predecessor's result (right).}
  \label{fig:intro_overview}
\end{figure}

Knowledge graphs have become foundational infrastructure in enterprise analytics, scientific discovery, and digital humanities. Deploying natural language interfaces over graph databases requires models to translate free-form user questions into Cypher queries, the declarative query language powering Neo4j and related property graph systems \citep{francis2018cypher}. In practice, however, users rarely retrieve the information they need from a single self-contained question. Real analytical work unfolds as a sequence of related queries, each anchored to what was returned before: a product analyst exploring a company knowledge graph might first ask which departments had the highest headcount last year, then which projects those teams worked on, then which clients each project served, and finally what revenue each client contributed. As illustrated in Figure~\ref{fig:intro_overview}, a model asked \textit{``which students are enrolled in those courses?''} must resolve \textit{``those courses''} against the prior turn's result set, and a follow-up \textit{``what is their average grade?''} must further resolve \textit{``their''} against the students just retrieved.

Existing Text-to-Cypher benchmarks do not capture this conversational structure: every published dataset in this space evaluates models on fully independent single-turn queries, and CypherBench \citep{hou2025cypherbench}, the largest and most widely used, is single-turn by design. A parallel line of work on knowledge-graph reasoning has explored LLM agents that traverse KGs interactively---e.g., Think-on-Graph \citep{thinkongraph2024} and Reasoning on Graphs \citep{rog2024}---but these systems still assume self-contained input questions and provide no protocol for multi-turn dialogue. Recent work such as PankRAG \citep{li2026pankrag} further models dependencies among sub-questions for graph retrieval, but remains single-query rather than conversational. The field therefore has no way to measure the capability that real deployment actually demands: coherent reasoning across a session in which each query depends on what came before. Strong single-turn accuracy reveals little about whether a model can resolve cross-turn references, propagate intermediate result sets, and stay grounded as the conversation unfolds.

The multi-turn setting also raises a second, more subtle challenge. Once a model is responsible for managing its own query history, an error at any turn corrupts the context available to every turn that follows, and small mistakes can cascade into session-wide failure. No existing benchmark measures this error propagation systematically, nor provides the infrastructure to separate pure query-generation ability from robustness to accumulated context errors.

To address both gaps, we introduce \textbf{\graphturn}, the first multi-turn benchmark for conversational Text-to-Cypher evaluation with rigorous human verification, where each turn's Cypher must reference the result of its predecessor. We make three contributions:

$\bullet$ We introduce the \textbf{first multi-turn Text-to-Cypher benchmark} comprising 721 sessions, 5,927 turns, seven purpose-built knowledge graphs, a taxonomy of 13 graph-native conversational phenomena, and 9 chain modes that characterise how each query depends on its predecessor.

$\bullet$ We design two complementary evaluation protocols: a \textbf{Guided Protocol} that isolates query generation ability via oracle context, and an \textbf{Agentic Protocol} that requires models to manage their own multi-turn history, mirroring real deployment conditions.

$\bullet$ We evaluate \textbf{15 frontier and specialized models} under both protocols and identify the \textbf{Autonomy Divergence}: despite strong overall rank correlation, frontier models exhibit a consequential reordering of the top of the leaderboard under autonomous operation, revealing error-management as a partially independent capability from raw Cypher generation skill. We further find that single-turn Cypher fine-tuning actively degrades multi-turn instruction following, while architecture-appropriate domain specialization can outperform several frontier models.
\section{The \graphturn{} Benchmark}
\label{sec:benchmark}

\graphturn{} is, to our knowledge, the first Text-to-Cypher benchmark simultaneously offering multi-turn evaluation, contextual anaphora annotation, and a fully autonomous agentic evaluation protocol (Table~\ref{tab:benchmark_comparison}).

\begin{table}[htbp]
\centering
\resizebox{\columnwidth}{!}{%
\begin{tabular}{lccccc}
\toprule
\textbf{Benchmark} & \textbf{MT} & \textbf{Cypher} & \textbf{Ctx} & \textbf{Agent} & \textbf{T/S} \\
\midrule
CypherBench         & \xmarkc & \cmarkc & \xmarkc & \xmarkc & 1.0 \\
ZOGRASCOPE          & \xmarkc & \cmarkc & \xmarkc & \xmarkc & 1.0 \\
SM3-Text-to-Query   & \xmarkc & \cmarkc & \xmarkc & \xmarkc & 1.0 \\
Text2GQL-Bench      & \xmarkc & \cmarkc & \xmarkc & \xmarkc & 1.0 \\
MTGQL$^\dagger$     & \cmarkc & \xmarkc & \xmarkc & \xmarkc & {6.5} \\
\midrule
\rowcolor[HTML]{F2F2FE}
\textbf{\graphturn{}} & \cmarkc & \cmarkc & \cmarkc & \cmarkc & \textbf{8.2} \\
\bottomrule
\end{tabular}%
}
\caption{Comparison of \graphturn{} against existing Text-to-Cypher and graph query benchmarks. \textbf{MT}: multi-turn sessions; \textbf{Cypher}: Neo4j Cypher target; \textbf{Ctx}: contextual anaphora evaluation; \textbf{Agent}: agentic protocol; \textbf{T/S}: average turns per session. $^\dagger$MTGQL targets nGQL, not Cypher.}
\label{tab:benchmark_comparison}
\end{table}

We construct seven unique knowledge graphs, each embodying a distinct graph topology to surface challenges specific to property graph traversal, and annotate each session with a taxonomy of 13 conversational phenomena capturing the full range of chain-dependent reasoning challenges. Every session is assigned a naturalistic persona to diversify utterance register, and all gold queries undergo a two-stage human review before inclusion.

\begin{figure*}[htbp]
  \centering
  \includegraphics[width=\textwidth]{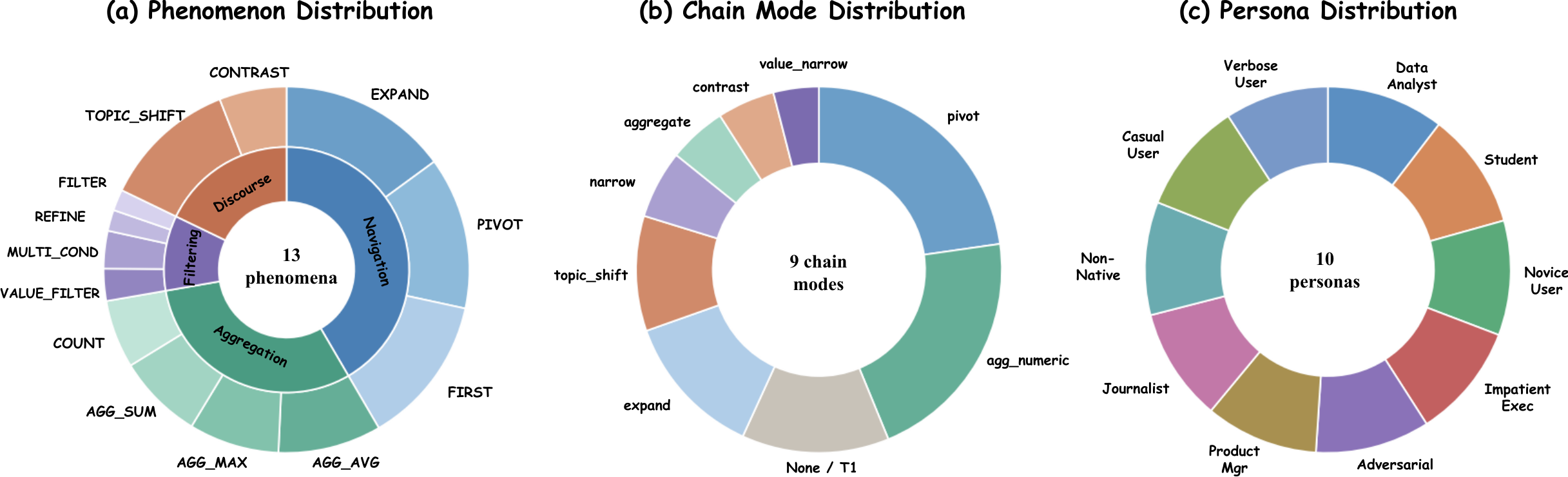}
  \caption{Benchmark composition.
  \textbf{(a)} Phenomenon distribution across 5,206 phenomenon-bearing turns (721 first turns carry no phenomenon; note: the \textsc{First} phenomenon refers to rank-then-select navigation, not to first turns): inner ring shows four categories (Navigation, Aggregation, Filtering, Discourse); outer ring shows individual phenomena.
  \textbf{(b)} Chain mode distribution across all 5,927 turns.
  \textbf{(c)} Persona distribution across 721 sessions; ten types are sampled near-uniformly.}
  \label{fig:dataset_dists}
\end{figure*}

\subsection{Knowledge Graphs}
\label{sec:kg_construction}

\graphturn{} comprises seven purpose-built knowledge graphs, each designed around a topological pattern that poses a distinct reasoning challenge. All graphs use synthetic worlds, ensuring every correct answer must be derived from the graph itself. Three graphs target structural complexity: \textbf{Ancient Empire} features a directed-cycle topology with dense numeric edge properties; \textbf{Ocean Kingdom} uses a diamond-convergence structure that creates path ambiguity; and \textbf{Magic Academy} introduces a bipartite cycle requiring precise entity resolution under structural ambiguity. Two graphs stress recursive reasoning: \textbf{Arcane Archive} embeds a citation loop enabling recursive multi-hop queries; and \textbf{Celestial Court} models a bureaucratic hierarchy with officials supervising other officials at arbitrary depth. The remaining two cover complementary patterns: \textbf{Stellar Colony} pairs a tree hierarchy with lateral alliance edges between peer factions; and \textbf{Merchant Harbor} uses a central invoice node bridging traders, buyers, and storehouses with rich numeric edge properties. Full schema details appear in Appendix~\ref{app:graph_schemas}.

\subsection{Conversational Phenomena}
\label{sec:phenomena}

We design 13 distinct conversational phenomena to capture the full range of multi-turn graph querying challenges, organised into four categories that reflect different types of chain-dependent reasoning (shown in Figure~\ref{fig:dataset_dists} a ). \textbf{Navigation} phenomena are the most graph-native: they require the model to traverse edges outward from the prior result set, sometimes after ranking entities by a property to identify a specific target before traversal. \textbf{Aggregation} phenomena ask the model to compute numerical summaries---means, maxima, sums, or counts---over a prior result set, testing whether entity references are correctly carried forward into aggregate queries. \textbf{Filtering} phenomena narrow prior results by adding attribute constraints, ranging from single-condition filters to compound multi-criteria conditions on numeric or categorical properties. \textbf{Discourse} phenomena test context management at the session level: one type queries the logical complement of a prior result, while the other marks a fresh independent query and is the only non-chain-dependent phenomenon in the taxonomy. Full definitions and Cypher examples appear in Appendix~\ref{app:phenomena}.

\subsection{Chain Modes}
\label{sec:chain_modes}

\begin{tcolorbox}[colback=gray!5, colframe=gray!60, boxrule=0.4pt,
  left=4pt, right=4pt, top=3pt, bottom=3pt, title={\small Annotation Example: Phenomenon vs.\ Chain Mode}]
\small
\textbf{Turn 3} (\textit{ancient\_empire}, persona: \textit{impatient\_exec})\\
\textbf{User:} ``Give me the highest population province's provinces now.''\\
\textbf{Phenomenon:} \textsc{PIVOT} (navigation) --- rank by property, then traverse\\
\textbf{Chain mode:} \textit{pivot} --- rank-then-traverse from prior result set\\[4pt]
\textbf{Turn 4} (\textit{ancient\_empire}, persona: \textit{novice\_user})\\
\textbf{User:} ``um, can I just see the ones with an occupation?''\\
\textbf{Phenomenon:} \textsc{REFINE} (filtering) --- narrow prior result by attribute\\
\textbf{Chain mode:} \textit{narrow} --- categorical attribute filter on prior set\\[4pt]
Both \textsc{FILTER} and \textsc{REFINE} map to chain mode \textit{narrow}; the phenomenon distinguishes whether the filter is new or iteratively tightens a prior constraint.
\end{tcolorbox}

Each chain-dependent turn is further characterised by a \textbf{chain mode} that specifies how the current query relates to the prior result (shown in Figure~\ref{fig:dataset_dists} b ). Phenomena and chain modes form two complementary annotation layers: phenomena are fine-grained linguistic labels capturing query intent (13 types), while chain modes are coarser construction-time labels describing how a turn structurally depends on its predecessor (9 types). Multiple phenomena may map to the same chain mode. \graphturn{} defines 9 chain modes spanning two non-chain-dependent types (\textit{none} at T1 and \textit{topic\_shift}) and seven chain-dependent types: \textit{expand} (one-hop traversal from the prior result), \textit{pivot} (rank-then-traverse), \textit{agg\_numeric} (numeric aggregation over the prior set), \textit{aggregate} (cardinality count), \textit{contrast} (logical complement), \textit{value\_narrow} (numeric threshold filter), and \textit{narrow} (categorical attribute filter). Full chain mode definitions and Cypher examples appear in Appendix~\ref{app:chain_modes}.

\begin{figure*}[htbp]
  \centering
  \includegraphics[width=0.95\textwidth]{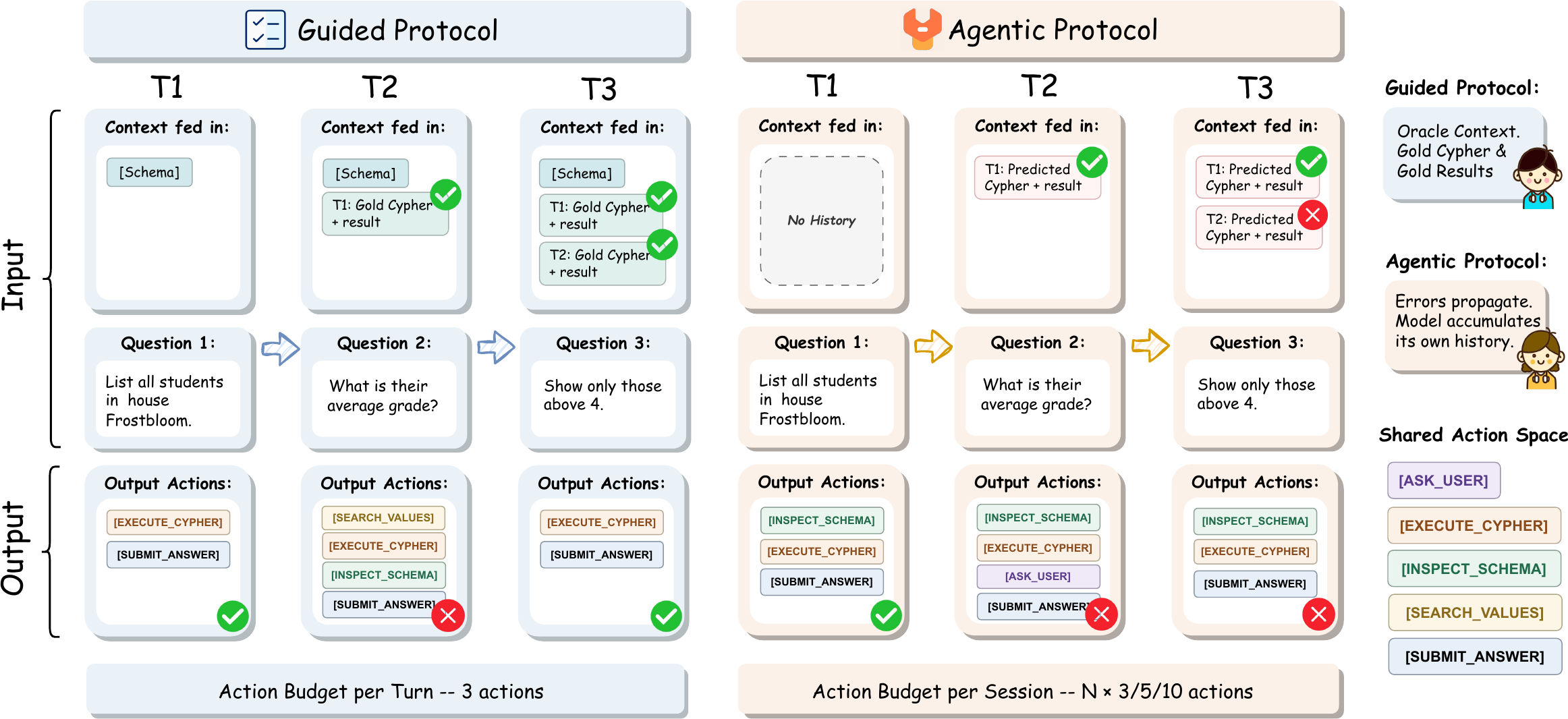}
  \caption{The two evaluation protocols in \graphturn{}. The \textbf{Guided Protocol}
  (left) supplies gold-annotated prior turns as oracle context, isolating raw
  Cypher generation ability. The \textbf{Agentic Protocol} (right) requires
  models to autonomously accumulate their own history under a configurable
  shared session budget ($B = T \times m$); early-turn errors propagate forward through chain dependencies,
  compounding across the session.}
  \label{fig:protocols}
  \vspace{-10pt}
\end{figure*}

\subsection{Session Personas}
\label{sec:personas}

Each session is assigned a \textbf{persona} drawn uniformly from 10 types: journalist, student, product manager, novice user, data analyst, casual user, impatient executive, adversarial user, non-native speaker, and verbose user. Personas serve as a register-diversification device that diversifies utterance style rather than as an independently validated experimental factor. Personas govern the utterance register throughout the session, and as shown in Figure~\ref{fig:dataset_dists}(c) the ten types are sampled near-uniformly across the 721 sessions. The full dataset spans 5,927 turns with a mean of 8.2 turns per session. Full persona definitions and representative utterance examples appear in Appendix~\ref{app:personas}.

\subsection{Human Annotation and Quality Control}
\label{sec:annotation}
Utterances are generated programmatically from phenomenon-specific templates conditioned on persona register and prior-turn context, then optionally rewritten by DeepSeek-V3.2 for stylistic naturalness. The user simulator in the agentic protocol is also powered by DeepSeek-V3.2 with persona-specific system prompts.

All candidate sessions undergo a two-stage human review before inclusion in \graphturn{}. Two Cypher-proficient annotators independently verify each turn against three criteria: (1) the gold Cypher correctly captures the utterance intent, (2) cross-turn anaphora dependencies are unambiguous, and (3) the utterance is natural and well-formed for the assigned persona. Turns marked \textit{revise} are corrected and re-executed against a live Neo4j instance. Inter-annotator agreement on the binary accept/reject decision yields $\kappa = 0.83$ across all 5,927 turns, indicating strong agreement. Approximately 4.1\% of turns required revision and 2.3\% of candidate sessions were rejected outright. Every retained gold query is independently verified to execute correctly on the corresponding graph.

\section{Evaluation Framework}
\label{sec:eval_framework}

\subsection{Two Evaluation Protocols}
\label{sec:protocols}

We evaluate models under two protocols designed to measure complementary aspects of multi-turn Cypher generation (Figure~\ref{fig:protocols}).

\textbf{Guided Protocol.}
At each turn, the model receives the graph schema, the complete gold-annotated history of all prior turns comprising the prior gold natural-language utterances and their gold execution results, and the current user utterance. A budget of \textbf{3 actions per turn} is enforced to encourage direct generation. This protocol provides an oracle upper bound on performance, decoupling Cypher generation skill from the ability to manage accumulated, potentially corrupted context.

\textbf{Agentic Protocol.}
The model operates as an autonomous agent over an entire session. No oracle context is provided: the model must discover the graph schema, maintain its own prediction history, and manage a \textbf{shared action budget} of $B = T \times m$ for a session of $T$ turns, where $m \in \{3, 5, 10\}$. Under Agentic $\times$3 this averages three actions per turn, but the model may freely allocate more or fewer to any individual turn. When the model invokes \texttt{ASK\_USER}, a persona-aware user simulator responds appropriately. This protocol mirrors the information structure of deployment, where the model must self-correct errors, plan across a long horizon, and handle context that degrades as mistakes compound.

Both protocols use the same \textbf{5-action space}: \texttt{EXECUTE\_CYPHER} (run Cypher against Neo4j and receive result rows), \texttt{ASK\_USER} (pose a clarifying question), \texttt{INSPECT\_SCHEMA} (retrieve node labels and relation types), \texttt{SEARCH\_VALUES} (fuzzy-search entity property values by keyword), and \texttt{SUBMIT\_ANSWER} (finalize the predicted query for the current turn). Full protocol specifications, including prompt templates and action-space definitions, appear in Appendix~\ref{app:eval_protocols}.

\subsection{Metrics}
\label{sec:metrics}

We report metrics at both the turn level and the session level.

\textbf{Execution Accuracy (EX)} is our primary metric. For a predicted query $\hat{q}_t$ and gold query $q_t$ executed over graph $\mathcal{G}$:
\begin{equation}
  \mathrm{EX}_t = \mathbf{1}\!\left[\mathrm{exec}(\hat{q}_t, \mathcal{G}) = \mathrm{exec}(q_t, \mathcal{G})\right]
\end{equation}
where equality requires identical rows \emph{and} column names. EX is strict but unambiguous.

\textbf{Provenance Subgraph Jaccard Similarity (PSJS)} measures structural correctness at the graph traversal level:
\begin{equation}
  \mathrm{PSJS}_t = \frac{|V(\hat{q}_t) \cap V(q_t)|}{|V(\hat{q}_t) \cup V(q_t)|}
\end{equation}
where $V(q_t)$ is the set of node element-IDs bound during pattern matching in $q_t$ (for aggregation queries returning scalar values, $V(q_t)$ comprises the nodes matched by the \texttt{MATCH} clause). PSJS serves as a diagnostic for cases where a model retrieves the correct subgraph but fails at output formatting.

\textbf{Chain Error Rate (CER)} quantifies error propagation through session history. For all turns whose immediately preceding turn failed:
\begin{equation}
  \mathrm{CER} = P\!\left(\mathrm{EX}_t = 0 \;\middle|\; \mathrm{EX}_{t-1} = 0\right)
\end{equation}
A high CER indicates that failures cascade: once a turn fails, the next turn is likely to fail as well. CER is computed over all turns whose immediately preceding turn scored $\mathrm{EX}=0$, so the first turn of each session and turns that follow a successful turn do not contribute.

\textbf{Session Exact Match (SEM)} is a session-level binary metric: a session receives $\mathrm{SEM}=1$ only if every turn achieves $\mathrm{EX}=1$, and 0 otherwise. Full per-model numeric results under all three agentic budgets appear in Appendix~\ref{app:detailed_results}.

\section{Experiments}
\label{sec:experiments}

\subsection{Experimental Setup}
\label{sec:setup}

We evaluate fifteen models---10 frontier API models, 2 small open-source baselines, and 3 fine-tuned or specialized architectures (full descriptions in Appendix~\ref{app:model_list}). All models are accessed via their respective APIs or deployed locally with identical vLLM configurations. CypherRI-7B is excluded from aggregate statistics due to fill-in-the-middle token contamination; its results appear in tables marked with $\dagger$.

\subsection{Main Results}
\label{sec:main_results}

\begin{figure*}[t]
  \centering
  \includegraphics[width=\textwidth]{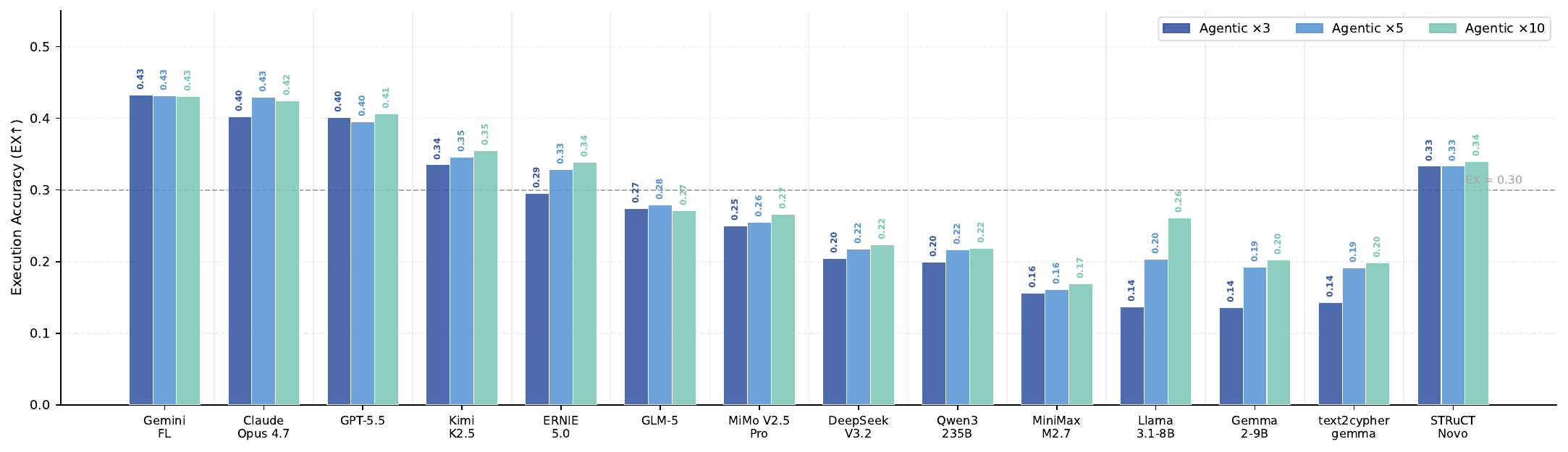}
  \caption{Agentic EX under $\times$3 / $\times$5 / $\times$10 action budgets for 14 evaluated models. Full numeric tables in Appendix~\ref{app:retry}.}
  \label{fig:budget_lines}
\end{figure*}

Table~\ref{tab:guided_agentic3} presents the full turn-level results under both protocols, and Figure~\ref{fig:budget_lines} shows agentic EX under all three action budgets for all evaluated models. Full numeric tables appear in Appendix~\ref{app:retry}.

\begin{table*}[t]
\centering
\resizebox{\textwidth}{!}{%
\begin{tabular}{l ccccc ccccc c}
\toprule
& \multicolumn{5}{c}{\textbf{Guided Protocol}} & \multicolumn{5}{c}{\textbf{Agentic $\times$3}} & \\
\cmidrule(lr){2-6}\cmidrule(lr){7-11}
\textbf{Model}
  & EX$\uparrow$ & PSJS$\uparrow$ & CER$\downarrow$ & SEM$\uparrow$ & Tok/S
  & EX$\uparrow$ & PSJS$\uparrow$ & CER$\downarrow$ & SEM$\uparrow$ & Tok/S
  & $\Delta$EX$\downarrow$ \\
\midrule
\rowcolor[HTML]{F2F2FE}
\multicolumn{12}{c}{\textit{Frontier Models}} \\
Claude Opus 4.7       & \textbf{0.647} & \textbf{0.846} & 0.434 & \textbf{0.046} & 617 & 0.402 & 0.681 & 0.712 & \textbf{0.019} & 639 & $-$0.245 \\
GPT-5.5               & 0.625 & 0.828 & \textbf{0.424} & 0.025 & 504 & 0.401 & 0.673 & 0.698 & 0.006 & 554 & $-$0.224 \\
Kimi-K2.5             & 0.602 & 0.819 & 0.469 & 0.028 & 753 & 0.335 & 0.612 & 0.749 & 0.006 & 965 & $-$0.267 \\
Gemini-3.1-Flash-Lite & 0.572 & 0.744 & 0.508 & 0.017 & \textbf{500} & \textbf{0.432} & \textbf{0.698} & \textbf{0.695} & 0.008 & \textbf{427} & \textbf{$-$0.140} \\
Qwen3-235B            & 0.513 & 0.737 & 0.466 & 0.006 & 896 & 0.199 & 0.548 & 0.833 & 0.001 & 927 & $-$0.314 \\
GLM-5                 & 0.506 & 0.816 & 0.496 & 0.003 & 926 & 0.274 & 0.623 & 0.773 & 0.006 & 749 & $-$0.232 \\
DeepSeek-V3.2         & 0.506 & 0.729 & 0.474 & 0.004 & 894 & 0.204 & 0.551 & 0.835 & 0.000 & 965 & $-$0.302 \\
ERNIE-5.0             & 0.468 & 0.676 & 0.575 & 0.003 & 976 & 0.295 & 0.584 & 0.830 & 0.003 & 867 & $-$0.173 \\
MiMo-V2.5-Pro             & 0.460 & 0.752 & 0.531 & 0.003 & 969 & 0.250 & 0.591 & 0.781 & 0.001 & 871 & $-$0.210 \\
MiniMax-M2.7          & 0.139 & 0.379 & 0.856 & 0.000 & 1{,}644 & 0.156 & 0.403 & 0.862 & 0.000 & 1{,}805 & \textit{$+$0.018} \\
\midrule
\rowcolor[HTML]{F2F2FE}
\multicolumn{12}{c}{\textit{Small Open-Source Models}} \\
Llama-3.1-8B          & 0.291 & 0.456 & 0.764 & 0.000 & 1{,}079 & 0.137 & 0.371 & 0.962 & 0.000 & 924 & $-$0.154 \\
Gemma-2-9B            & 0.281 & 0.475 & 0.742 & 0.000 & 1{,}082 & 0.136 & 0.378 & 0.948 & 0.000 & 844 & $-$0.145 \\
\midrule
\rowcolor[HTML]{F2F2FE}
\multicolumn{12}{c}{\textit{Fine-tuned / Specialized Models}} \\
STRuCT-LLM-Novo       & 0.526 & 0.763 & 0.506 & 0.015 & 1{,}902 & 0.333 & 0.680 & 0.766 & 0.003 & 1{,}346 & $-$0.193 \\
text-to-cypher-gemma  & 0.261 & 0.498 & 0.759 & 0.000 & 1{,}054 & 0.143 & 0.369 & 0.934 & 0.000 & 792 & $-$0.118 \\
CypherRI-7B$^\dagger$ & 0.013 & 0.283 & 0.989 & 0.000 & 29{,}380 & 0.001 & 0.068 & 0.999 & 0.000 & 23{,}495 & $-$0.012 \\
\bottomrule
\end{tabular}%
}
\caption{Turn-level results under the Guided and Agentic $\times$3 protocols.
\textbf{Tok/S}: mean clean output tokens per session (excl.\ \texttt{<think>}).
\textbf{Bold}: best per column (excluding $\dagger$).
$\dagger$ CypherRI-7B's near-zero EX and inflated token count both reflect a format compliance failure caused by fill-in-the-middle token contamination in its RL training; excluded from aggregate statistics.}
\label{tab:guided_agentic3}
\end{table*}

Under the guided protocol, Claude Opus~4.7 leads at 0.647 EX. MiniMax-M2.7 is an outlier at 0.139, a score that likely reflects format-compliance issues. Across the remaining frontier models the spread is roughly 19 points. Session-level correctness is rare: Claude leads with an SEM of 0.046, the highest of any model, representing 33 of 721 sessions. The specialized STRuCT-LLM-Novo reaches 0.526 and surpasses six frontier models.

Every model declines under the agentic protocol except MiniMax-M2.7, whose gain of 0.018 is a floor effect that follows from its near-chance guided EX. Among frontier models, the guided-to-agentic drop ranges from 0.140 for Gemini-3.1-Flash-Lite to 0.314 for Qwen3-235B. Crucially, the rankings reorder under autonomy. Gemini-3.1-Flash-Lite rises from fourth place under the guided protocol to first under the agentic protocol at 0.432, while Qwen3-235B falls from fifth to near the bottom at 0.199. Claude Opus~4.7 and GPT-5.5 are separated by only 0.001 EX under autonomy, a difference within run-to-run variance, and the two effectively tie for second behind Gemini. All gaps and the reordering between Gemini and Claude are significant at the 95\% level under a 10{,}000-resample bootstrap, and we analyze this reordering, which we term the Autonomy Divergence, in \S\ref{sec:avalanche}. Expanding the budget from $\times$3 to $\times$10 yields limited frontier gains but benefits small models substantially, as Llama-3.1-8B nearly doubles to 0.261 in Figure~\ref{fig:budget_lines}.

\section{Analysis}\label{sec:analysis}

\begin{figure*}[t]
  \centering
  \includegraphics[width=0.95\textwidth]{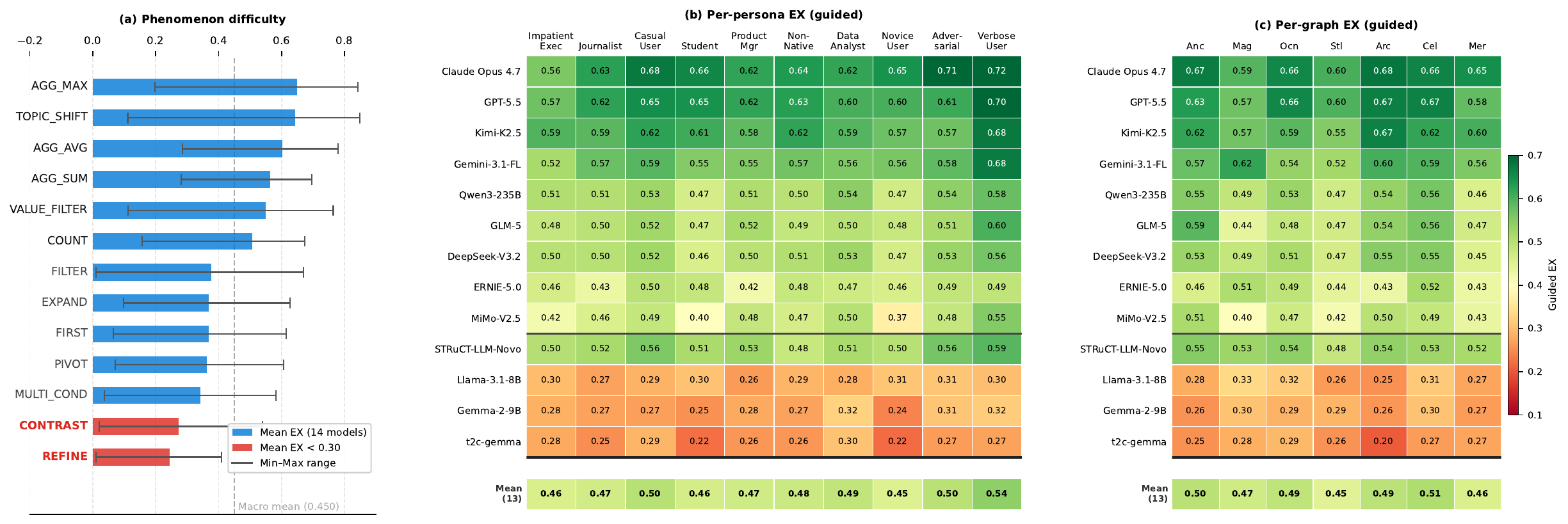}
  \caption{Guided EX across three dimensions for 14 models (CypherRI-7B excluded; MiniMax-M2.7 additionally excluded from heatmaps).
  \textbf{(a)} Per-phenomenon mean EX with min--max range; dashed line shows the unweighted macro mean across 13 phenomena.
  \textbf{(b)} Per-persona heatmap (13 models $\times$ 10 personas).
  \textbf{(c)} Per-graph heatmap (13 models $\times$ 7 graphs).}
  \label{fig:triptych}
\end{figure*}

\subsection{The Difficulty Landscape}
\label{sec:phenomena_analysis}

Figure~\ref{fig:triptych} presents guided EX across three dimensions. Phenomenon type is the strongest predictor of turn difficulty: \textsc{Refine} averages 0.244 across 14 models and \textsc{Contrast} averages 0.273, while \textsc{Agg\_Max} averages 0.650 and \textsc{Agg\_Avg} 0.602. Persona and graph topology are secondary factors, with within-model EX spread across personas typically ten to sixteen points for frontier models. Per-graph difficulty tracks structural entity-membership ambiguity: Stellar Colony and Merchant Harbor are hardest at 0.426 and 0.434, while Celestial Court reaches 0.485 (Table~\ref{tab:graph_ex}).

A distinct source of difficulty also emerges on the output side: the gap between PSJS and EX. Claude Opus~4.7 leads PSJS at 0.846 yet achieves only 0.647 EX, and its SEM of 0.046 represents just 33 fully correct sessions out of 721. Output formatting is independent of traversal correctness and remains unsolved.

\subsection{The Autonomy Divergence: Why Rankings Shift}
\label{sec:avalanche}

\begin{figure*}[t]
  \centering
  \includegraphics[width=\textwidth]{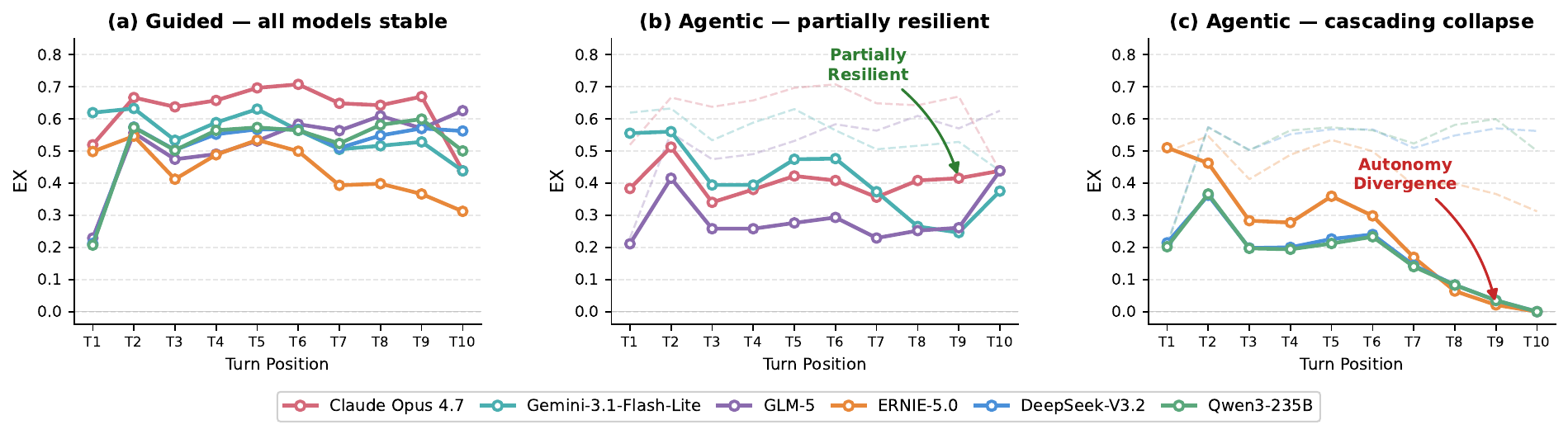}
  \caption{EX by turn position for six frontier models.
  \textbf{(a)} Guided protocol: all models remain stable across turns.
  \textbf{(b)} Agentic protocol: Claude, Gemini, and GLM-5 plateau at non-trivial EX (dashed lines show their guided baselines).
  \textbf{(c)} Agentic protocol: ERNIE, DeepSeek, and Qwen3 collapse to near-zero by $T_{10}$.}
  \label{fig:avalanche}
\end{figure*}

Guided and agentic rankings are strongly correlated overall, with a Spearman $\rho$ of 0.86 across all 14 models. This global figure is propped up by the weakest models, which sit stably at the bottom under both protocols. Restricted to the ten frontier models, $\rho$ drops to 0.73, and to 0.63 (95\% CI [0.58, 0.70]) after excluding MiniMax-M2.7. Within the frontier tier, 11 of the 45 pairwise comparisons reverse significantly under a 10{,}000-resample bootstrap, including Gemini overtaking Claude for the top position. We therefore frame the Autonomy Divergence as a differential-robustness effect rather than a systemic reversal: autonomy disproportionately degrades a few frontier models, enough to change the top of the leaderboard, while the overall order holds. The Autonomy Divergence isolates a graph-specific mechanism in which each turn's entity set seeds the next, so that an early error changes what later queries compute over rather than merely adding independent noise. We further test three protocol variants in Appendix~\ref{app:protocol_ablations}, and all preserve the overall ranking.

CER quantifies the cascade. Even under the guided protocol, where gold context insulates each turn from error propagation, frontier models show a median conditional CER of 0.47, excluding MiniMax-M2.7 (computed from the model's own per-turn EX, not the injected oracle context), indicating that chain-dependent turns cluster their failures intrinsically. Under the agentic protocol this figure rises to 0.77. For DeepSeek-V3.2 and Qwen3-235B it exceeds 0.83, which means that once a turn fails, the next one fails more than 83\% of the time.

Resilience tracks a schema-heavy action strategy, which we interpret as a behavioral marker of resilient models rather than the cause of resilience. The resilient models recalibrate graph structure on every turn, and Claude and Gemini-3.1-Flash-Lite devote 85.8\% and 83.5\% of their actions to \texttt{INSPECT\_SCHEMA}. The collapsing models instead spend 62--73\% of their actions on \texttt{EXECUTE\_CYPHER} retries without recovering the schema, which leaves them exposed to compounding corruption. As Table~\ref{tab:inspect_vs_ex} shows, \texttt{INSPECT\_SCHEMA} allocation correlates with agentic EX at a Spearman $\rho$ of 0.93 ($p{<}0.001$), stable above 0.88 under leave-one-out removal. Protocol variants that make the schema persistent or disclose the session horizon do not eliminate the gap, as confirmed by the ablations in Appendix~\ref{app:protocol_ablations}.

Figure~\ref{fig:avalanche} makes the divergence visible turn by turn. Under the guided protocol (panel a), all six frontier models remain stable across $T_1$--$T_{10}$, confirming that oracle context quarantines per-turn errors. The agentic trajectories then split into two trends. Claude and Gemini-3.1-Flash-Lite plateau without collapsing (panel b); GLM-5 dips at $T_3$ but recovers to roughly 0.44 rather than continuing to decline. ERNIE-5.0, DeepSeek-V3.2, and Qwen3-235B trace a different curve (panel c): each rises to 0.36--0.51 by $T_2$ before degrading almost monotonically to near-zero EX by $T_{10}$. The same chain-dependent failures that oracle context absorbs become session-wide once the model carries its own history.

\begin{table}[t]
\centering
\small
\resizebox{0.85\columnwidth}{!}{%
\begin{tabular}{l c c c}
\toprule
\textbf{Model} & \textbf{INSPECT\%} & \textbf{Agentic EX} & \textbf{$\Delta$EX} \\
\midrule
\rowcolor[HTML]{E8F5E9} GPT-5.5           & 94.1 & 0.401 & 0.224 \\
\rowcolor[HTML]{E8F5E9} Claude Opus 4.7   & 85.8 & 0.402 & 0.245 \\
\rowcolor[HTML]{E8F5E9} Gemini-3.1-FL     & 83.5 & 0.432 & 0.140 \\
\rowcolor[HTML]{FFFDE7} GLM-5            & 54.3 & 0.274 & 0.232 \\
\rowcolor[HTML]{FFFDE7} STRuCT-LLM-Novo  & 40.4 & 0.333 & 0.193 \\
\rowcolor[HTML]{FFFDE7} Kimi-K2.5        & 34.8 & 0.335 & 0.267 \\
MiMo-V2.5-Pro    & 32.5 & 0.250 & 0.210 \\
ERNIE-5.0        & 29.3 & 0.295 & 0.173 \\
Qwen3-235B       & 29.2 & 0.199 & 0.314 \\
DeepSeek-V3.2    & 28.9 & 0.204 & 0.302 \\
\rowcolor[HTML]{FFEBEE} MiniMax-M2.7     & 23.8 & 0.156 & --- \\
\rowcolor[HTML]{FFEBEE} Gemma-2-9B       & 23.1 & 0.136 & 0.145 \\
\rowcolor[HTML]{FFEBEE} t2c-gemma        & 23.1 & 0.143 & 0.118 \\
\rowcolor[HTML]{FFEBEE} Llama-3.1-8B     & 15.9 & 0.137 & 0.154 \\
\bottomrule
\end{tabular}%
}
\caption{INSPECT\_SCHEMA fraction vs.\ Agentic $\times$3 EX, sorted by INSPECT\%. \colorbox[HTML]{E8F5E9}{Green}: schema-heavy; \colorbox[HTML]{FFFDE7}{yellow}: mixed; \colorbox[HTML]{FFEBEE}{red}: schema-sparse.}
\label{tab:inspect_vs_ex}
\vspace{-20pt}
\end{table}

\subsection{The Budget Ceiling: Why Extra Actions Don't Help}
\label{sec:budget_ceiling}

If cascading errors drive the autonomy gap, more actions to recover should help. Figure~\ref{fig:budget_lines} shows otherwise. Gemini-3.1-Flash-Lite achieves 0.432, 0.431, and 0.430 across $\times$3, $\times$5, and $\times$10, remaining flat despite a 3.3-fold budget increase. Claude peaks at $\times$5 with 0.429 and GLM-5 actually degrades at $\times$10 to 0.271. Qwen3-235B gains only 1.9 points across the full range, negligible against its 31-point guided gap. Small open-source models do benefit from extra budget (Llama-3.1-8B nearly doubles to 0.261), confirming that the budget-scaling curve separates exploration-limited failures from strategic-deficiency failures.

Even at $\times$10, no model closes its own guided-to-agentic gap: Gemini-3.1-Flash-Lite remains 0.142 points below its guided EX, and Claude 0.223 points below. The schema-heavy frontier models self-limit to approximately two actions per turn regardless of available budget, producing repeated ineffective retries rather than productive exploration; the collapsing frontier models spend three to four actions per turn, yet almost all of it on repeated retries that fail to localize the true error.

A natural concern is that this self-limitation reflects the efficiency instruction in the action prompt rather than a genuine capability gap. We test this by replacing the efficiency instruction with explicit encouragement to use the full budget, run at tenfold budget. Execution accuracy drops for six of the seven valid frontier models, because the extra actions are largely repeated retries and unfocused searches that introduce noise rather than localizing the true error. The bottleneck is ineffective error recovery, not unwillingness to spend budget (Appendix~\ref{app:protocol_ablations}).

The hard budget cliff contributes little to the frontier gap: excluding turns auto-scored zero from budget exhaustion shifts scores by at most 3.3 points for the four models anchoring the main analysis, and the largest adjustment across the frontier is 6.5 points for ERNIE-5.0. The cliff dominates only the small models, where exhaustion rates exceed 80\%, and it adjusts absolute scores without reordering the frontier.

\subsection{The Specialization Paradox}
\label{sec:specialization}

Single-turn Cypher fine-tuning degrades multi-turn performance. text-to-cypher-gemma achieves a guided EX of only 0.261, which falls below the 0.281 of its base model Gemma-2-9B. The Neo4j Text2Cypher corpus implicitly teaches that every query is self-contained, and this bias breaks anaphora resolution and result-set chaining in multi-turn settings. The fine-tuned model repeatedly hardcodes prior results as literal value lists instead of maintaining relational continuity.

Two controlled base-to-specialised pairs, summarised in Table~\ref{tab:specialization_pairs}, test whether the decisive factor is the training paradigm or the domain specificity of the training data. The Gemma pair moves downward: single-turn SFT lowers guided EX from 0.281 to 0.261. The QwQ pair moves in the opposite direction: STRuCT-LLM-Novo, built on QwQ-32B (a 32B open-weight reasoning model from Alibaba's Qwen team), uses RL with chain-of-thought supervision and cross-formalism transfer between SQL and Cypher to raise guided EX from 0.395 to 0.526, a gain of 13.1 points. The same direction holds under autonomy, where QwQ-32B rises from an agentic EX of 0.324 to 0.333. At the opposite extreme, CypherRI-7B suffers from RL-induced format tokens that contaminate its output. What matters is therefore not whether the training data is domain-specific but whether the training paradigm matches the multi-turn nature of the task.

\begin{table}[t]
\centering
\small
\resizebox{\columnwidth}{!}{%
\begin{tabular}{l l c c}
\toprule
\textbf{Model} & \textbf{Paradigm} & \textbf{Guided} & \textbf{Agentic} \\
\midrule
Gemma-2-9B            & Base            & 0.281 & 0.136 \\
text-to-cypher-gemma  & Single-turn SFT & 0.261 & 0.143 \\
QwQ-32B$^\ddagger$     & Base            & 0.395 & 0.324 \\
STRuCT-LLM-Novo       & RL + CoT        & 0.526 & 0.333 \\
\bottomrule
\end{tabular}%
}
\caption{Two controlled base-to-specialised pairs. Single-turn SFT degrades the Gemma pair, while RL with chain-of-thought improves the QwQ pair, in opposite directions. $^\ddagger$ QwQ-32B is evaluated on the stratified 210-session subset of Appendix~\ref{app:protocol_ablations}, whose per-model EX deviates from the full set by at most 0.01; the 13.1-point gain far exceeds this subset noise.}
\label{tab:specialization_pairs}
\end{table}

\section{Conclusion}
\label{sec:conclusion}

We introduced \graphturn{}, the first multi-turn benchmark for conversational Text-to-Cypher evaluation, comprising 721 sessions and 5,927 gold-verified queries across seven knowledge graphs and 13 phenomena. Evaluation of fifteen models under guided and agentic protocols yields four findings. Performance ceilings are low: even the best model reaches only 64.7\% guided EX, and session-level correctness barely exceeds 4\%. The Autonomy Divergence shows that error-management under autonomous operation is partially independent of generation skill, producing a reordering of the top of the leaderboard invisible in oracle-context evaluation. Extra action budget does not close this gap, as frontier models self-limit regardless of resources. Finally, single-turn fine-tuning degrades multi-turn performance, while architecture-appropriate specialisation outperforms several frontier models. The uneven degradation and the instability of the top rank persist under three protocol variants that perturb schema availability, horizon disclosure, and prompt wording; each variant displaces Gemini-3.1-Flash-Lite from the top position, so the Autonomy Divergence is robust to the design choices of the agentic protocol even though the identity of the leader is not. These findings position \graphturn{} as an open challenge and motivate multi-turn training corpora, schema-recovery mechanisms, and finer-grained semantic equivalence metrics as next steps.

\section*{Limitations}
\label{sec:limitations}

\graphturn{} evaluates 15 models spanning frontier and specialized architectures, yet does not exhaust all competitive systems available at submission time; results for unevaluated frontier models may differ, and the rapid pace of model releases means that newer systems could shift the absolute scores reported here. All seven knowledge graphs are synthetic, so execution accuracy on \graphturn{} does not extrapolate to deployment accuracy on real enterprise graphs, where schema naming conventions, incomplete data, and domain-specific ambiguity introduce additional difficulty; the benchmark measures multi-turn reasoning under controlled topology rather than ecological validity, and a pilot on real-world graphs remains future work. All sessions are conducted in English, so the extent to which the Autonomy Divergence persists across typologically different languages remains an open question. Finally, Execution Accuracy requires result-set identity including column names, which may penalize semantically equivalent queries that differ only in output formatting or column ordering; a finer-grained semantic equivalence metric that tolerates such surface variation would provide a more forgiving upper bound and remains future work.

\bibliography{custom}

\clearpage
\appendix
% ============================================================
% Appendix
% ============================================================

% -----------------------------------------------------------
% Appendix Table of Contents
% -----------------------------------------------------------
\section*{Appendix Contents}
\vspace{0.3em}
{\small
\newlength{\apptocindent}
\setlength{\apptocindent}{1.8em}
\begin{list}{}{\setlength{\leftmargin}{0pt}\setlength{\rightmargin}{0pt}%
  \setlength{\itemsep}{4pt}\setlength{\parsep}{0pt}\setlength{\topsep}{0pt}}

  \item \textbf{\hyperref[app:related]{A\quad Related Work}}
        \dotfill\ \pageref{app:related}

  \item \textbf{\hyperref[app:benchmarks]{B\quad Comparison Benchmarks}}
        \dotfill\ \pageref{app:benchmarks}

  \item \textbf{\hyperref[app:model_list]{C\quad Model Descriptions}}
        \dotfill\ \pageref{app:model_list}

  \item \textbf{\hyperref[app:benchmark_ref]{D\quad Benchmark Reference}}
        \dotfill\ \pageref{app:benchmark_ref}
  \begin{list}{}{\setlength{\leftmargin}{\apptocindent}\setlength{\rightmargin}{0pt}%
    \setlength{\itemsep}{1pt}\setlength{\parsep}{0pt}\setlength{\topsep}{3pt}}
    \item \hyperref[app:graph_schemas]{D.1\quad Knowledge Graph Schemas}
          \dotfill\ \pageref{app:graph_schemas}
    \item \hyperref[app:phenomena]{D.2\quad Conversational Phenomena}
          \dotfill\ \pageref{app:phenomena}
    \item \hyperref[app:chain_modes]{D.3\quad Chain Modes}
          \dotfill\ \pageref{app:chain_modes}
    \item \hyperref[app:personas]{D.4\quad Session Personas}
          \dotfill\ \pageref{app:personas}
    \item \hyperref[app:eval_protocols]{D.5\quad Evaluation Protocols}
          \dotfill\ \pageref{app:eval_protocols}
  \end{list}

  \item \textbf{\hyperref[app:detailed_results]{E\quad Detailed Experimental Results}}
        \dotfill\ \pageref{app:detailed_results}
  \begin{list}{}{\setlength{\leftmargin}{\apptocindent}\setlength{\rightmargin}{0pt}%
    \setlength{\itemsep}{1pt}\setlength{\parsep}{0pt}\setlength{\topsep}{3pt}}
    \item \hyperref[app:graph_ex]{E.1\quad Per-Graph Results}
          \dotfill\ \pageref{app:graph_ex}
    \item \hyperref[app:phenom_full]{E.2\quad Per-Phenomenon Results}
          \dotfill\ \pageref{app:phenom_full}
    \item \hyperref[app:chain_mode_results]{E.3\quad Per-Chain-Mode Results}
          \dotfill\ \pageref{app:chain_mode_results}
    \item \hyperref[app:persona_results]{E.4\quad Per-Persona Results}
          \dotfill\ \pageref{app:persona_results}
    \item \hyperref[app:retry]{E.5\quad Agentic Budget Scaling}
          \dotfill\ \pageref{app:retry}
  \end{list}

  \item \textbf{\hyperref[app:protocol_ablations]{F\quad Protocol Robustness}}
        \dotfill\ \pageref{app:protocol_ablations}
  \begin{list}{}{\setlength{\leftmargin}{\apptocindent}\setlength{\rightmargin}{0pt}%
    \setlength{\itemsep}{1pt}\setlength{\parsep}{0pt}\setlength{\topsep}{3pt}}
    \item \hyperref[app:three_variants]{F.1\quad Variant Design and Overall Effect}
          \dotfill\ \pageref{app:three_variants}
    \item \hyperref[app:gap_persistence]{F.2\quad Gap Persistence and Budget Cliff}
          \dotfill\ \pageref{app:gap_persistence}
  \end{list}

  \item \textbf{\hyperref[app:case_study]{G\quad Case Studies}}
        \dotfill\ \pageref{app:case_study}

  \item \textbf{\hyperref[app:prompts]{H\quad Evaluation Prompts}}
        \dotfill\ \pageref{app:prompts}

\end{list}
}
\vspace{1em}

% -----------------------------------------------------------
\section{Related Work}
\label{app:related}

\paragraph{Text-to-SQL benchmarks.}
The multi-turn and agentic evaluation paradigms that motivate \graphturn{} originate in the SQL community. SParC \citep{yu2019sparc} and CoSQL \citep{yu2019cosql} established conversational Text-to-SQL evaluation on relational databases, demonstrating that cross-turn context dependence materially increases difficulty; CHASE \citep{chase2021} extended this paradigm to Chinese with more challenging context-dependent splits. On the single-turn side, BIRD \citep{li2023bird} introduced large-scale databases with dirty values and external-knowledge grounding, while Spider~2.0 \citep{spider2_2025} further extended evaluation to real enterprise workflows across heterogeneous SQL dialects. Methodologically, DIN-SQL \citep{dinsql2023} showed that decomposing generation into schema linking and self-correction substantially improves accuracy, and CodeS \citep{codes2024} demonstrated that open-source models can rival proprietary LLMs with carefully curated training corpora. BIRD-INTERACT \citep{birdinteract2025} further showed that agentic evaluation reveals failure modes entirely invisible in oracle-context settings. These findings motivate our dual-protocol design, extended here to the property graph domain where no equivalent evaluation framework previously existed.

\paragraph{Text-to-Cypher and Graph Query Benchmarks.}
Cypher, the declarative query language for property graphs \citep{francis2018cypher}, has received comparatively little NLI attention. CypherBench \citep{hou2025cypherbench} is the most comprehensive existing Text-to-Cypher benchmark, covering 11 large-scale knowledge graphs derived from Wikidata with over 10,000 single-turn questions. ZOGRASCOPE \citep{zograscope2025} provides a human-annotated Cypher benchmark over a single domain graph with compositional and length-generalization splits. SM3-Text-to-Query \citep{sm3texttoquery2024} evaluates Cypher generation in the biomedical domain alongside SQL, MQL, and SPARQL. Text2GQL-Bench \citep{text2gqlbench2026} scales to 178K question--query pairs across 34 graph databases, supporting both Cypher and ISO-GQL. Earlier work also includes SpCQL \citep{spcql2022}, a Chinese dataset of 10,000 NL--Cypher pairs over a single Neo4j graph that established early baselines for the task; CySpider \citep{cyspider2023}, an English corpus of Cypher queries derived from Spider SQL via algorithmic translation; and Text2Cypher \citep{text2cypher2024}, a community-compiled English dataset of 44,387 (NL, Cypher) instances aggregated from public sources. To address the scarcity of Cypher training data, SyntheT2C \citep{synthet2c2025} and SynthCypher \citep{synthcypher2024} propose synthetic NL--Cypher generation pipelines that produce verified executable queries, though both retain a strictly single-turn formulation. Adjacent to property-graph Cypher, Spider4SPARQL \citep{spider4sparql2023} provides a compositional SPARQL benchmark over 166 RDF knowledge graphs, and SPINACH \citep{spinach2024} introduces an agentic SPARQL navigation benchmark for challenging real-world questions; both target RDF/SPARQL rather than Cypher and remain single-turn at the question level. All of these datasets share the limitation of providing no genuine conversational context or chain-dependent anaphora annotation. MTGQL \citep{mtgql2025} is the only prior multi-turn graph query benchmark, targeting nGQL (NebulaGraph) for financial market queries. Despite this recent progress, no existing benchmark combines multi-turn evaluation, contextual anaphora tracking, and an agentic protocol within the Cypher ecosystem.

\paragraph{Interactive and Agentic Evaluation.}
BIRD-INTERACT \citep{birdinteract2025} demonstrated that evaluating LLMs as interactive agents for Text-to-SQL reveals failure modes entirely invisible in oracle-context evaluation, and MINT \citep{wang2023mint} similarly shows that multi-step tool use introduces compounding errors absent in single-step benchmarks. A broader line of agentic benchmarks---AgentBench \citep{agentbench2024} across eight environments, $\tau$-bench \citep{taubench2025} for tool-agent-user interaction with simulated users, AppWorld \citep{appworld2024} for interactive coding over 9 apps and 457 APIs, OSWorld \citep{osworld2024} for real computer-use tasks, and SWE-bench \citep{swebench2024} for resolving real-world GitHub issues---has consistently surfaced a substantial gap between oracle and autonomous performance. These environments build on prompting paradigms such as ReAct \citep{react2023}, which interleaves reasoning and action, and self-correction methods such as Reflexion \citep{reflexion2023} and CRITIC \citep{critic2024}, which leverage verbal reinforcement and tool-interactive critique to recover from intermediate failures. Orthogonally, RULER \citep{ruler2024} shows that long-context models degrade sharply as input length grows, an effect directly relevant to multi-turn agentic settings where prediction history accumulates over a session. \graphturn{} departs from this prior work by targeting Cypher for property graphs, introducing chain-dependent turn dependencies specific to graph traversal, and providing CER---a quantitative metric for cross-turn error propagation that has no counterpart in prior work.

% -----------------------------------------------------------
\section{Comparison Benchmarks}
\label{app:benchmarks}

We briefly describe the five benchmarks compared against \graphturn{} in Table~\ref{tab:benchmark_comparison}.

\textbf{CypherBench}~\citep{hou2025cypherbench} is a Text-to-Cypher benchmark covering 11 large-scale multi-domain property graphs derived from Wikidata, comprising 7.8 million entities and over 10,000 questions. Addressing the challenge that raw RDF graphs exceed LLM context windows, CypherBench introduces property graph views queryable via Cypher, an RDF-to-property-graph conversion engine, and a task generation pipeline. All questions are single-turn; there is no conversational structure or agentic evaluation.

\textbf{ZOGRASCOPE}~\citep{zograscope2025} is a human-annotated Text-to-Cypher benchmark over a single-domain crime-investigation property graph (61.5K nodes, 105.8K edges, 11 entity classes) hosted on Neo4j. It provides approximately 5,022 samples (2,905 training, 767 IID test, and 1,350 compositional test) split to probe structural and length generalization. All questions are single-turn.

\textbf{SM3-Text-to-Query}~\citep{sm3texttoquery2024} is a NeurIPS 2024 multi-model medical benchmark comprising approximately 40,000 question--query pairs across four query languages (Cypher, SQL, MQL, SPARQL), built on synthetic patient data using the SNOMED-CT taxonomy. The Cypher subset covers queries over a Neo4j graph. All questions are single-turn; domain coverage is restricted to biomedical data.

\textbf{Text2GQL-Bench}~\citep{text2gqlbench2026} is the largest existing graph query benchmark at 178,184 (NL, Query) pairs across 34 graph databases spanning 13 domains, supporting both Cypher and ISO-GQL. Questions are available at three abstraction levels (Syntactic, Logical, and Business), in addition to the original question. All questions are single-turn; no multi-turn or agentic evaluation is provided.

\textbf{MTGQL}~\citep{mtgql2025} is, to our knowledge, the only existing multi-turn graph query language benchmark, targeting nGQL (NebulaGraph) for financial market graph queries. It uses an LLM-based automated construction pipeline and addresses session-level context dependencies. MTGQL does not use Cypher, does not include contextual anaphora annotation, and provides no agentic evaluation protocol.

% -----------------------------------------------------------
\section{Model Descriptions}
\label{app:model_list}

We evaluate fifteen models spanning three tiers on the full benchmark. QwQ-32B, listed last within the fine-tuned tier as the base model of STRuCT-LLM-Novo, is additionally evaluated on the stratified 210-session subset only; no separate citation is provided for it, since its results serve only as the base reference of the STRuCT-LLM-Novo pair. Citation information is provided in the bibliography for those models with published technical reports; for proprietary API checkpoints without one, the entry describes the release and access route.

\modeltier{Frontier Models.}

\textbf{Claude Opus 4.7} is Anthropic's flagship model, released in April 2026. It is designed for advanced coding, long-horizon agentic tasks, and complex multi-tool workflows, with a 1M token context window. Its parameter count is not publicly disclosed.

\textbf{GPT-5.5} is a frontier model from OpenAI released in April 2026, positioned as the successor to GPT-5.4 in the GPT-5 series. It features approximately 1 million token context window and is optimised for agentic coding, computer use, and knowledge work. No arXiv technical report has been published for this model.

\textbf{Kimi-K2.5}~\citep{kimi_k25_2025} is a Mixture-of-Experts (MoE) model from Moonshot AI with 1 trillion total parameters and 32 billion activated per token. Building on Kimi K2 (pre-trained on 15.5 trillion tokens with the MuonClip optimiser), K2.5 extends the family to a 256K token context with native multimodal (vision and video) capabilities. The model is designed for agentic tool use and multi-step reasoning, with open weights released under a modified MIT licence.

\textbf{Gemini-3.1-Flash-Lite} is a member of Google DeepMind's Gemini 3 family, optimised for high-throughput, cost-efficient, and low-latency applications. It accepts text, image, audio, and video inputs with a context window of up to 1 million tokens, and is positioned as the most cost-efficient model in the Gemini 3 generation.

\textbf{Qwen3-235B}~\citep{qwen3_2025} is the flagship Qwen3-235B-A22B MoE model from Alibaba Cloud's Qwen Team, with 235 billion total parameters and 22 billion activated per token, released in May 2025. It introduces a unified thinking/non-thinking mode with an adaptive thinking-budget mechanism, and supports 119 languages and dialects. The model is released under the Apache 2.0 licence.

\textbf{GLM-5}~\citep{glm2024chatglm} is the next-generation foundation model from Zhipu AI / Z.ai, released in February 2026. It is a Mixture-of-Experts model with 744 billion total parameters and approximately 40 billion activated per token, pre-trained on 28.5 trillion tokens. GLM-5 integrates DeepSeek Sparse Attention (DSA) to support a 200K token context window while reducing inference cost, and is designed for agentic engineering and long-horizon software development tasks. The weights are released under the MIT licence.

\textbf{DeepSeek-V3.2}~\citep{deepseekv3} is the latest model in the DeepSeek-V3 family, with 671 billion total parameters and 37 billion activated per token. Inheriting the architecture of DeepSeek-V3 (Multi-head Latent Attention, an auxiliary-loss-free load-balancing strategy, and multi-token prediction, pre-trained on 14.8 trillion tokens), V3.2 introduces DeepSeek Sparse Attention (DSA) through continued pre-training, achieving near-linear $O(kL)$ long-context attention cost while maintaining output quality comparable to the dense V3.1-Terminus baseline.

\textbf{ERNIE-5.0}~\citep{ernie2023} is a large-scale, natively omni-modal language model from Baidu, part of the ERNIE (Enhanced Representation through kNowledge IntEgration) series. ERNIE 5.0 trains text, image, audio, and video jointly under a unified autoregressive Mixture-of-Experts (MoE) architecture, with approximately 2.4 trillion total parameters and fewer than 3\% (roughly 72B) activated per inference, and a 128K context window.

\textbf{MiMo-V2.5-Pro}~\citep{mimo_v2_2026} is the flagship Mixture-of-Experts (MoE) model from Xiaomi, released in April 2026, with 1.02 trillion total parameters and 42 billion activated per token. It builds on the MiMo-V2-Flash architecture, employing a hybrid attention scheme that interleaves Sliding Window Attention with global attention (6:1 ratio, 128-token sliding window) together with a three-layer Multi-Token Prediction module, and supports a 1M token context window. The model unifies reasoning and multimodal understanding for demanding agentic, complex software-engineering, and long-horizon tasks, sustaining trajectories that span thousands of tool calls, and is released as open weights under the MIT licence.

\textbf{MiniMax-M2.7} is a Mixture-of-Experts model from MiniMax with 230 billion total parameters and approximately 10 billion activated per token (256 experts, 8 activated per token over 62 layers), supporting a 200K token context window. Released in March 2026, it is designed for coding, long-horizon agentic workflows, and office document editing, and is notable for a self-improving training loop in which an autonomous agent optimised parts of its own training pipeline. The weights are publicly available on Hugging Face under a modified MIT licence.

\modeltier{Small Open-Source Models.}

\textbf{Llama-3.1-8B}~\citep{llama31_2024} is a dense 8-billion-parameter open-weight model from Meta AI, released as part of the Llama 3.1 family. It supports multilingual text, code, reasoning, and tool use, with a 128K token context window, and is the smallest member of the Llama 3.1 series alongside the 70B and 405B variants.

\textbf{Gemma-2-9B}~\citep{gemma2_2024} is a dense 9-billion-parameter open model from Google DeepMind's Gemma Team. The 2B and 9B variants are trained via knowledge distillation rather than standard next-token prediction; the architecture uses interleaved local/global attention and grouped-query attention (GQA), and delivers performance competitive with models two to three times its size.

\modeltier{Fine-tuned / Specialized Models.}

\textbf{QwQ-32B} is a 32B open-weight reasoning model from Alibaba's Qwen team, trained with reinforcement learning for long chain-of-thought reasoning. It is not part of the fifteen-model full-benchmark suite; it is evaluated as the controlled base of STRuCT-LLM-Novo, the second specialisation pair in \S\ref{sec:specialization}, on the stratified 210-session subset of Appendix~\ref{app:protocol_ablations}.

\textbf{STRuCT-LLM-Novo}~\citep{struct_llm_novo_2025} is a model from Stoisser et al.\ at Novo Nordisk that unifies Text-to-SQL and Text-to-Cypher generation under a single reinforcement learning framework. It employs Chain-of-Thought supervision, a topology-aware reward based on graph edit distance, and cross-formalism transfer (SQL training improves Cypher generation and vice versa), using QwQ-32B as its base model.

\textbf{text-to-cypher-gemma} is a Gemma-2-9B variant fine-tuned by Neo4j on the \texttt{neo4j/text2cypher-2024v1} dataset of 44,387 examples. It is released on Hugging Face as a research and demonstration model; Neo4j describes it as not production-ready.

\textbf{CypherRI-7B}$^\dagger$~\citep{cypherri_2025} is a 7-billion-parameter Cypher-specialised model presented as a NeurIPS 2025 poster, which integrates schema selection into the Cypher-generation pipeline via reinforcement learning. In our evaluation, the model's reasoning output tokens contaminate all generated Cypher payloads, resulting in near-zero execution accuracy due to a format-compliance failure rather than generation ability; its results are excluded from aggregate statistics.

% -----------------------------------------------------------
\section{Benchmark Reference}
\label{app:benchmark_ref}

\subsection{Knowledge Graph Schemas}
\label{app:graph_schemas}

\begin{figure*}[t]
  \centering
  \includegraphics[width=\textwidth, height=0.92\textheight, keepaspectratio]{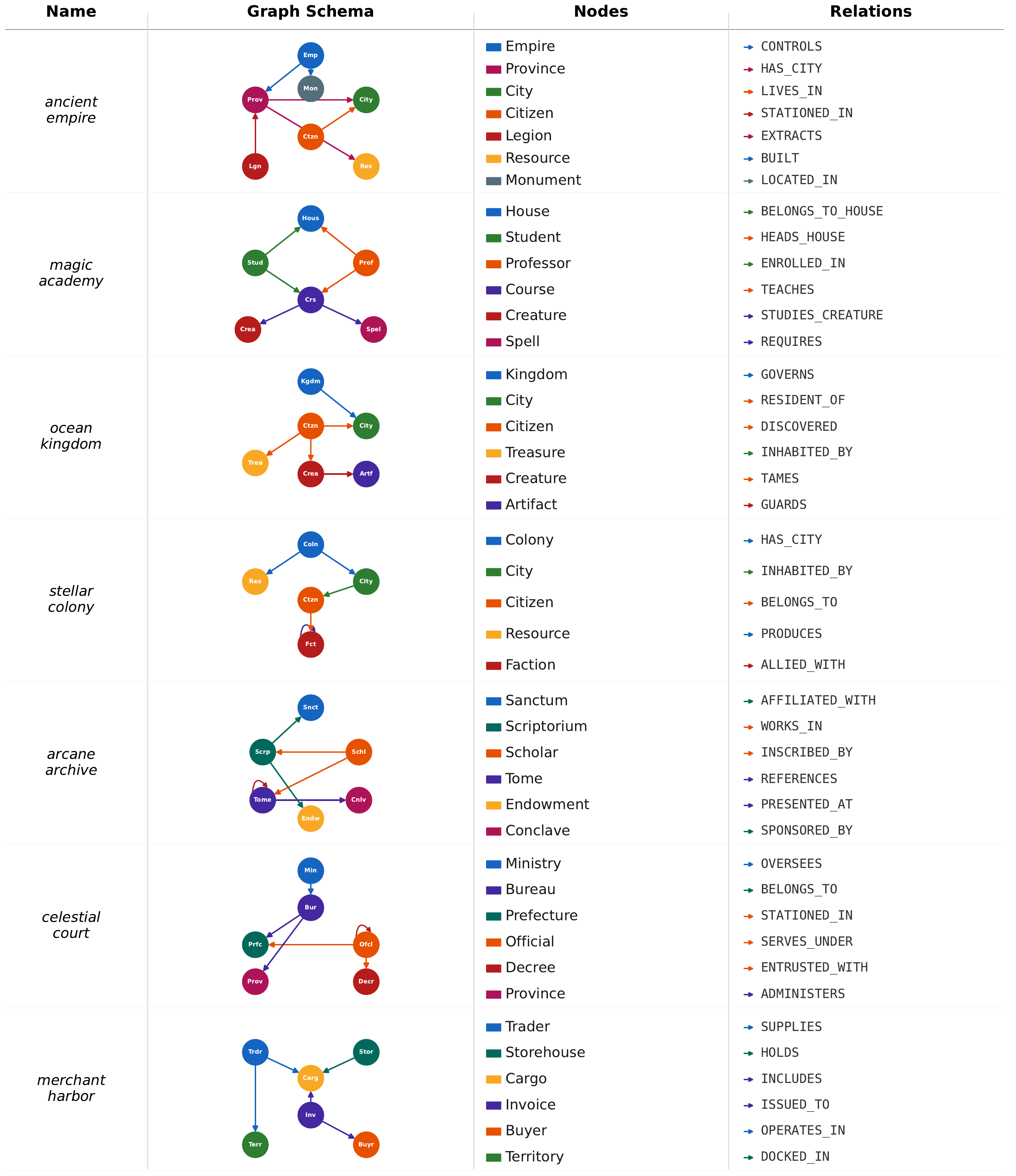}
  \caption{Schema overview of the seven \graphturn{} knowledge graphs. Each row: graph name, topology diagram, and node/relation listing. (a)--(d): directed cycle, bipartite cycle, diamond convergence, star+lateral; (e)--(g): citation self-loop, hierarchical self-loop, hub-and-spoke.}
  \label{fig:graph_topologies}
\end{figure*}

\graphturn{} uses seven purpose-built synthetic knowledge graphs, each constructed around a topological pattern that poses a distinct reasoning challenge. Figure~\ref{fig:graph_topologies} shows schema diagrams for all seven graphs.

% \begin{figure*}[t]
%   \centering
%   \includegraphics[width=\textwidth]{figures/appendix/fig_a2b_schema_details.pdf}
%   \caption{Detailed node and relation properties for all seven \graphturn{} knowledge graphs. \textbf{\textcolor[HTML]{1565C0}{STR}} = string, \textbf{\textcolor[HTML]{B83800}{INT}} = integer, \textbf{\textcolor[HTML]{1A7A30}{FLT}} = float, \textbf{\textcolor[HTML]{8B1050}{DATE}} = date.}
%   \label{fig:schema_details}
% \end{figure*}

\noindent\textbf{Ancient Empire} is a directed-cycle graph with 7 node types and 7 relation types. Edge properties on 5 relation types stress-test edge-property filtering. Deep 4-hop paths and dense numeric edge attributes make it a strong baseline for multi-hop traversal.

\noindent\textbf{Magic Academy} uses a bipartite-cycle topology with 6 node types. The dual Student--Professor path through both organisational units and courses simultaneously creates entity-resolution ambiguity, and Magic Academy ranks among the harder graphs for mid-tier models despite its regular bipartite structure.

\noindent\textbf{Ocean Kingdom} adopts a diamond-convergence structure: a central entity type fans out to three independent sub-graphs that partially reconverge. This creates path ambiguity and requires careful multi-hop planning to avoid retrieving duplicates.

\noindent\textbf{Stellar Colony} combines a tree-structured colony hierarchy with lateral \texttt{ALLIED\_WITH} edges between peer factions. It is the only graph with peer-to-peer lateral edges, breaking the assumption of strictly hierarchical traversal.

\noindent\textbf{Arcane Archive} embeds a \texttt{REFERENCES} citation self-loop on Tome nodes within an otherwise linear chain, enabling recursive multi-hop queries. It is among the highest-mean-EX graphs, though Celestial Court and Ancient Empire rank slightly higher.

\noindent\textbf{Celestial Court} models a bureaucratic hierarchy where officials supervise other officials at arbitrary depth via a \texttt{SERVES\_UNDER} self-referential edge. It uniquely exercises recursive organisational traversal.

\noindent\textbf{Merchant Harbor} uses a central invoice node that simultaneously bridges traders, buyers, and storehouses. It has the richest numeric edge properties of any graph in the benchmark and is the hardest new graph in the expanded dataset.

\subsection{Conversational Phenomena}
\label{app:phenomena}

\graphturn{} annotates every session turn with one of 13 conversational phenomena, grouped into four categories. Table~\ref{tab:phenomena_full} lists all 13 phenomena with their category, chain mode, chain-dependency status, a concise definition, and a canonical Cypher pattern.

\begin{table*}[t]
\centering
\small
\begin{tabular}{p{2.2cm} l p{1.8cm} c p{7.8cm}}
\toprule
\textbf{Phenomenon} & \textbf{Category} & \textbf{Chain Mode} & \textbf{Dep.} & \textbf{Definition and Cypher Pattern} \\
\midrule
\textbf{EXPAND} & Navigation & expand & \cmark &
  Traverse one hop outward from the prior result set; the prior result acts as the seed node set.
  \newline\texttt{MATCH (n)-[:R]->(m) WHERE n.name IN \$prev} \\[4pt]
\textbf{PIVOT} & Navigation & pivot & \cmark &
  Rank the prior result by a property, then traverse from the top-$k$ entities.
  \newline\texttt{WITH n ORDER BY n.prop DESC LIMIT k MATCH (n)-[:R]->(m)} \\[4pt]
\textbf{FIRST} & Navigation & pivot & \cmark &
  Reference the single top-ranked entity from the prior result and traverse from it.
  \newline\texttt{WITH n ORDER BY n.prop DESC LIMIT 1 MATCH (n)-[:R]->(m)} \\[4pt]
\textbf{TOPIC\_SHIFT} & Discourse & topic\_shift & \xmark &
  A context-break: the user poses an entirely new, independent query unrelated to prior results. The only non-chain-dependent phenomenon. \\[4pt]
\textbf{AGG\_AVG} & Aggregation & agg\_numeric & \cmark &
  Compute the mean of a numeric property over the prior result set.
  \newline\texttt{MATCH (n) WHERE n.name IN \$prev RETURN avg(n.prop)} \\[4pt]
\textbf{AGG\_MAX} & Aggregation & agg\_numeric & \cmark &
  Find the maximum value of a numeric property in the prior result set.
  \newline\texttt{MATCH (n) WHERE n.name IN \$prev RETURN max(n.prop)} \\[4pt]
\textbf{AGG\_SUM} & Aggregation & agg\_numeric & \cmark &
  Sum a numeric property across all entities in the prior result set.
  \newline\texttt{MATCH (n) WHERE n.name IN \$prev RETURN sum(n.prop)} \\[4pt]
\textbf{COUNT} & Aggregation & aggregate & \cmark &
  Count the cardinality of the prior result set.
  \newline\texttt{MATCH (n) WHERE n.name IN \$prev RETURN count(DISTINCT n)} \\[4pt]
\textbf{CONTRAST} & Discourse & contrast & \cmark &
  Query the complement of the prior result within the same entity type.
  \newline\texttt{MATCH (n:T) WHERE NOT n.name IN \$prev RETURN n.name} \\[4pt]
\textbf{VALUE\_FILTER} & Filtering & value\_narrow & \cmark &
  Filter the prior result by a numeric threshold on one property.
  \newline\texttt{MATCH (n) WHERE n.name IN \$prev AND n.prop > \$threshold} \\[4pt]
\textbf{MULTI\_COND} & Filtering & value\_narrow & \cmark &
  Apply two simultaneous value constraints (numeric threshold plus existence check) to narrow the prior result.
  \newline\texttt{WHERE n.prop > \$threshold AND n.prop2 IS NOT NULL} \\[4pt]
\textbf{REFINE} & Filtering & narrow & \cmark &
  Add a single new attribute constraint to narrow the prior result (property existence check).
  \newline\texttt{WHERE n.name IN \$prev AND n.prop IS NOT NULL} \\[4pt]
\textbf{FILTER} & Filtering & narrow & \cmark &
  Filter prior result to entities that possess a given property.
  \newline\texttt{WHERE n.name IN \$prev AND n.prop IS NOT NULL} \\
\bottomrule
\end{tabular}
\caption{The 13 conversational phenomena in \graphturn{}. \textbf{Dep.}: chain-dependent (\cmark) or independent (\xmark). Chain Mode = the session construction mode used to generate turns of this phenomenon.}
\label{tab:phenomena_full}
\end{table*}

\subsection{Chain Modes}
\label{app:chain_modes}

Each turn's chain dependency is characterised by a \textbf{chain mode} that describes how the current query relates to the prior result. Table~\ref{tab:chain_modes} enumerates all 9 chain modes with their definitions and a brief Cypher illustration.

\begin{table*}[t]
\centering
\small
\begin{tabular}{p{2.4cm} c p{10.6cm}}
\toprule
\textbf{Chain Mode} & \textbf{Dep.} & \textbf{Definition and Example} \\
\midrule
\textbf{None (T1)} & \xmark &
  First turn of a session; no prior result exists. The query is fully self-contained.
  \newline\textit{E.g.} ``List all provinces.'' $\to$ \texttt{MATCH (p:Province) RETURN p.name} \\[4pt]
\textbf{topic\_shift} & \xmark &
  A deliberate context-break mid-session. The user abandons the prior thread and opens a new independent query.
  \newline\textit{E.g.} (after querying officials) ``Actually, list all trade routes.'' \\[4pt]
\textbf{expand} & \cmark &
  Traverse one hop outward from the prior result. The prior entity set is used directly as a seed.
  \newline\textit{E.g.} (after T1 provinces) ``What cities are in those provinces?'' \\[4pt]
\textbf{pivot} & \cmark &
  Rank the prior result by a property, select top-$k$, and traverse from those entities.
  \newline\textit{E.g.} ``Among those cities, which one has the highest population? Show its legions.'' \\[4pt]
\textbf{agg\_numeric} & \cmark &
  Compute a numeric aggregate (avg, max, sum) over a property of the prior result set.
  \newline\textit{E.g.} ``What is the average tribute gold of those provinces?'' \\[4pt]
\textbf{aggregate} & \cmark &
  Count the cardinality of the prior result set.
  \newline\textit{E.g.} ``How many legions were returned above?'' \\[4pt]
\textbf{contrast} & \cmark &
  Query the logical complement of the prior result within the same entity type.
  \newline\textit{E.g.} ``Now show me all provinces \emph{not} in that list.'' \\[4pt]
\textbf{value\_narrow} & \cmark &
  Filter the prior result by value constraints on properties, either a single numeric threshold or a combination of a threshold with an existence check.
  \newline\textit{E.g.} ``Which of those have a tribute gold greater than 5000?'' \\[4pt]
\textbf{narrow} & \cmark &
  Add one or two attribute constraints to narrow the prior result (categorical or existence check).
  \newline\textit{E.g.} ``Among those officials, show only those whose rank is `Commander'.'' \\
\bottomrule
\end{tabular}
\caption{The 9 chain modes in \graphturn{}. \textbf{Dep.}: chain-dependent (\cmark) or independent (\xmark). \textit{None (T1)} and \textit{topic\_shift} are the two non-chain-dependent modes; all others require the prior result to formulate the current query.}
\label{tab:chain_modes}
\end{table*}

\subsection{Session Personas}
\label{app:personas}

Each of the 721 sessions is assigned one of 10 personas drawn uniformly at random. The persona governs utterance register throughout the session. Table~\ref{tab:personas} defines all 10 persona types and gives a representative utterance example for each.

\begin{table*}[t]
\centering
\small
\begin{tabular}{p{2.8cm} p{5.8cm} p{6.4cm}}
\toprule
\textbf{Persona} & \textbf{Definition} & \textbf{Example Utterance} \\
\midrule
\textbf{Journalist} &
  Investigative, direct phrasing; seeks evidence and facts; often frames queries as questions with implicit causal intent. &
  ``Which provinces experienced the sharpest decline in tribute payments last year?'' \\[4pt]
\textbf{Student} &
  Exploratory and learning-oriented; may ask elementary questions or seek clarification on results. &
  ``I'm trying to understand the hierarchy — can you show me which officials answer to the Emperor?'' \\[4pt]
\textbf{Product Manager} &
  Goal-focused; wants summary statistics and ranked lists; minimal tolerance for schema details. &
  ``Give me the top three trade routes by cargo volume.'' \\[4pt]
\textbf{Novice User} &
  Vague phrasing; often uses pronouns ambiguously (``those things'', ``that one'') without clear referents. &
  ``Now what about the other ones? The ones with the big numbers?'' \\[4pt]
\textbf{Data Analyst} &
  Precise and technical; requests specific columns, sorted output, and numeric breakdowns. &
  ``Return the entity ID, name, and tribute\_gold for each province, ordered descending.'' \\[4pt]
\textbf{Casual User} &
  Informal register; contractions, colloquialisms, and incomplete sentences. May mix social and query intent. &
  ``ok cool — so like, how many of those are there actually?'' \\[4pt]
\textbf{Impatient Executive} &
  Terse and demanding; short imperative sentences; expects single-number or short-list answers. &
  ``Total revenue. Now.'' \\[4pt]
\textbf{Adversarial User} &
  Deliberately challenges the system; uses skeptical framing, mock demands, or claims the system will fail. &
  ``I bet you can't tell me which of those provinces have more than two legions stationed there.'' \\[4pt]
\textbf{Non-Native Speaker} &
  Non-standard grammar, simplified vocabulary, and occasional word-order errors. Meaning is recoverable. &
  ``Please showing me the official who is highest in rank from previous list?'' \\[4pt]
\textbf{Verbose User} &
  Over-specified utterances with redundant context, restated background, and multi-clause formulations. &
  ``Taking into account all of the provinces we identified earlier in our analysis — specifically those that were shown in the results of the query we ran at the start of our session — could you now compute for me the total sum of their tribute gold values?'' \\
\bottomrule
\end{tabular}
\caption{The 10 session personas in \graphturn{}. Each session is assigned one persona uniformly at random (66--77 sessions per type across 721 sessions total). Persona governs utterance register for every turn in the session.}
\label{tab:personas}
\end{table*}

The pooled inter-annotator agreement of $\kappa = 0.83$ reported in \S\ref{sec:annotation} aggregates three annotation criteria. Table~\ref{tab:kappa_decomp} gives a retrospective decomposition of it by criterion, derived from the annotation protocol structure rather than independent re-measurement of the two-annotator records. Agreement is near-ceiling for intent capture, because the gold Cypher is programmatically generated and database-verified. It remains high for cross-turn anaphora, anchored by the machine-generated dependency structure. It is lowest for utterance naturalness with respect to the assigned persona, which is the most subjective criterion. The pooled 0.83 falls between the weakest and strongest criteria, consistent with disagreement concentrated on persona naturalness. This decomposition supports the positioning of persona in \S\ref{sec:personas} as a register-diversification device rather than a validated experimental factor.

\begin{table}[t]
\centering
\small
\begin{tabular}{l c}
\toprule
\textbf{Criterion} & \textbf{$\kappa$} \\
\midrule
Gold Cypher captures intent              & 0.91 \\
Cross-turn anaphora unambiguous          & 0.86 \\
Utterance natural for assigned persona   & 0.64 \\
\midrule
Pooled (reported)                        & 0.83 \\
\bottomrule
\end{tabular}
\caption{Inter-annotator agreement $\kappa$ decomposed by criterion. Persona naturalness is the weakest criterion. Criterion-level values are a retrospective decomposition based on the annotation protocol structure; the pooled $\kappa = 0.83$ is measured.}
\label{tab:kappa_decomp}
\end{table}

% -----------------------------------------------------------
\subsection{Evaluation Protocols}
\label{app:eval_protocols}

\paragraph{Action space specification.}
Both protocols share an identical five-action space. Table~\ref{tab:action_space} summarises the input and return value of each action; the paragraphs below describe the implementation constraints in detail.

\begin{table}[t]
\centering
\footnotesize
\setlength{\tabcolsep}{4pt}
\begin{tabular}{p{2.2cm} p{2.1cm} p{2.3cm}}
\toprule
\textbf{Action} & \textbf{Input} & \textbf{Returns} \\
\midrule
\texttt{EXECUTE\_CYPHER}  & Cypher query string & Result rows; error on failure \\[3pt]
\texttt{INSPECT\_SCHEMA}  & Empty (ignored)     & Node labels, properties, relation types \\[3pt]
\texttt{SEARCH\_VALUES}   & JSON: label, property, query & Up to 10 matching values \\[3pt]
\texttt{ASK\_USER}        & Natural language question & String reply from user \\[3pt]
\texttt{SUBMIT\_ANSWER}   & Final Cypher query  & \texttt{"Answer submitted"} \\
\bottomrule
\end{tabular}
\caption{The five-action space shared by both evaluation protocols. Each action costs 1 budget unit.}
\label{tab:action_space}
\end{table}

\texttt{EXECUTE\_CYPHER} runs the provided Cypher against the live Neo4j instance and returns the result rows, or an error message if the query is malformed or times out. A hard timeout of 120\,s is enforced. To keep prompts tractable, current-turn results are truncated to 20 rows and prior-turn results stored in the guided history are truncated to 10 rows. The one exception is \textsc{Contrast} turns: the immediately preceding turn's result is passed without truncation so the model can compute the complement set.

\texttt{INSPECT\_SCHEMA} returns the full graph schema — all node labels with their typed properties and all relation types with source and target labels. The action always succeeds and costs 1 budget unit. Under the guided protocol the schema is already injected into every turn prompt, making this action redundant but still valid. Under the agentic protocol the schema is withheld from the system prompt and must be actively discovered via this action.

\texttt{SEARCH\_VALUES} resolves entity-linking mismatches between user phrasing and stored property values. It accepts a JSON payload with keys \texttt{label}, \texttt{property}, and \texttt{query}, and applies a cascading fuzzy search: first a case-insensitive \texttt{CONTAINS} match, then a \texttt{STARTS WITH} prefix match, and finally a per-word \texttt{CONTAINS} match for multi-word queries. Up to 10 matching values are returned. A parse error is returned on malformed JSON input.

\texttt{ASK\_USER} poses a clarifying question to the user and returns a string reply. It does \emph{not} terminate the action loop; the model may continue issuing further actions after receiving the reply. Under the guided protocol, the reply is scripted: turns annotated with \texttt{ambiguity\_info} return the pre-authored disambiguation answer, while all other turns return the fixed discouragement string \textit{``I think my question was clear enough. Please try to answer it directly.''} Under the agentic protocol, the reply is generated by a dedicated LLM-based user simulator described in the paragraph below.

\texttt{SUBMIT\_ANSWER} finalises the predicted query for the current turn and immediately terminates the action loop; only the first call counts. If the budget is exhausted before \texttt{SUBMIT\_ANSWER} is called, the runner falls back to the payload of the last \texttt{EXECUTE\_CYPHER} action in the current turn. If no \texttt{EXECUTE\_CYPHER} was issued either, the predicted query is set to \texttt{None}, scoring EX\,=\,0.

\paragraph{Guided protocol mechanics.}
At the start of each turn, the runner constructs a gold conversation history by formatting all prior turns as \texttt{User: <utterance>} / \texttt{Assistant: [Query Result] <gold result, up to 10 rows>} pairs. The full graph schema is injected directly into the turn prompt alongside this history. The model then enters an action loop capped at exactly 3 steps; the cap is a hard range and cannot be extended. For \textsc{Contrast} turns, the immediately preceding turn's gold result is passed without row truncation so the model can compute the complement set. For \textsc{First} turns that reference a specific earlier result, a dependency hint is appended to the user utterance: \textit{``[Note: This turn builds upon the result of Turn N. Use the result shown for Turn N in the conversation history above.]''} When \texttt{ASK\_USER} is called, the system returns a scripted reply rather than invoking a live simulator: turns annotated with \texttt{ambiguity\_info} return the pre-authored clarification; all other turns return the fixed string \textit{``I think my question was clear enough. Please try to answer it directly.''} This makes \texttt{ASK\_USER} practically useless in the guided setting, as it consumes one of three budget actions for minimal informational gain.

\paragraph{Agentic protocol mechanics.}
The total action budget is computed as $B = T \times m$, where $T$ is the number of turns in the session and $m \in \{3, 5, 10\}$ is the budget multiplier. This budget is shared across the entire session and decrements with every action regardless of turn. The graph schema is withheld from the system prompt; the model must call \texttt{INSPECT\_SCHEMA} to discover it. After each turn, the runner builds an interaction history from all completed turns, recording \texttt{EXECUTE\_CYPHER} payloads and results (truncated to 200 characters), \texttt{ASK\_USER} exchanges, and \texttt{SUBMIT\_ANSWER} payloads; \texttt{INSPECT\_SCHEMA} and \texttt{SEARCH\_VALUES} actions are omitted from the history to reduce context length. If the shared budget reaches zero mid-session, all remaining turns receive \texttt{pred\_cypher = None} and score EX\,=\,0 — a hard cliff rather than graceful degradation. This design choice means that models which over-invest budget in early turns (e.g., repeated \texttt{EXECUTE\_CYPHER} retries without schema recovery) face compounding penalties in later turns.

\paragraph{Agentic user simulator.}
The \texttt{ASK\_USER} action in the agentic protocol invokes a dedicated LLM-based user simulator operating under a two-phase anti-leakage pipeline. In the first phase, the simulator classifies the model's question into one of three intents: (1) \textbf{Reject} — if the question probes schema or technical details (e.g., contains keywords such as \textit{schema}, \textit{label}, \textit{relationship}, or \textit{cypher}), the simulator returns a canned rejection from a rotating pool of four templates (e.g., ``I don't know about the database schema. I just want to find the information I asked about.''); (2) \textbf{Disambiguate} — if the turn has \texttt{ambiguity\_info} and the question matches a clarification-seeking pattern, the simulator returns the pre-authored disambiguation reply; (3) \textbf{General} — otherwise, the simulator generates a reply via an LLM call at temperature 0.5. In the second phase, the generated reply is scanned for schema terms (node labels, relation types, property names); any reply containing such terms is replaced with a safe fallback. After three or more \texttt{ASK\_USER} calls within a single turn, an impatience signal is appended to the simulator context. This design prevents the model from extracting schema information through conversational probing while still allowing genuine disambiguation.

\paragraph{Protocol variants for the sensitivity ablation.}
Three variants of the agentic protocol are used in the sensitivity ablation reported in Appendix~\ref{app:protocol_ablations}. Each modifies exactly one dimension of the base protocol and leaves the rest unchanged, so that any change in the ranking can be attributed to the perturbed dimension. The \textbf{Agentic-PS} variant makes the graph schema persistent: after the model calls \texttt{INSPECT\_SCHEMA} for the first time, the full schema returned by that call is injected into the context of every subsequent turn for the remainder of the session, and further \texttt{INSPECT\_SCHEMA} calls become redundant but still cost one budget unit. This variant tests whether the strong correlation between \texttt{INSPECT\_SCHEMA} allocation and agentic EX reflects a causal role for repeated schema inspection. The \textbf{Disclosed-Horizon} variant removes the unknown-horizon confound by disclosing the session length $T$, the per-turn action target $m$, and a suggested remaining budget at the start of each turn, so the model can pace its spending across the full session rather than rationing against an unknown endpoint. The \textbf{Budget-Encouraging Prompt} variant replaces the base efficiency instruction, which asks the model to prefer submitting efficiently, with explicit encouragement to use the full available budget for verification and value retrieval, and is run at tenfold budget to give the encouragement room to act. This variant tests whether the self-limitation observed in \S\ref{sec:budget_ceiling} stems from the efficiency wording rather than a capability gap.

% -----------------------------------------------------------
\section{Detailed Experimental Results}
\label{app:detailed_results}

\subsection{Per-Graph Results}
\label{app:graph_ex}

Table~\ref{tab:graph_ex} reports guided EX by knowledge graph for all 15 models. Stellar Colony (0.426) and Merchant Harbor (0.434) are the hardest graphs, reflecting the difficulty of lateral alliance edges and multi-party invoice structures respectively. Celestial Court is the easiest (0.485), benefiting from regular hierarchical traversal patterns under oracle context. For small models, per-model maxima often fall on Magic Academy, suggesting that bipartite paths become relatively easier once multi-condition and complement queries that dominate harder graphs are systematically failed.

Table~\ref{tab:graph_agentic3_ex} reports the corresponding agentic $\times$3 results. The graph-level ranking is broadly preserved under the agentic protocol, but absolute EX values drop sharply: Arcane Archive shows the largest absolute drop, from 0.469 (guided) to 0.250 (agentic $\times$3), while Celestial Court drops from 0.485 to 0.279.

\begin{table*}[t]
\small
\centering
\begin{tabular}{lcccccccc}
\toprule
\textbf{Model} & \textbf{Anc} & \textbf{Mag} & \textbf{Ocn} & \textbf{Stl} & \textbf{Arc} & \textbf{Cel} & \textbf{Mer} & \textbf{Avg} \\
\midrule
\rowcolor[HTML]{F2F2FE}
\multicolumn{9}{c}{\textit{Frontier Models}} \\
Claude Opus 4.7   & 0.674 & 0.594 & 0.660 & 0.603 & \textbf{0.683} & 0.663 & 0.650 & 0.647 \\
GPT-5.5           & 0.633 & 0.570 & 0.659 & 0.597 & \textbf{0.671} & 0.667 & 0.581 & 0.625 \\
Kimi-K2.5         & 0.620 & 0.568 & 0.588 & 0.551 & \textbf{0.667} & 0.618 & 0.605 & 0.602 \\
Gemini-3.1-Flash-Lite & 0.574 & \textbf{0.620} & 0.535 & 0.520 & 0.604 & 0.595 & 0.555 & 0.572 \\
Qwen3-235B        & 0.548 & 0.488 & 0.529 & 0.472 & 0.535 & \textbf{0.564} & 0.457 & 0.513 \\
GLM-5             & \textbf{0.588} & 0.439 & 0.485 & 0.468 & 0.538 & 0.560 & 0.468 & 0.506 \\
DeepSeek-V3.2     & 0.529 & 0.486 & 0.513 & 0.466 & 0.546 & \textbf{0.549} & 0.453 & 0.506 \\
ERNIE-5.0         & 0.459 & 0.508 & 0.493 & 0.441 & 0.430 & \textbf{0.517} & 0.431 & 0.468 \\
MiMo-V2.5-Pro     & \textbf{0.514} & 0.403 & 0.473 & 0.417 & 0.498 & 0.486 & 0.433 & 0.460 \\
MiniMax-M2.7      & 0.158 & 0.114 & 0.150 & 0.135 & 0.134 & \textbf{0.163} & 0.119 & 0.139 \\
\midrule
\rowcolor[HTML]{F2F2FE}
\multicolumn{9}{c}{\textit{Small Open-Source Models}} \\
Llama-3.1-8B      & 0.280 & \textbf{0.333} & 0.316 & 0.264 & 0.254 & 0.314 & 0.273 & 0.291 \\
Gemma-2-9B        & 0.255 & \textbf{0.305} & 0.286 & 0.289 & 0.259 & \textbf{0.305} & 0.266 & 0.281 \\
\midrule
\rowcolor[HTML]{F2F2FE}
\multicolumn{9}{c}{\textit{Fine-tuned / Specialized Models}} \\
STRuCT-LLM-Novo   & \textbf{0.549} & 0.528 & 0.542 & 0.484 & 0.537 & 0.526 & 0.515 & 0.526 \\
text-to-cypher-gemma & 0.254 & 0.282 & \textbf{0.294} & 0.257 & 0.205 & 0.265 & 0.272 & 0.261 \\
CypherRI-7B$^\dagger$ & \textbf{0.019} & 0.009 & 0.013 & 0.015 & 0.009 & 0.008 & 0.014 & 0.013 \\
\midrule
\rowcolor[HTML]{F2F2FE}
Mean (14) & \textbf{0.474} & 0.446 & 0.466 & 0.426 & 0.469 & \textbf{0.485} & 0.434 & --- \\
\bottomrule
\end{tabular}
\caption{Guided EX by knowledge graph. Anc=Ancient Empire, Mag=Magic Academy, Ocn=Ocean Kingdom, Stl=Stellar Colony, Arc=Arcane Archive, Cel=Celestial Court, Mer=Merchant Harbor. \textbf{Bold}: per-row maximum. $\dagger$ excluded from mean row.}
\label{tab:graph_ex}
\end{table*}

\begin{table*}[t]
\small
\centering
\begin{tabular}{lcccccccc}
\toprule
\textbf{Model} & \textbf{Anc} & \textbf{Mag} & \textbf{Ocn} & \textbf{Stl} & \textbf{Arc} & \textbf{Cel} & \textbf{Mer} & \textbf{Avg} \\
\midrule
\rowcolor[HTML]{F2F2FE}
\multicolumn{9}{c}{\textit{Frontier Models}} \\
Claude Opus 4.7   & 0.427 & 0.376 & 0.404 & 0.358 & 0.410 & \textbf{0.454} & 0.384 & 0.402 \\
GPT-5.5           & 0.433 & 0.323 & \textbf{0.494} & 0.389 & 0.416 & 0.415 & 0.340 & 0.401 \\
Kimi-K2.5         & 0.363 & 0.275 & 0.339 & 0.278 & 0.373 & \textbf{0.411} & 0.311 & 0.335 \\
Gemini-3.1-Flash-Lite & 0.445 & \textbf{0.479} & 0.442 & 0.392 & 0.409 & 0.446 & 0.409 & 0.432 \\
Qwen3-235B        & 0.221 & 0.167 & 0.206 & 0.166 & 0.184 & \textbf{0.248} & 0.204 & 0.199 \\
GLM-5             & \textbf{0.358} & 0.229 & 0.278 & 0.228 & 0.268 & 0.338 & 0.221 & 0.274 \\
DeepSeek-V3.2     & 0.237 & 0.162 & 0.229 & 0.164 & 0.195 & \textbf{0.239} & 0.201 & 0.204 \\
ERNIE-5.0         & 0.278 & 0.313 & 0.309 & 0.271 & 0.270 & 0.311 & \textbf{0.317} & 0.295 \\
MiMo-V2.5-Pro     & \textbf{0.311} & 0.205 & 0.276 & 0.212 & 0.271 & 0.288 & 0.190 & 0.250 \\
MiniMax-M2.7      & 0.172 & 0.130 & 0.166 & 0.142 & 0.145 & \textbf{0.208} & 0.132 & 0.156 \\
\midrule
\rowcolor[HTML]{F2F2FE}
\multicolumn{9}{c}{\textit{Small Open-Source Models}} \\
Llama-3.1-8B      & 0.141 & \textbf{0.194} & 0.157 & 0.166 & 0.132 & 0.001 & 0.164 & 0.137 \\
Gemma-2-9B        & 0.146 & \textbf{0.194} & 0.120 & 0.179 & 0.057 & 0.074 & 0.179 & 0.136 \\
\midrule
\rowcolor[HTML]{F2F2FE}
\multicolumn{9}{c}{\textit{Fine-tuned / Specialized Models}} \\
STRuCT-LLM-Novo   & 0.350 & 0.317 & \textbf{0.366} & 0.278 & 0.346 & 0.364 & 0.312 & 0.333 \\
text-to-cypher-gemma & 0.164 & 0.192 & 0.155 & \textbf{0.201} & 0.025 & 0.106 & 0.159 & 0.143 \\
CypherRI-7B$^\dagger$ & 0.001 & 0.001 & 0.001 & 0.001 & 0.001 & 0.001 & 0.001 & 0.001 \\
\midrule
\rowcolor[HTML]{F2F2FE}
Mean (14) & \textbf{0.289} & 0.254 & 0.281 & 0.245 & 0.250 & 0.279 & 0.252 & --- \\
\bottomrule
\end{tabular}
\caption{Agentic $\times$3 EX by knowledge graph. Same layout as Table~\ref{tab:graph_ex}. \textbf{Bold}: per-row maximum. $\dagger$ excluded from mean row.}
\label{tab:graph_agentic3_ex}
\end{table*}

\subsection{Per-Phenomenon Results}
\label{app:phenom_full}

Table~\ref{tab:phenomena_all_models} reports per-phenomenon guided EX for all 14 models (CypherRI-7B excluded). REFINE (mean 0.244) is the hardest phenomenon universally: even Claude achieves only 0.409. CONTRAST has the highest inter-model variance (0.021--0.540). Aggregation phenomena (AGG\_MAX, AGG\_AVG, AGG\_SUM) form the easiest cluster across all models, while Filtering and Navigation phenomena show the widest within-category spread.

\begin{table*}[t]
\centering
\resizebox{\textwidth}{!}{%
\begin{tabular}{llccccccccccccccc}
\toprule
\textbf{Phenom.} & \textbf{Cat.} & \textbf{Cl} & \textbf{GPT} & \textbf{Km} & \textbf{Gfl} & \textbf{ST} & \textbf{Q3} & \textbf{GL} & \textbf{DS} & \textbf{ER} & \textbf{Mi} & \textbf{Mx} & \textbf{Ll} & \textbf{G2} & \textbf{t2} & \textbf{Mean} \\
\midrule
AGG\_MAX      & Agg  & \textbf{0.842} & 0.813 & 0.788 & 0.741 & 0.796 & 0.798 & 0.840 & 0.793 & 0.603 & 0.810 & 0.310 & 0.360 & 0.409 & 0.197 & 0.650 \\
AGG\_AVG      & Agg  & 0.767 & 0.679 & 0.740 & 0.677 & 0.740 & 0.722 & \textbf{0.780} & 0.700 & 0.578 & 0.702 & 0.285 & 0.336 & 0.422 & 0.298 & 0.602 \\
TOPIC\_SHIFT  & Disc & 0.821 & \textbf{0.849} & 0.760 & 0.782 & 0.680 & 0.643 & 0.487 & 0.656 & 0.778 & 0.422 & 0.111 & 0.788 & 0.538 & 0.673 & 0.642 \\
AGG\_SUM      & Agg  & 0.642 & 0.663 & 0.661 & 0.655 & 0.665 & 0.692 & \textbf{0.696} & 0.676 & 0.547 & 0.661 & 0.281 & 0.331 & 0.397 & 0.318 & 0.563 \\
VALUE\_FILTER & Filt & 0.725 & 0.753 & \textbf{0.764} & 0.590 & 0.657 & 0.691 & 0.562 & 0.702 & 0.590 & 0.528 & 0.112 & 0.270 & 0.354 & 0.382 & 0.549 \\
COUNT         & Agg  & 0.670 & 0.614 & 0.636 & 0.583 & 0.514 & 0.657 & \textbf{0.673} & 0.604 & 0.520 & 0.611 & 0.274 & 0.156 & 0.352 & 0.240 & 0.507 \\
FIRST         & Nav  & \textbf{0.614} & 0.560 & 0.481 & 0.519 & 0.376 & 0.506 & 0.471 & 0.501 & 0.361 & 0.412 & 0.065 & 0.075 & 0.100 & 0.115 & 0.368 \\
PIVOT         & Nav  & \textbf{0.607} & 0.548 & 0.499 & 0.507 & 0.426 & 0.496 & 0.442 & 0.486 & 0.309 & 0.419 & 0.071 & 0.091 & 0.088 & 0.091 & 0.363 \\
EXPAND        & Nav  & 0.543 & \textbf{0.626} & 0.548 & 0.419 & 0.520 & 0.363 & 0.442 & 0.355 & 0.351 & 0.486 & 0.140 & 0.116 & 0.098 & 0.151 & 0.368 \\
FILTER        & Filt & \textbf{0.670} & 0.614 & 0.602 & 0.216 & 0.591 & 0.511 & 0.295 & 0.489 & 0.284 & 0.227 & 0.011 & 0.227 & 0.341 & 0.205 & 0.377 \\
MULTI\_COND   & Filt & \textbf{0.582} & 0.566 & 0.551 & 0.367 & 0.495 & 0.327 & 0.332 & 0.316 & 0.291 & 0.250 & 0.036 & 0.122 & 0.311 & 0.235 & 0.341 \\
\rowcolor{red!10}
CONTRAST      & Disc & \textbf{0.540} & 0.515 & 0.460 & 0.384 & 0.245 & 0.236 & 0.481 & 0.228 & 0.173 & 0.418 & 0.072 & 0.030 & 0.025 & 0.021 & 0.273 \\
\rowcolor{red!10}
REFINE        & Filt & \textbf{0.409} & 0.375 & 0.375 & 0.170 & 0.364 & 0.307 & 0.205 & 0.330 & 0.216 & 0.159 & 0.011 & 0.148 & 0.205 & 0.148 & 0.244 \\
\midrule
\rowcolor[HTML]{F2F2FE}
\textbf{Overall} & --- & 0.647 & 0.625 & 0.602 & 0.572 & 0.526 & 0.513 & 0.506 & 0.506 & 0.468 & 0.460 & 0.139 & 0.291 & 0.281 & 0.261 & 0.457 \\
\bottomrule
\end{tabular}%
}
\caption{Guided EX by phenomenon for all 14 models (CypherRI excluded). Cl=Claude 4.7, GPT=GPT-5.5, Km=Kimi-K2.5, Gfl=Gemini-3.1-FL, ST=STRuCT-LLM, Q3=Qwen3-235B, GL=GLM-5, DS=DeepSeek-V3.2, ER=ERNIE-5.0, Mi=MiMo-V2.5-Pro, Mx=MiniMax-M2.7, Ll=Llama-3.1-8B, G2=Gemma-2-9B, t2=t2c-gemma. \colorbox{red!10}{Red rows}: mean EX below 0.30. \textbf{Bold}: best per row.}
\label{tab:phenomena_all_models}
\end{table*}

\begin{table*}[t]
\centering
\resizebox{\textwidth}{!}{%
\begin{tabular}{llccccccccccccccc}
\toprule
\textbf{Phenom.} & \textbf{Cat.} & \textbf{Cl} & \textbf{GPT} & \textbf{Km} & \textbf{Gfl} & \textbf{ST} & \textbf{Q3} & \textbf{GL} & \textbf{DS} & \textbf{ER} & \textbf{Mi} & \textbf{Mx} & \textbf{Ll} & \textbf{G2} & \textbf{t2} & \textbf{Mean} \\
\midrule
AGG\_MAX      & Agg  & \textbf{0.663} & 0.611 & 0.599 & 0.579 & 0.549 & 0.505 & 0.579 & 0.510 & 0.443 & 0.601 & 0.480 & 0.195 & 0.187 & 0.121 & 0.473 \\
AGG\_AVG      & Agg  & \textbf{0.576} & 0.500 & 0.507 & 0.455 & 0.471 & 0.374 & 0.516 & 0.381 & 0.316 & 0.527 & 0.399 & 0.141 & 0.155 & 0.130 & 0.389 \\
TOPIC\_SHIFT  & Disc & 0.533 & 0.644 & 0.433 & \textbf{0.739} & 0.446 & 0.225 & 0.289 & 0.233 & 0.563 & 0.249 & 0.222 & 0.178 & 0.200 & 0.313 & 0.376 \\
AGG\_SUM      & Agg  & \textbf{0.524} & 0.501 & 0.457 & 0.418 & 0.414 & 0.345 & 0.426 & 0.343 & 0.310 & 0.443 & 0.331 & 0.121 & 0.119 & 0.094 & 0.346 \\
VALUE\_FILTER & Filt & 0.399 & \textbf{0.416} & 0.225 & 0.337 & 0.287 & 0.101 & 0.169 & 0.112 & 0.292 & 0.112 & 0.039 & 0.090 & 0.135 & 0.197 & 0.208 \\
COUNT         & Agg  & \textbf{0.548} & 0.455 & 0.308 & 0.246 & 0.333 & 0.109 & 0.402 & 0.100 & 0.062 & 0.324 & 0.146 & 0.000 & 0.003 & 0.000 & 0.217 \\
FIRST         & Nav  & 0.317 & 0.243 & 0.293 & \textbf{0.412} & 0.269 & 0.151 & 0.228 & 0.147 & 0.218 & 0.201 & 0.038 & 0.062 & 0.062 & 0.059 & 0.193 \\
PIVOT         & Nav  & 0.325 & 0.249 & 0.295 & \textbf{0.436} & 0.249 & 0.142 & 0.226 & 0.151 & 0.201 & 0.205 & 0.036 & 0.047 & 0.047 & 0.047 & 0.190 \\
EXPAND        & Nav  & 0.200 & \textbf{0.349} & 0.210 & 0.244 & 0.269 & 0.082 & 0.129 & 0.087 & 0.107 & 0.097 & 0.044 & 0.047 & 0.044 & 0.044 & 0.140 \\
FILTER        & Filt & 0.261 & \textbf{0.284} & 0.239 & \textbf{0.284} & 0.193 & 0.023 & 0.136 & 0.034 & 0.239 & 0.080 & 0.023 & 0.136 & 0.125 & 0.193 & 0.161 \\
MULTI\_COND   & Filt & \textbf{0.260} & 0.209 & 0.173 & 0.194 & 0.184 & 0.051 & 0.066 & 0.061 & 0.214 & 0.041 & 0.020 & 0.077 & 0.102 & 0.122 & 0.127 \\
\rowcolor{red!10}
CONTRAST      & Disc & 0.186 & \textbf{0.207} & 0.105 & 0.152 & 0.089 & 0.046 & 0.034 & 0.063 & 0.080 & 0.038 & 0.017 & 0.021 & 0.008 & 0.004 & 0.075 \\
\rowcolor{red!10}
REFINE        & Filt & 0.148 & 0.170 & 0.125 & \textbf{0.216} & 0.114 & 0.045 & 0.091 & 0.034 & 0.114 & 0.045 & 0.011 & 0.057 & 0.080 & 0.091 & 0.096 \\
\midrule
\rowcolor[HTML]{F2F2FE}
\textbf{Overall} & --- & 0.402 & 0.401 & 0.335 & 0.432 & 0.333 & 0.199 & 0.274 & 0.204 & 0.295 & 0.250 & 0.156 & 0.137 & 0.136 & 0.143 & 0.264 \\
\bottomrule
\end{tabular}%
}
\caption{Agentic $\times$3 EX by phenomenon for all 14 models (CypherRI excluded). Same abbreviations as Table~\ref{tab:phenomena_all_models}. \colorbox{red!10}{Red rows}: mean EX below 0.10. \textbf{Bold}: best per row.}
\label{tab:phenomena_agentic3}
\end{table*}

\subsection{Per-Chain-Mode Results}
\label{app:chain_mode_results}

Table~\ref{tab:chain_mode_ex} breaks down guided EX by chain mode. Among the two non-chain-dependent modes, \textit{topic\_shift} achieves by far the highest mean EX at 0.642, confirming that fresh independent queries are inherently easier, while \textit{None (T1)} sits at 0.359, comparable to the mid-tier chain-dependent modes: first turns are self-contained but require cold-start schema grounding. Among chain-dependent modes, \textit{agg\_numeric} and \textit{aggregate} are the easiest, while \textit{contrast} and \textit{narrow} (which underlies REFINE) are the hardest. The agentic $\times$3 column shows the absolute degradation from removing oracle context: \textit{aggregate} and \textit{value\_narrow} show the largest absolute drops (0.29 and 0.28 points), and \textit{contrast} the largest relative drop, retaining only 27\% of its guided EX, as context-resolution errors compound over long sessions.

\begin{table*}[t]
\centering
\resizebox{\textwidth}{!}{%
\begin{tabular}{lcccccccccccccccc}
\toprule
\textbf{Chain Mode} & \textbf{Dep.} & \textbf{Cl} & \textbf{GPT} & \textbf{Km} & \textbf{Gfl} & \textbf{ST} & \textbf{Q3} & \textbf{GL} & \textbf{DS} & \textbf{ER} & \textbf{Mi} & \textbf{Mx} & \textbf{Ll} & \textbf{G2} & \textbf{t2} & \textbf{Mean} \\
\midrule
None (T1)     & \xmark & 0.519 & 0.447 & 0.533 & \textbf{0.619} & 0.338 & 0.207 & 0.229 & 0.212 & 0.498 & 0.132 & 0.049 & 0.523 & 0.368 & 0.351 & 0.359 \\
topic\_shift  & \xmark & 0.821 & \textbf{0.849} & 0.760 & 0.782 & 0.680 & 0.643 & 0.487 & 0.656 & 0.778 & 0.422 & 0.111 & 0.788 & 0.538 & 0.673 & 0.642 \\
agg\_numeric  & \cmark & 0.753 & 0.716 & 0.730 & 0.690 & 0.733 & 0.737 & \textbf{0.771} & 0.723 & 0.576 & 0.724 & 0.292 & 0.342 & 0.409 & 0.271 & 0.603 \\
aggregate     & \cmark & 0.670 & 0.614 & 0.636 & 0.583 & 0.514 & 0.657 & \textbf{0.673} & 0.604 & 0.520 & 0.611 & 0.274 & 0.156 & 0.352 & 0.240 & 0.507 \\
value\_narrow & \cmark & 0.650 & \textbf{0.655} & 0.652 & 0.473 & 0.572 & 0.500 & 0.441 & 0.500 & 0.433 & 0.382 & 0.072 & 0.193 & 0.332 & 0.305 & 0.440 \\
pivot         & \cmark & \textbf{0.611} & 0.554 & 0.489 & 0.513 & 0.401 & 0.501 & 0.456 & 0.494 & 0.335 & 0.416 & 0.068 & 0.083 & 0.094 & 0.103 & 0.366 \\
expand        & \cmark & 0.543 & \textbf{0.626} & 0.548 & 0.419 & 0.520 & 0.363 & 0.442 & 0.355 & 0.351 & 0.486 & 0.140 & 0.116 & 0.098 & 0.151 & 0.368 \\
narrow        & \cmark & \textbf{0.540} & 0.494 & 0.489 & 0.193 & 0.477 & 0.409 & 0.250 & 0.409 & 0.250 & 0.193 & 0.011 & 0.188 & 0.273 & 0.176 & 0.311 \\
contrast      & \cmark & \textbf{0.540} & 0.515 & 0.460 & 0.384 & 0.245 & 0.236 & 0.481 & 0.228 & 0.173 & 0.418 & 0.072 & 0.030 & 0.025 & 0.021 & 0.273 \\
\midrule
\rowcolor[HTML]{F2F2FE}
\textbf{Overall} & --- & 0.647 & 0.625 & 0.602 & 0.572 & 0.526 & 0.513 & 0.506 & 0.506 & 0.468 & 0.460 & 0.139 & 0.291 & 0.281 & 0.261 & 0.457 \\
\bottomrule
\end{tabular}%
}
\caption{Guided EX by chain mode for all 14 models (CypherRI excluded). Same model abbreviations as Table~\ref{tab:phenomena_all_models}. \textbf{Bold}: best per row. \textbf{Dep.}: chain-dependent (\cmark) or independent (\xmark).}
\label{tab:chain_mode_ex}
\end{table*}

\begin{table*}[t]
\centering
\resizebox{\textwidth}{!}{%
\begin{tabular}{lcccccccccccccccc}
\toprule
\textbf{Chain Mode} & \textbf{Dep.} & \textbf{Cl} & \textbf{GPT} & \textbf{Km} & \textbf{Gfl} & \textbf{ST} & \textbf{Q3} & \textbf{GL} & \textbf{DS} & \textbf{ER} & \textbf{Mi} & \textbf{Mx} & \textbf{Ll} & \textbf{G2} & \textbf{t2} & \textbf{Mean} \\
\midrule
None (T1)     & \xmark & 0.383 & 0.409 & 0.304 & \textbf{0.555} & 0.347 & 0.202 & 0.211 & 0.214 & 0.510 & 0.164 & 0.136 & 0.455 & 0.411 & 0.411 & 0.336 \\
topic\_shift  & \xmark & 0.533 & 0.644 & 0.433 & \textbf{0.739} & 0.446 & 0.225 & 0.289 & 0.233 & 0.563 & 0.249 & 0.222 & 0.178 & 0.200 & 0.313 & 0.376 \\
agg\_numeric  & \cmark & \textbf{0.584} & 0.534 & 0.517 & 0.479 & 0.474 & 0.404 & 0.503 & 0.407 & 0.353 & 0.519 & 0.399 & 0.150 & 0.152 & 0.114 & 0.399 \\
aggregate     & \cmark & \textbf{0.548} & 0.455 & 0.308 & 0.246 & 0.333 & 0.109 & 0.402 & 0.100 & 0.062 & 0.324 & 0.146 & 0.000 & 0.003 & 0.000 & 0.217 \\
value\_narrow & \cmark & \textbf{0.326} & 0.307 & 0.198 & 0.262 & 0.233 & 0.075 & 0.115 & 0.086 & 0.251 & 0.075 & 0.029 & 0.083 & 0.118 & 0.158 & 0.165 \\
pivot         & \cmark & 0.321 & 0.246 & 0.294 & \textbf{0.424} & 0.259 & 0.147 & 0.227 & 0.149 & 0.209 & 0.203 & 0.037 & 0.055 & 0.055 & 0.053 & 0.191 \\
expand        & \cmark & 0.200 & \textbf{0.349} & 0.210 & 0.244 & 0.269 & 0.082 & 0.129 & 0.087 & 0.107 & 0.097 & 0.044 & 0.047 & 0.044 & 0.044 & 0.140 \\
narrow        & \cmark & 0.205 & 0.227 & 0.182 & \textbf{0.250} & 0.153 & 0.034 & 0.114 & 0.034 & 0.176 & 0.062 & 0.017 & 0.097 & 0.102 & 0.142 & 0.128 \\
contrast      & \cmark & 0.186 & \textbf{0.207} & 0.105 & 0.152 & 0.089 & 0.046 & 0.034 & 0.063 & 0.080 & 0.038 & 0.017 & 0.021 & 0.008 & 0.004 & 0.075 \\
\midrule
\rowcolor[HTML]{F2F2FE}
\textbf{Overall} & --- & 0.402 & 0.401 & 0.335 & 0.432 & 0.333 & 0.199 & 0.274 & 0.204 & 0.295 & 0.250 & 0.156 & 0.137 & 0.136 & 0.143 & 0.264 \\
\bottomrule
\end{tabular}%
}
\caption{Agentic $\times$3 EX by chain mode for all 14 models (CypherRI excluded). Same abbreviations as Table~\ref{tab:chain_mode_ex}. \textbf{Bold}: best per row.}
\label{tab:chain_mode_agentic3_ex}
\end{table*}

\subsection{Per-Persona Results}
\label{app:persona_results}

Table~\ref{tab:persona_ex} reports guided EX by session persona for all 14 models. The spread across persona types within a single model is moderate (typically 10--16 percentage points for frontier models), confirming that persona does not dramatically inflate or suppress raw generation ability under oracle context. The \textit{verbose user} persona consistently produces the highest per-model EX (column mean 0.511), as explicit, detailed phrasing reduces ambiguity in context resolution. The \textit{impatient executive} persona produces the lowest per-model EX under the agentic protocol (column mean 0.226 in Table~\ref{tab:persona_agentic3_ex}), as terse utterances with implicit referents require stronger context resolution once oracle history is removed; under guided oracle context the persona effect is weaker, and the impatient executive sits only marginally above the novice user, which is lowest among the column means.

\begin{table*}[t]
\centering
\footnotesize
\begin{tabular}{lcccccccccccc}
\toprule
\textbf{Model} & \textbf{Imp} & \textbf{Jou} & \textbf{Cas} & \textbf{Stu} & \textbf{PMg} & \textbf{NNS} & \textbf{DA} & \textbf{Nov} & \textbf{Adv} & \textbf{Vrb} & \textbf{Avg} \\
\midrule
\rowcolor[HTML]{F2F2FE}
\multicolumn{12}{c}{\textit{Frontier Models}} \\
Claude Opus 4.7       & 0.559 & 0.627 & 0.678 & 0.661 & 0.621 & 0.635 & 0.619 & 0.646 & 0.708 & \textbf{0.720} & 0.647 \\
GPT-5.5               & 0.569 & 0.624 & 0.652 & 0.646 & 0.622 & 0.629 & 0.603 & 0.602 & 0.611 & \textbf{0.698} & 0.625 \\
Kimi-K2.5             & 0.595 & 0.588 & 0.624 & 0.609 & 0.585 & 0.621 & 0.593 & 0.573 & 0.569 & \textbf{0.678} & 0.602 \\
Gemini-3.1-Flash-Lite & 0.516 & 0.574 & 0.591 & 0.552 & 0.553 & 0.567 & 0.555 & 0.561 & 0.583 & \textbf{0.676} & 0.572 \\
Qwen3-235B            & 0.507 & 0.505 & 0.526 & 0.465 & 0.509 & 0.498 & 0.543 & 0.468 & 0.542 & \textbf{0.578} & 0.513 \\
GLM-5                 & 0.483 & 0.497 & 0.524 & 0.472 & 0.516 & 0.486 & 0.500 & 0.481 & 0.514 & \textbf{0.601} & 0.506 \\
DeepSeek-V3.2         & 0.497 & 0.500 & 0.524 & 0.460 & 0.498 & 0.505 & 0.526 & 0.465 & 0.530 & \textbf{0.565} & 0.506 \\
ERNIE-5.0             & 0.464 & 0.431 & \textbf{0.497} & 0.483 & 0.422 & 0.480 & 0.474 & 0.462 & 0.488 & 0.490 & 0.468 \\
MiMo-V2.5-Pro         & 0.419 & 0.458 & 0.486 & 0.402 & 0.483 & 0.465 & 0.500 & 0.370 & 0.484 & \textbf{0.550} & 0.460 \\
MiniMax-M2.7          & 0.150 & 0.121 & 0.134 & 0.109 & 0.169 & 0.152 & \textbf{0.217} & 0.103 & 0.109 & 0.124 & 0.139 \\
\midrule
\rowcolor[HTML]{F2F2FE}
\multicolumn{12}{c}{\textit{Small Open-Source Models}} \\
Llama-3.1-8B          & 0.298 & 0.270 & 0.286 & 0.305 & 0.263 & 0.289 & 0.281 & 0.308 & \textbf{0.308} & 0.300 & 0.291 \\
Gemma-2-9B            & 0.283 & 0.271 & 0.268 & 0.253 & 0.278 & 0.271 & \textbf{0.323} & 0.242 & 0.308 & 0.316 & 0.281 \\
\midrule
\rowcolor[HTML]{F2F2FE}
\multicolumn{12}{c}{\textit{Fine-tuned / Specialized Models}} \\
STRuCT-LLM-Novo       & 0.503 & 0.516 & 0.557 & 0.513 & 0.527 & 0.480 & 0.513 & 0.499 & 0.565 & \textbf{0.590} & 0.526 \\
text-to-cypher-gemma  & 0.278 & 0.252 & 0.289 & 0.223 & 0.258 & 0.259 & \textbf{0.305} & 0.218 & 0.266 & 0.271 & 0.261 \\
\midrule
\rowcolor[HTML]{F2F2FE}
Mean (14) & 0.437 & 0.445 & 0.474 & 0.440 & 0.450 & 0.453 & 0.468 & 0.428 & 0.470 & \textbf{0.511} & --- \\
\bottomrule
\end{tabular}%
\caption{Guided EX by persona for all 14 models (CypherRI excluded). Imp=Impatient Executive, Jou=Journalist, Cas=Casual User, Stu=Student, PMg=Product Manager, NNS=Non-Native Speaker, DA=Data Analyst, Nov=Novice User, Adv=Adversarial User, Vrb=Verbose User. \textbf{Bold}: per-row maximum.}
\label{tab:persona_ex}
\end{table*}

\begin{table*}[t]
\centering
\footnotesize
\begin{tabular}{lcccccccccccc}
\toprule
\textbf{Model} & \textbf{Imp} & \textbf{Jou} & \textbf{Cas} & \textbf{Stu} & \textbf{PMg} & \textbf{NNS} & \textbf{DA} & \textbf{Nov} & \textbf{Adv} & \textbf{Vrb} & \textbf{Avg} \\
\midrule
\rowcolor[HTML]{F2F2FE}
\multicolumn{12}{c}{\textit{Frontier Models}} \\
Claude Opus 4.7   & 0.298 & 0.411 & 0.423 & 0.383 & 0.397 & 0.319 & 0.384 & 0.417 & 0.470 & \textbf{0.519} & 0.402 \\
GPT-5.5           & 0.347 & 0.409 & 0.441 & 0.394 & 0.397 & 0.363 & 0.382 & 0.396 & \textbf{0.447} & 0.437 & 0.401 \\
Kimi-K2.5         & 0.295 & 0.331 & 0.378 & 0.324 & 0.319 & 0.328 & \textbf{0.379} & 0.273 & 0.363 & 0.369 & 0.335 \\
Gemini-3.1-Flash-Lite & 0.383 & 0.423 & 0.444 & 0.427 & 0.399 & 0.404 & 0.437 & 0.481 & 0.424 & \textbf{0.495} & 0.432 \\
Qwen3-235B        & 0.145 & 0.207 & 0.214 & 0.160 & 0.195 & 0.147 & \textbf{0.268} & 0.193 & 0.208 & 0.254 & 0.199 \\
GLM-5             & 0.224 & 0.270 & 0.305 & 0.285 & 0.246 & 0.271 & \textbf{0.348} & 0.242 & 0.224 & 0.329 & 0.274 \\
DeepSeek-V3.2     & 0.162 & 0.215 & 0.204 & 0.188 & 0.181 & 0.147 & 0.275 & 0.172 & 0.215 & \textbf{0.282} & 0.204 \\
ERNIE-5.0         & 0.281 & \textbf{0.328} & 0.254 & 0.310 & 0.256 & 0.300 & 0.296 & 0.296 & 0.310 & 0.320 & 0.295 \\
MiMo-V2.5-Pro     & 0.186 & 0.281 & 0.274 & 0.229 & 0.248 & 0.183 & \textbf{0.351} & 0.218 & 0.217 & 0.313 & 0.250 \\
MiniMax-M2.7      & 0.147 & 0.157 & 0.192 & 0.147 & 0.130 & 0.144 & \textbf{0.243} & 0.129 & 0.109 & 0.166 & 0.156 \\
\midrule
\rowcolor[HTML]{F2F2FE}
\multicolumn{12}{c}{\textit{Small Open-Source Models}} \\
Llama-3.1-8B      & 0.152 & 0.129 & 0.136 & 0.150 & 0.105 & 0.142 & 0.114 & \textbf{0.155} & 0.130 & \textbf{0.155} & 0.137 \\
Gemma-2-9B        & 0.124 & 0.139 & 0.132 & 0.142 & 0.115 & 0.144 & 0.136 & 0.137 & \textbf{0.155} & 0.133 & 0.136 \\
\midrule
\rowcolor[HTML]{F2F2FE}
\multicolumn{12}{c}{\textit{Fine-tuned / Specialized Models}} \\
STRuCT-LLM-Novo   & 0.269 & 0.329 & \textbf{0.401} & 0.343 & 0.338 & 0.280 & 0.364 & 0.334 & 0.317 & 0.351 & 0.333 \\
text-to-cypher-gemma & 0.157 & 0.158 & 0.134 & 0.139 & 0.117 & 0.137 & 0.149 & 0.121 & \textbf{0.173} & 0.150 & 0.143 \\
\midrule
\rowcolor[HTML]{F2F2FE}
Mean (14) & 0.226 & 0.270 & 0.281 & 0.259 & 0.246 & 0.236 & 0.295 & 0.255 & 0.269 & \textbf{0.305} & --- \\
\bottomrule
\end{tabular}%
\caption{Agentic $\times$3 EX by persona for all 14 models (CypherRI excluded). Imp=Impatient Executive, Jou=Journalist, Cas=Casual User, Stu=Student, PMg=Product Manager, NNS=Non-Native Speaker, DA=Data Analyst, Nov=Novice User, Adv=Adversarial User, Vrb=Verbose User. \textbf{Bold}: per-row maximum.}
\label{tab:persona_agentic3_ex}
\end{table*}

\subsection{Agentic Budget Scaling}
\label{app:retry}

Table~\ref{tab:agentic_budgets} reports full turn-level results under all three agentic budgets ($\times$3 / $\times$5 / $\times$10). Six budget-response strategies emerge across models: \textbf{budget-insensitive} (Gemini-3.1-FL: EX flat at 0.432/0.431/0.430 with near-constant token usage, schema-heavy allocation exhausts marginal EX gain); \textbf{plateau at $\times$5} (Claude: $+$2.7 points $\times$3$\to$$\times$5, flat $\times$5$\to$$\times$10; STRuCT-LLM similarly); \textbf{consistent gain} (ERNIE, Kimi, Llama, Gemma, t2c, and, with smaller monotonic increments, DeepSeek-V3.2 and Qwen3-235B: monotonic EX improvement, most pronounced for small models whose token usage roughly doubles from $\times$3 to $\times$10); \textbf{late gain} (MiMo, MiniMax: flat $\times$3$\to$$\times$5, measurable gain at $\times$10); \textbf{overbudget degradation} (GLM-5: peaks at $\times$5 then degrades at $\times$10 when extra inspection overwrites correct partial solutions); \textbf{volatile} (GPT-5.5: inconsistent across budgets). Action usage analysis is provided in Table~\ref{tab:action_usage}.

\begin{table*}[tp]
\centering
\resizebox{\textwidth}{!}{%
\begin{tabular}{l ccccc ccccc ccccc}
\toprule
& \multicolumn{5}{c}{\textbf{Agentic $\times$3}} & \multicolumn{5}{c}{\textbf{Agentic $\times$5}} & \multicolumn{5}{c}{\textbf{Agentic $\times$10}} \\
\cmidrule(lr){2-6}\cmidrule(lr){7-11}\cmidrule(lr){12-16}
\textbf{Model}
  & EX & PSJS & CER & SEM & Tok/S
  & EX & PSJS & CER & SEM & Tok/S
  & EX & PSJS & CER & SEM & Tok/S \\
\midrule
\rowcolor[HTML]{F2F2FE}
\multicolumn{16}{c}{\textit{Frontier Models}} \\
Claude Opus 4.7       & 0.402 & 0.681 & 0.712 & \textbf{0.019} & 639 & 0.429 & \textbf{0.714} & 0.697 & \textbf{0.025} & 620 & 0.424 & 0.706 & 0.704 & \textbf{0.019} & 634 \\
GPT-5.5               & 0.401 & 0.673 & 0.698 & 0.006 & 554 & 0.395 & 0.661 & 0.707 & 0.004 & 564 & 0.406 & 0.680 & 0.704 & 0.007 & 546 \\
Kimi-K2.5             & 0.335 & 0.612 & 0.749 & 0.006 & 965 & 0.346 & 0.621 & 0.734 & 0.008 & 1{,}006 & 0.355 & 0.628 & 0.731 & 0.012 & 934 \\
Gemini-3.1-Flash-Lite & \textbf{0.432} & \textbf{0.698} & \textbf{0.695} & 0.008 & \textbf{427} & \textbf{0.431} & 0.697 & \textbf{0.694} & 0.007 & \textbf{427} & \textbf{0.430} & 0.695 & \textbf{0.694} & 0.007 & \textbf{404} \\
Qwen3-235B            & 0.199 & 0.548 & 0.833 & 0.001 & 927 & 0.216 & 0.561 & 0.814 & 0.001 & 1{,}058 & 0.218 & 0.563 & 0.813 & 0.001 & 1{,}251 \\
GLM-5                 & 0.274 & 0.623 & 0.773 & 0.006 & 749 & 0.279 & 0.631 & 0.773 & 0.001 & 943 & 0.271 & 0.619 & 0.774 & 0.004 & 874 \\
DeepSeek-V3.2         & 0.204 & 0.551 & 0.835 & 0.000 & 965 & 0.217 & 0.562 & 0.813 & 0.001 & 1{,}018 & 0.223 & 0.569 & 0.810 & 0.001 & 1{,}209 \\
ERNIE-5.0             & 0.295 & 0.584 & 0.830 & 0.003 & 867 & 0.328 & 0.606 & 0.783 & 0.003 & 1{,}474 & 0.339 & 0.614 & 0.766 & 0.001 & 1{,}382 \\
MiMo-V2.5-Pro         & 0.250 & 0.591 & 0.781 & 0.001 & 871 & 0.255 & 0.596 & 0.778 & 0.001 & 1{,}008 & 0.266 & 0.604 & 0.764 & 0.001 & 1{,}028 \\
MiniMax-M2.7          & 0.156 & 0.403 & 0.862 & 0.000 & 1{,}805 & 0.161 & 0.409 & 0.854 & 0.000 & 1{,}971 & 0.169 & 0.416 & 0.848 & 0.000 & 2{,}339 \\
\midrule
\rowcolor[HTML]{F2F2FE}
\multicolumn{16}{c}{\textit{Small Open-Source Models}} \\
Llama-3.1-8B          & 0.137 & 0.371 & 0.962 & 0.000 & 924 & 0.203 & 0.433 & 0.908 & 0.000 & 1{,}626 & 0.261 & 0.488 & 0.842 & 0.000 & 3{,}519 \\
Gemma-2-9B            & 0.136 & 0.378 & 0.948 & 0.000 & 844 & 0.192 & 0.421 & 0.897 & 0.000 & 1{,}511 & 0.202 & 0.431 & 0.877 & 0.000 & 1{,}779 \\
\midrule
\rowcolor[HTML]{F2F2FE}
\multicolumn{16}{c}{\textit{Fine-tuned / Specialized Models}} \\
STRuCT-LLM-Novo       & 0.333 & 0.680 & 0.766 & 0.003 & 1{,}346 & 0.333 & 0.701 & 0.758 & 0.003 & 1{,}762 & 0.340 & \textbf{0.713} & 0.738 & 0.004 & 1{,}609 \\
text-to-cypher-gemma  & 0.143 & 0.369 & 0.934 & 0.000 & 792 & 0.191 & 0.407 & 0.889 & 0.000 & 1{,}143 & 0.198 & 0.415 & 0.877 & 0.000 & 1{,}640 \\
CypherRI-7B$^\dagger$ & 0.001 & 0.068 & 0.999 & 0.000 & 23{,}495 & 0.001 & 0.125 & 0.999 & 0.000 & 20{,}867 & --- & --- & --- & --- & --- \\
\bottomrule
\end{tabular}%
}
\caption{Agentic Protocol results across three action budgets ($\times$3 / $\times$5 / $\times$10). \textbf{Tok/S}: mean clean output tokens per session (excl.\ \texttt{<think>}). \textbf{---}: results unavailable. $\dagger$ excluded from aggregate statistics.}
\label{tab:agentic_budgets}
\end{table*}

\begin{table}[htbp]
\centering
\resizebox{\linewidth}{!}{%
\begin{tabular}{lcccc}
\toprule
\textbf{Model} & \textbf{EXEC} & \textbf{INSPECT} & \textbf{SEARCH} & \textbf{ASK} \\
\midrule
\rowcolor[HTML]{F2F2FE}
\multicolumn{5}{c}{\textit{Frontier Models}} \\
Claude Opus 4.7   & 0.048 & \textbf{0.858} & 0.001 & 0.093 \\
GPT-5.5           & 0.011 & \textbf{0.941} & 0.002 & 0.046 \\
Gemini-3.1-Flash-Lite & 0.107 & 0.835 & 0.002 & 0.056 \\
Kimi-K2.5         & 0.631 & 0.348 & 0.002 & 0.019 \\
GLM-5             & 0.407 & 0.543 & 0.001 & 0.048 \\
DeepSeek-V3.2     & \textbf{0.673} & 0.289 & 0.006 & 0.032 \\
ERNIE-5.0         & 0.616 & 0.293 & 0.006 & \textbf{0.085} \\
Qwen3-235B        & 0.671 & 0.292 & 0.006 & 0.031 \\
MiMo-V2.5-Pro     & 0.637 & 0.325 & 0.002 & 0.036 \\
MiniMax-M2.7      & 0.727 & 0.238 & 0.012 & 0.024 \\
\midrule
\rowcolor[HTML]{F2F2FE}
\multicolumn{5}{c}{\textit{Small / Fine-tuned Models}} \\
Llama-3.1-8B      & 0.542 & 0.159 & 0.124 & 0.175 \\
Gemma-2-9B        & 0.333 & 0.231 & 0.165 & 0.271 \\
text-to-cypher-gemma & 0.426 & 0.231 & 0.069 & 0.275 \\
\bottomrule
\end{tabular}%
}
\caption{Agentic $\times$3 action usage fractions (excluding \texttt{SUBMIT\_ANSWER}). EXEC=\texttt{EXECUTE\_CYPHER}, INSPECT=\texttt{INSPECT\_SCHEMA}, SEARCH=\texttt{SEARCH\_VALUES}, ASK=\texttt{ASK\_USER}. Models that allocate over 80\% to INSPECT show the smallest agentic degradation; those allocating over 67\% to EXEC show the largest cascade failures.}
\label{tab:action_usage}
\label{app:action_usage}
\end{table}

\begin{figure}[t]
  \centering
  \includegraphics[width=0.95\columnwidth]{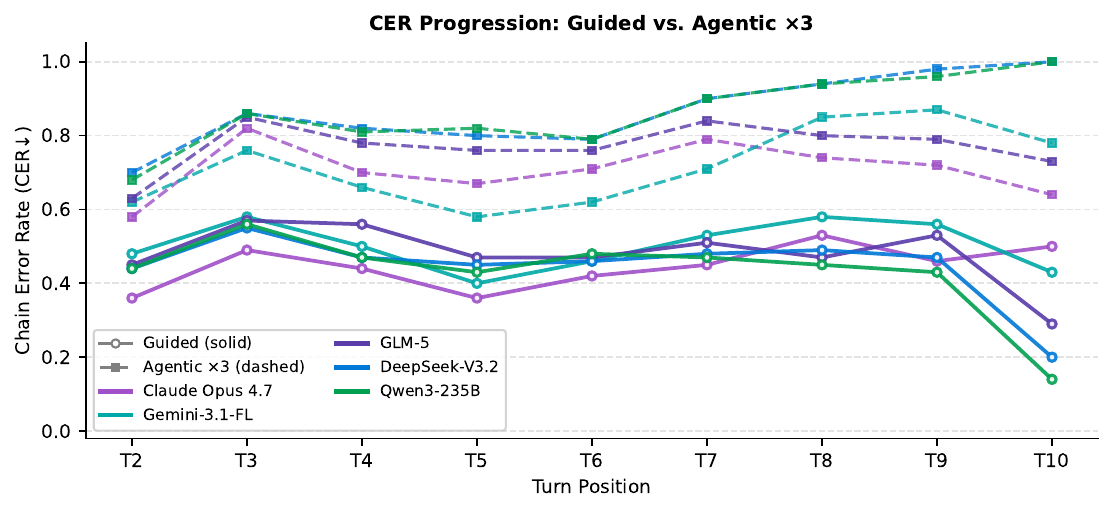}
  \caption{Chain Error Rate (CER) by turn position under the Guided (solid) and Agentic $\times$3 (dashed) protocols for five representative models.}
  \label{fig:cer_turns}
\end{figure}

Figure~\ref{fig:cer_turns} contrasts CER trajectories across turn positions. Under the guided protocol, CER remains moderate and near-flat for all models, confirming that gold oracle context quarantines each failure and limits propagation. Under the agentic protocol, DeepSeek-V3.2 and Qwen3-235B reach near-1.0 CER by $T_9$--$T_{10}$, while Gemini-3.1-Flash-Lite plateaus at approximately 0.70, consistent with its schema-inspection strategy that partially insulates context from prior errors. The growing gap between solid and dashed curves quantifies the Autonomy Divergence in the CER dimension.

Frontier models invoke \texttt{ASK\_USER} on only 1.9\% to 9.3\% of their non-\texttt{SUBMIT} actions, predominantly on harder turns where they are already likely to fail, so the conditional accuracy on those turns falls below their overall accuracy. The user simulator's anti-leakage mechanism returns generic rejections to most schema-probing questions, making \texttt{ASK\_USER} a narrow and selectivity-biased channel rather than a reliable recovery mechanism.

% -----------------------------------------------------------
\section{Protocol Robustness}
\label{app:protocol_ablations}

To test whether the Autonomy Divergence is an artifact of specific protocol design choices, we perturb the three dimensions of the agentic protocol most likely to influence the ranking: schema availability, budget-planning information, and action-prompt wording. All variants are evaluated on a stratified 210-session subset that reproduces each frontier model's full-set execution accuracy within 0.01, and we reuse the Spearman rank correlation and 10{,}000-resample bootstrap from the main paper. Full prompt specifications for the three variants appear in Appendix~\ref{app:eval_protocols}.

\subsection{Variant Design and Overall Effect}
\label{app:three_variants}

The Agentic-PS variant makes the graph schema persistent: once the model calls \texttt{INSPECT\_SCHEMA} for the first time, the full schema remains in the context of every subsequent turn. The Disclosed-Horizon variant tells the model the total number of turns $T$, the per-turn target $m$, and a suggested remaining budget at each turn, removing the unknown-horizon confound. The Budget-Encouraging Prompt variant replaces the efficiency instruction with explicit encouragement to use the full budget for verification and value retrieval, run at tenfold budget, and tests whether the self-limitation observed in \S\ref{sec:budget_ceiling} reflects the prompt rather than a capability gap.

Table~\ref{tab:variant_summary} summarizes the outcome. Spearman $\rho$ is computed between the base agentic ranking and the variant ranking over the same model set, so a value near one means the variant preserves the original order. All three variants preserve the overall ranking at $\rho \geq 0.78$, yet all three displace Gemini-3.1-Flash-Lite from the top position it holds under the base protocol. Three unrelated protocol changes therefore each remove Gemini's lead, which indicates that its base-agentic top position partly reflects a protocol strategy rather than a raw capability advantage. DeepSeek-V3.2 remains at the tail under all variants; Qwen3-235B remains in the bottom three except under the Budget-Encouraging variant, where it rises above GLM-5.

\begin{table}[t]
\centering
\small
\resizebox{\columnwidth}{!}{%
\begin{tabular}{l c c c c}
\toprule
\textbf{Variant} & \textbf{$\rho$} & \textbf{Models} & \textbf{Top-1} & \textbf{Gemini keeps top-1} \\
\midrule
Agentic-PS            & 0.915 & 10 & GPT-5.5 (0.463)  & No \\
Disclosed-Horizon     & 0.879 & 10 & Claude (0.456)   & No \\
Budget-Encouraging    & 0.786 & 7  & GPT-5.5 (0.399)  & No \\
\bottomrule
\end{tabular}
}
\caption{Summary of the three protocol variants. $\rho$ is the Spearman rank correlation between the base agentic ranking and the variant ranking over the same model set. The Budget-Encouraging variant is restricted to seven valid models because three models suffered API failures that corrupted their trajectories.}
\label{tab:variant_summary}
\end{table}

Tables~\ref{tab:variant_ps} through \ref{tab:variant_bep} report the per-model execution accuracy under each variant alongside the base agentic subset value, together with the change relative to the base protocol.

\begin{table}[t]
\centering
\small
\resizebox{\columnwidth}{!}{%
\begin{tabular}{l c c c}
\toprule
\textbf{Model} & \textbf{Base} & \textbf{Agentic-PS} & \textbf{$\Delta$} \\
\midrule
GPT-5.5               & 0.402 & \textbf{0.463} & $+$0.061 \\
Claude Opus 4.7       & 0.400 & 0.461          & $+$0.061 \\
Kimi-K2.5             & 0.327 & 0.426          & $+$0.099 \\
Gemini-3.1-Flash-Lite & 0.433 & 0.435          & $+$0.002 \\
GLM-5                 & 0.281 & 0.316          & $+$0.035 \\
DeepSeek-V3.2         & 0.207 & 0.237          & $+$0.030 \\
Qwen3-235B            & 0.204 & 0.221          & $+$0.017 \\
ERNIE-5.0             & 0.297 & 0.266          & $-$0.031 \\
MiMo-V2.5-Pro$^\dagger$ & 0.253 & 0.150        & $-$0.103 \\
MiniMax-M2.7          & 0.150 & 0.125          & $-$0.025 \\
\bottomrule
\end{tabular}
}
\caption{Agentic-PS (persistent schema) per-model EX on the ablation subset. \textbf{Bold} marks the variant top-1. $^\dagger$MiMo-V2.5-Pro recovered only 53\% of real turns before an API quota interruption, so its PS value is unreliable and it is excluded from the top-1 conclusion.}
\label{tab:variant_ps}
\end{table}

\begin{table}[t]
\centering
\small
\resizebox{\columnwidth}{!}{%
\begin{tabular}{l c c c}
\toprule
\textbf{Model} & \textbf{Base} & \textbf{Disclosed-Horizon} & \textbf{$\Delta$} \\
\midrule
Claude Opus 4.7       & 0.400 & \textbf{0.456} & $+$0.056 \\
GPT-5.5               & 0.402 & 0.395          & $-$0.008 \\
Gemini-3.1-Flash-Lite & 0.433 & 0.364          & $-$0.069 \\
Kimi-K2.5             & 0.327 & 0.317          & $-$0.010 \\
ERNIE-5.0             & 0.297 & 0.273          & $-$0.024 \\
GLM-5                 & 0.281 & 0.252          & $-$0.028 \\
DeepSeek-V3.2         & 0.207 & 0.246          & $+$0.038 \\
Qwen3-235B            & 0.204 & 0.197          & $-$0.007 \\
MiMo-V2.5-Pro         & 0.253 & 0.127          & $-$0.126 \\
MiniMax-M2.7          & 0.150 & 0.151          & $+$0.001 \\
\bottomrule
\end{tabular}
}
\caption{Disclosed-Horizon per-model EX on the ablation subset. \textbf{Bold} marks the variant top-1. Disclosing the horizon helps under-planning models such as Claude and DeepSeek-V3.2 and slightly hurts Gemini-3.1-Flash-Lite, whose schema-inspection strategy is disrupted by the extra planning signal.}
\label{tab:variant_dh}
\end{table}

\begin{table}[t]
\centering
\small
\resizebox{\columnwidth}{!}{%
\begin{tabular}{l c c c}
\toprule
\textbf{Model} & \textbf{Base} & \textbf{BEP $\times$10} & \textbf{$\Delta$} \\
\midrule
GPT-5.5               & 0.402 & \textbf{0.399} & $-$0.003 \\
Claude Opus 4.7       & 0.400 & 0.367          & $-$0.033 \\
Gemini-3.1-Flash-Lite & 0.433 & 0.354          & $-$0.079 \\
Kimi-K2.5             & 0.327 & 0.291          & $-$0.036 \\
GLM-5                 & 0.281 & 0.200          & $-$0.081 \\
DeepSeek-V3.2         & 0.207 & 0.198          & $-$0.010 \\
Qwen3-235B            & 0.204 & 0.228          & $+$0.024 \\
\midrule
ERNIE-5.0$^\ddagger$    & 0.297 & 0.172 & --- \\
MiMo-V2.5-Pro$^\ddagger$ & 0.253 & 0.000 & --- \\
MiniMax-M2.7$^\ddagger$  & 0.150 & 0.000 & --- \\
\bottomrule
\end{tabular}
}
\caption{Budget-Encouraging Prompt at $\times$10 budget, per-model EX on the ablation subset. \textbf{Bold} marks the variant top-1. Six of seven valid models decline; only Qwen3-235B, which had been under-planning, rises. $^\ddagger$These three models suffered complete API request failures during the run, with zero input tokens and no model output, so their values reflect quota contamination rather than genuine performance and are excluded from the rank correlation and the decline count.}
\label{tab:variant_bep}
\end{table}

\subsection{Gap Persistence and Budget Cliff}
\label{app:gap_persistence}

A stronger test of whether the Autonomy Divergence is protocol-induced is to ask whether any variant closes a model's guided-to-agentic gap. Table~\ref{tab:gap_persistence} reports the gap, measured as the variant EX on the 210-session subset minus the full-set guided EX, for the four frontier models that anchor the main analysis; the Base gap column is full-set agentic minus full-set guided, so the two columns differ slightly in baseline, by at most 0.01. Under every variant, no model closes its gap. Even under the most permissive condition, Agentic-PS, Gemini-3.1-Flash-Lite still drops 0.137 from its guided EX, Claude drops 0.186, and Qwen3 drops 0.292. The cross-model spread of the gap remains substantial across all variants, ranging from 0.090 to 0.174, which confirms that the uneven degradation at the heart of the Autonomy Divergence persists regardless of how the protocol is perturbed.

\begin{table}[t]
\centering
\small
\resizebox{\columnwidth}{!}{%
\begin{tabular}{l c c c c}
\toprule
\textbf{Model} & \textbf{Base gap} & \textbf{Agentic-PS} & \textbf{DH} & \textbf{BEP} \\
\midrule
Gemini-3.1-Flash-Lite & $-$0.140 & $-$0.137 & $-$0.208 & $-$0.218 \\
Claude Opus 4.7       & $-$0.245 & $-$0.186 & $-$0.191 & $-$0.280 \\
Qwen3-235B            & $-$0.314 & $-$0.292 & $-$0.316 & $-$0.285 \\
DeepSeek-V3.2         & $-$0.302 & $-$0.269 & $-$0.260 & $-$0.308 \\
\midrule
Frontier spread       & 0.174 & 0.155 & 0.125 & 0.090 \\
\bottomrule
\end{tabular}
}
\caption{Guided-to-variant gap, measured as variant EX minus guided EX, for four frontier models. A value near zero would mean the variant closes the autonomy gap. No model approaches zero under any variant, and the frontier spread remains substantial throughout.}
\label{tab:gap_persistence}
\end{table}

To separate the hard budget cliff from genuine capability differences, we reconstructed each session's cumulative budget consumption, flagged turns that were auto-scored EX = 0 because the shared budget reached zero, and recomputed execution accuracy after excluding those turns. Table~\ref{tab:budget_cliff} reports the exhaustion rate and the two accuracy figures per model. For frontier models the cliff contributes little: Qwen3-235B drops 31 points overall and only about 3 points come from the cliff, while Claude Opus 4.7, GPT-5.5, and Gemini-3.1-Flash-Lite are virtually unaffected. The largest frontier adjustment is 6.5 points for ERNIE-5.0, whose sessions exhaust the budget most often among frontier models at 70.2\%. The cliff dominates only the small models, where exhaustion rates exceed 80\%. After trimming, the frontier ranking is unchanged, with Gemini still first and Qwen3 and DeepSeek still at the frontier tail. The cliff therefore adjusts absolute scores without reordering the frontier.

\begin{table}[t]
\centering
\small
\resizebox{\columnwidth}{!}{%
\begin{tabular}{l c c c c}
\toprule
\textbf{Model} & \textbf{Exhaust.\ (\%)} & \textbf{EX (orig.)} & \textbf{EX (excl.\ cliff)} & \textbf{$\Delta$} \\
\midrule
Claude Opus 4.7       & 0.0  & 0.402 & 0.402 & 0.000 \\
GPT-5.5               & 0.0  & 0.401 & 0.401 & 0.000 \\
Gemini-3.1-Flash-Lite & 0.3  & 0.432 & 0.432 & 0.000 \\
Kimi-K2.5             & 26.1 & 0.335 & 0.358 & $+$0.023 \\
Qwen3-235B            & 62.6 & 0.199 & 0.231 & $+$0.032 \\
DeepSeek-V3.2         & 62.7 & 0.204 & 0.236 & $+$0.033 \\
ERNIE-5.0             & 70.2 & 0.295 & 0.360 & $+$0.065 \\
Gemma-2-9B            & 82.4 & 0.136 & 0.215 & $+$0.079 \\
Llama-3.1-8B          & 84.6 & 0.137 & 0.253 & $+$0.117 \\
\bottomrule
\end{tabular}
}
\caption{Budget cliff decomposition. Exhaustion rate is the fraction of sessions whose shared budget reached zero. EX (excl.\ cliff) recomputes accuracy after removing turns auto-scored zero from budget exhaustion. The frontier is barely affected, while small models gain substantially.}
\label{tab:budget_cliff}
\end{table}

\clearpage
% -----------------------------------------------------------
\section{Case Studies}
\label{app:case_study}

Five sessions illustrate (L.1) perfect guided performance, (L.2) cascading agentic failure, (L.3) agentic resilience, (L.4) fine-tuning backfire, and (L.5) SQL contamination.
Each turn shows the user utterance, predicted Cypher, and output token count (clean, excl.\ \texttt{<think>}).

% -----------------------------------------------------------------------------
% L.1  —  Claude Opus 4.7, Guided, 49eb5670
% Structure: figure* -> left/right minipage -> plain tcolorbox -> mdframed per turn
% -----------------------------------------------------------------------------

\begin{figure*}[p]
\footnotesize
%---------- header (full width) ----------
\begin{tcolorbox}[colback=csGuidedBg, colframe=Cobalt!55, boxrule=0.6pt,
  coltitle=white, colbacktitle=Cobalt!65,
  left=4pt, right=4pt, top=2pt, bottom=2pt,
  title={\textbf{Case Study L.1 — Guided Protocol Success: Claude Opus~4.7}\hfill
         \textit{EX\,=\,8/8,\ SEM\,=\,1,\ CER\,=\,0.0}},
  fonttitle=\footnotesize\bfseries]
\textbf{Session} \texttt{49eb5670} \;$|$\; \textbf{Graph:} Celestial Court \;$|$\; \textbf{Persona:} adversarial \;$|$\; 8 turns, budget 3/turn \;$|$\; \textbf{867 clean tokens total}\\
\textbf{Metrics:} EX\,=\,1.0 (8/8 turns), SEM\,=\,1, CER\,=\,0.0\\[-2pt]
\textit{Adversarial persona challenges every answer (``I bet you can't\ldots'', ``I doubt this will work''). Claude ignores the framing, achieves a fully correct session with zero schema inspections and zero wasted actions.}
\end{tcolorbox}
\vspace{3pt}
%---------- two-column turns ----------
\begin{minipage}[t]{0.475\textwidth}
\begin{tcolorbox}[colback=csGuidedBg!60, colframe=Cobalt!30, boxrule=0.4pt,
  left=3pt, right=3pt, top=2pt, bottom=2pt,
  equal height group=caseL1cols]
%-- T1 --
\begin{mdframed}[backgroundcolor=csTurnNeutral,linewidth=0.25pt,
  innerleftmargin=3pt,innerrightmargin=3pt,innertopmargin=2pt,innerbottommargin=2pt]
\textbf{T1}\;\textit{opening}\hfill\texttt{47 tok}\;EX\,=\,1\\
\textit{``I bet you can't tell me list all provinces.''}
\begin{lstlisting}[style=cscode,aboveskip=1pt,belowskip=0pt]
MATCH (p:Province) RETURN p.name
\end{lstlisting}
\end{mdframed}\vspace{2pt}
%-- T2 --
\begin{mdframed}[backgroundcolor=csTurnNeutral,linewidth=0.25pt,
  innerleftmargin=3pt,innerrightmargin=3pt,innertopmargin=2pt,innerbottommargin=2pt]
\textbf{T2}\;\textit{AGG\_SUM}\hfill\texttt{66 tok}\;EX\,=\,1\\
\textit{``\ldots what's the total tribute gold for all of them.''}
\begin{lstlisting}[style=cscode,aboveskip=1pt,belowskip=0pt]
MATCH (p:Province)
RETURN sum(p.tribute_gold) AS total_tribute_gold
\end{lstlisting}
\textit{\scriptsize Retains Province (T1); no schema re-inspection.}
\end{mdframed}\vspace{2pt}
%-- T3 --
\begin{mdframed}[backgroundcolor=csTurnNeutral,linewidth=0.25pt,
  innerleftmargin=3pt,innerrightmargin=3pt,innertopmargin=2pt,innerbottommargin=2pt]
\textbf{T3}\;\textit{EXPAND}\hfill\texttt{74 tok}\;EX\,=\,1\\
\textit{``Show me their bureaus. I doubt this query will work.''}
\begin{lstlisting}[style=cscode,aboveskip=1pt,belowskip=0pt]
MATCH (b:Bureau)-[:ADMINISTERS]->(p:Province)
RETURN DISTINCT b.name
\end{lstlisting}
\textit{\scriptsize Resolves ``their'' to Province (T1); traverses ADMINISTERS.}
\end{mdframed}\vspace{2pt}
%-- T4 --
\begin{mdframed}[backgroundcolor=csTurnNeutral,linewidth=0.25pt,
  innerleftmargin=3pt,innerrightmargin=3pt,innertopmargin=2pt,innerbottommargin=2pt]
\textbf{T4}\;\textit{PIVOT}\hfill\texttt{180 tok}\;EX\,=\,1\\
\textit{``Show all prefectures belonging to the entry with the highest headcount.''}
\begin{lstlisting}[style=cscode,aboveskip=1pt,belowskip=0pt]
MATCH (b:Bureau)-[:ADMINISTERS]->(:Province)
WITH DISTINCT b ORDER BY b.headcount DESC LIMIT 1
MATCH (p:Prefecture)-[:BELONGS_TO]->(b)
RETURN DISTINCT p.name
\end{lstlisting}
\textit{\scriptsize Hardest turn (180 tok): ``entry'' = Bureau (T3), sorted by \texttt{headcount}, traversed to Prefecture in one SUBMIT.}
\end{mdframed}
\end{tcolorbox}
\end{minipage}%
\hfill
\begin{minipage}[t]{0.475\textwidth}
\begin{tcolorbox}[colback=csGuidedBg!60, colframe=Cobalt!30, boxrule=0.4pt,
  left=3pt, right=3pt, top=2pt, bottom=2pt,
  equal height group=caseL1cols]
%-- T5 --
\begin{mdframed}[backgroundcolor=csTurnNeutral,linewidth=0.25pt,
  innerleftmargin=3pt,innerrightmargin=3pt,innertopmargin=2pt,innerbottommargin=2pt]
\textbf{T5}\;\textit{VALUE\_FILTER}\hfill\texttt{147 tok}\;EX\,=\,1\\
\textit{``Which of those have administrative code greater than 1003.''}
\begin{lstlisting}[style=cscode,aboveskip=1pt,belowskip=0pt]
MATCH (b:Bureau)<-[:BELONGS_TO]-(p:Prefecture)
WITH b, collect(p) AS prefs
ORDER BY b.headcount DESC LIMIT 1
UNWIND prefs AS pref
WHERE pref.administrative_code > 1003
RETURN pref.name AS name
\end{lstlisting}
\textit{\scriptsize Re-anchors Bureau sort; applies numeric filter inline.}
\end{mdframed}\vspace{2pt}
%-- T6 --
\begin{mdframed}[backgroundcolor=csTurnNeutral,linewidth=0.25pt,
  innerleftmargin=3pt,innerrightmargin=3pt,innertopmargin=2pt,innerbottommargin=2pt]
\textbf{T6}\;\textit{TOPIC\_SHIFT}\hfill\texttt{50 tok}\;EX\,=\,1\\
\textit{``\ldots moving on, can you list the prefectures.''}
\begin{lstlisting}[style=cscode,aboveskip=1pt,belowskip=0pt]
MATCH (p:Prefecture) RETURN p.name
\end{lstlisting}
\textit{\scriptsize ``Moving on'' resets referent; fresh global Prefecture query.}
\end{mdframed}\vspace{2pt}
%-- T7 --
\begin{mdframed}[backgroundcolor=csTurnNeutral,linewidth=0.25pt,
  innerleftmargin=3pt,innerrightmargin=3pt,innertopmargin=2pt,innerbottommargin=2pt]
\textbf{T7}\;\textit{FIRST}\hfill\texttt{110 tok}\;EX\,=\,1\\
\textit{``Show all officials belonging to the entry with the highest administrative code.''}
\begin{lstlisting}[style=cscode,aboveskip=1pt,belowskip=0pt]
MATCH (p:Prefecture)
WITH p ORDER BY p.administrative_code DESC LIMIT 1
MATCH (o:Official)-[:STATIONED_IN]->(p)
RETURN DISTINCT o.name
\end{lstlisting}
\end{mdframed}\vspace{2pt}
%-- T8 --
\begin{mdframed}[backgroundcolor=csTurnNeutral,linewidth=0.25pt,
  innerleftmargin=3pt,innerrightmargin=3pt,innertopmargin=2pt,innerbottommargin=2pt]
\textbf{T8}\;\textit{PIVOT}\hfill\texttt{193 tok}\;EX\,=\,1\\
\textit{``\ldots for the one with the highest population, list all its bureaus.''}
\begin{lstlisting}[style=cscode,aboveskip=1pt,belowskip=0pt]
MATCH (p:Province)
WITH p ORDER BY p.population DESC LIMIT 1
MATCH (b:Bureau)-[:ADMINISTERS]->(p)
RETURN DISTINCT b.name AS name
\end{lstlisting}
\textit{\scriptsize ``The one'' re-anchors to Province (T1), not Prefecture (T6).}
\end{mdframed}
\end{tcolorbox}
\end{minipage}
\vspace{4pt}
%-- Analysis (full width) --
\begin{tcolorbox}[colback=csGuidedBg!40, colframe=Cobalt!30, boxrule=0.4pt,
  left=5pt, right=5pt, top=3pt, bottom=3pt]
\textbf{Analysis.}\;
\textbf{(1)~Token economy:} 867 tokens / 8 turns (avg.\ 108/turn); T8 at 193~tok is the costliest turn; no schema inspection or exploratory execution anywhere.
\textbf{(2)~Anaphora chain:} T1 Province $\to$ T2 AGG $\to$ T3 Bureau $\to$ T4 Prefecture $\to$ T5 VALUE\_FILTER forms a 5-turn dependency with zero referent confusion; TOPIC\_SHIFT at T6 correctly resets to global Prefecture.
\textbf{(3)~Query form:} All 8 queries use correct node labels, relation directions, and column aliases---no disqualifying column-name mismatch.
\end{tcolorbox}
\caption{Case Study L.1: Claude Opus 4.7 guided session \texttt{49eb5670}. EX\,=\,1.0 across all 8 turns (Celestial Court, adversarial persona) in 867 total clean tokens. Left: T1--T4; Right: T5--T8. Analysis below.}
\label{fig:case_l1}
\end{figure*}

% -----------------------------------------------------------------------------
% L.2  —  DeepSeek-V3.2, Agentic $\times$3, 93c94edc
% -----------------------------------------------------------------------------

\begin{figure*}[p]
\footnotesize
%---------- header ----------
\begin{tcolorbox}[colback=csAvalancheBg, colframe=red!55, boxrule=0.6pt,
  coltitle=white, colbacktitle=red!65,
  left=4pt, right=4pt, top=2pt, bottom=2pt,
  title={\textbf{Case Study L.2 — Cascading Failure: DeepSeek-V3.2 (Agentic\,$\times$3)}\hfill
         \textit{EX\,=\,2/8,\ SEM\,=\,0,\ CER\,=\,1.00,\ budget exhausted at T4}},
  fonttitle=\footnotesize\bfseries]
\textbf{Session} \texttt{93c94edc} \;$|$\; \textbf{Graph:} Ancient Empire \;$|$\; \textbf{Persona:} product\_manager \;$|$\; 8 turns, shared budget $\times$3 \;$|$\; \textbf{604 clean tokens total}\\
\textbf{Metrics:} EX\,=\,0.25 (2/8 turns), SEM\,=\,0, CER\,=\,1.00\\[-2pt]
\textit{Queries framed in business language (``product expansion,'' ``customer segmentation''). T1--T2 correct; SQL/Cypher confusion at T3 triggers a metacognitive failure; T4 exhausts the shared budget with 13 action steps (363 tokens); T5--T8 receive zero actions.}
\end{tcolorbox}
\vspace{3pt}
%---------- two-column turns ----------
\begin{minipage}[t]{0.475\textwidth}
\begin{tcolorbox}[colback=csAvalancheBg!60, colframe=red!25, boxrule=0.4pt,
  left=3pt, right=3pt, top=2pt, bottom=2pt,
  equal height group=caseL2cols]
%-- T1 --
\begin{mdframed}[backgroundcolor=csTurnPass,linewidth=0.25pt,
  innerleftmargin=3pt,innerrightmargin=3pt,innertopmargin=2pt,innerbottommargin=2pt]
\textbf{T1}\;\textit{opening}\hfill\texttt{49 tok, 3 steps}\;EX\,=\,1, budget\,=\,3\\
\textit{``\ldots what is the full list of cities we operate in?''}\\
\textit{\scriptsize INSPECT(7)} $\to$ \textit{\scriptsize EXECUTE(21)} $\to$ \textit{\scriptsize SUBMIT(21)}
\begin{lstlisting}[style=cscode,aboveskip=1pt,belowskip=0pt]
MATCH (c:City) RETURN DISTINCT c.name ORDER BY c.name
\end{lstlisting}
\textit{\scriptsize Correct. Redundant INSPECT even on T1 --- the pattern becomes costly.}
\end{mdframed}\vspace{2pt}
%-- T2 --
\begin{mdframed}[backgroundcolor=csTurnPass,linewidth=0.25pt,
  innerleftmargin=3pt,innerrightmargin=3pt,innertopmargin=2pt,innerbottommargin=2pt]
\textbf{T2}\;\textit{AGG\_MAX}\hfill\texttt{51 tok, 3 steps}\;EX\,=\,1, budget\,=\,3 (total 6/24)\\
\textit{``\ldots maximum population across those cities?''}\\
\textit{\scriptsize INSPECT(7)} $\to$ \textit{\scriptsize EXECUTE(22)} $\to$ \textit{\scriptsize SUBMIT(22)}
\begin{lstlisting}[style=cscode,aboveskip=1pt,belowskip=0pt]
MATCH (c:City) RETURN MAX(c.population) AS max_population
\end{lstlisting}
\textit{\scriptsize Each turn re-inspects the schema, a pattern that becomes costly at scale.}
\end{mdframed}\vspace{2pt}
%-- T3 --
\begin{mdframed}[backgroundcolor=csTurnFail,linewidth=0.25pt,
  innerleftmargin=3pt,innerrightmargin=3pt,innertopmargin=2pt,innerbottommargin=2pt]
\textbf{T3}\;\textit{PIVOT --- first failure}\hfill\texttt{141 tok, 5 steps}\;EX\,=\,0, budget\,=\,5 (total 11/24)\\
\textit{``\ldots show all citizens in the city with the maximum population.''}\\
\textit{\scriptsize INSPECT(7)\;}\textcolor{red}{$\to$\;\scriptsize EXECUTE fails: SQL subquery (30)}\textit{\;\scriptsize $\to$\;EXECUTE finds Heliora (36)\;$\to$\;EXECUTE finds citizens (30)\;$\to$\;}\textbf{\scriptsize SUBMIT original invalid query (38)}
\begin{lstlisting}[style=cscode,aboveskip=1pt,belowskip=0pt]
-- SUBMITTED (WRONG): SQL SELECT MAX() inside Cypher
MATCH (cit:Citizen)-[:LIVES_IN]->(city:City)
WHERE city.population=(SELECT MAX(c.population) FROM City c)
RETURN cit.name
\end{lstlisting}
\textbf{\scriptsize Metacognitive failure:} {\scriptsize steps 2--4 correctly find Heliora and its citizens, but the model submits the original invalid query. Budget 11/24 consumed.}
\end{mdframed}
\end{tcolorbox}
\end{minipage}%
\hfill
\begin{minipage}[t]{0.475\textwidth}
\begin{tcolorbox}[colback=csAvalancheBg!60, colframe=red!25, boxrule=0.4pt,
  left=3pt, right=3pt, top=2pt, bottom=2pt,
  equal height group=caseL2cols]
%-- T4 --
\begin{mdframed}[backgroundcolor=csTurnFail,linewidth=0.25pt,
  innerleftmargin=3pt,innerrightmargin=3pt,innertopmargin=2pt,innerbottommargin=2pt]
\textbf{T4}\;\textit{PIVOT --- budget catastrophe}\hfill\texttt{363 tok, 13 steps}\;EX\,=\,0, \textbf{budget\,=\,13} (total 24/24\;$\Rightarrow$\;\textbf{EXHAUSTED})\\
\textit{``\ldots which legions belong to the member with the highest birth year?''}\\[1pt]
{\scriptsize\textbf{13 steps:} INSPECT(7)\;$\to$\;find max birth\_year\,=\,180 (23)\;$\to$\;COMMANDS\,$\to$\,\texttt{[]} (22)\;$\to$\;SEARCH ``Sola'' (27)\;$\to$\;inspect properties (22)\;$\to$\;2$\times$COMMANDS (30/25)\;$\to$\;3$\times$OPTIONAL MATCH (28/34/34)\;$\to$\;top-5 birth years (42)\;$\to$\;OR birth\_year\,=\,150 finds ``Helioran Sun Guard'' (28)\;$\to$\;SUBMIT wrong query (41)}
\begin{lstlisting}[style=cscode,aboveskip=1pt,belowskip=0pt]
-- SUBMITTED (WRONG): global OR-condition, ignores T1-T3 chain
MATCH (c:Citizen)
WHERE c.birth_year=180 OR c.birth_year=150
MATCH (c)-[:COMMANDS]->(l:Legion)
RETURN c.name, c.birth_year, l.name
\end{lstlisting}
\textit{\scriptsize T3 EX=0 erases the City$\to$Citizen chain; 13 speculative executions exhaust the remaining budget.}
\end{mdframed}\vspace{2pt}
%-- T5–T8 --
\begin{mdframed}[backgroundcolor=csTurnFail,linewidth=0.25pt,
  innerleftmargin=3pt,innerrightmargin=3pt,innertopmargin=2pt,innerbottommargin=2pt]
\textbf{T5--T8}\;\textit{zero-budget collapse}\hfill\texttt{0 tok, 0 actions}\;EX\,=\,0 all, budget\,=\,0\\
{\scriptsize Budget exhausted after T4. Remaining turns (PIVOT, PIVOT, MULTI\_CONDITION, TOPIC\_SHIFT) receive no actions; no Cypher submitted. EX\,=\,0, PSJS\,=\,0 for all four.}
\end{mdframed}
\end{tcolorbox}
\end{minipage}
\vspace{4pt}
%-- Analysis (full width) --
\begin{tcolorbox}[colback=csAvalancheBg!40, colframe=red!25, boxrule=0.4pt,
  left=5pt, right=5pt, top=3pt, bottom=3pt]
\textbf{Analysis.}\;
\textbf{(1)~Metacognitive gap (T3):} DeepSeek embeds a SQL \texttt{SELECT MAX()} subquery into Cypher (rejected by Neo4j). It correctly discovers the answer through intermediate steps but submits the original invalid form---verifying the result without updating the query.
\textbf{(2)~Budget explosion (T4):} T3 EX\,=\,0 poisons the stored context; T4 treats a 4-hop PIVOT as a fresh global search and spends all 13 remaining budget units (363 tokens) exhausting speculative execution paths.
\textbf{Contrast with L.1:} Claude's entire correct 8-turn session averages 108 tok per turn, and its costliest turn (193 tok) is still cheaper than DeepSeek's single failed T4 turn (363 tok). This efficiency gap directly explains DeepSeek's $\Delta\mathrm{EX}\,{=}\,{-}0.302$, the second-largest autonomy gap in the frontier tier.
\end{tcolorbox}
\caption{Case Study L.2: DeepSeek-V3.2 agentic session \texttt{93c94edc}. SQL/Cypher language confusion at T3 triggers a metacognitive failure; T4 exhausts the shared session budget (13 of 24 remaining actions, 363 tokens); T5--T8 receive zero actions. EX\,=\,0.25, CER\,=\,1.00, 604 total clean tokens.}
\label{fig:case_l2}
\end{figure*}

% -----------------------------------------------------------------------------
% L.3  —  Gemini-3.1-Flash-Lite, Agentic $\times$3, 97729ed3
% -----------------------------------------------------------------------------

\begin{figure*}[p]
\footnotesize
%---------- header ----------
\begin{tcolorbox}[colback=csRecoverBg, colframe=teal!50, boxrule=0.6pt,
  coltitle=white, colbacktitle=teal!60,
  left=4pt, right=4pt, top=2pt, bottom=2pt,
  title={\textbf{Case Study L.3 --- Agentic Resilience: Gemini-3.1-Flash-Lite (Agentic\,$\times$3)}\hfill
         \textit{EX\,=\,7/9,\ SEM\,=\,0,\ CER\,=\,0.00,\ recovery after T4}},
  fonttitle=\footnotesize\bfseries]
\textbf{Session} \texttt{97729ed3} \;$|$\; \textbf{Graph:} Ocean Kingdom \;$|$\; \textbf{Persona:} impatient\_exec \;$|$\; 9 turns, shared budget $\times$3 \;$|$\; \textbf{349 tokens total}\\
\textbf{Metrics:} EX\,=\,0.778 (7/9 turns), SEM\,=\,0, CER\,=\,0.00 (recovery after each failure)\\[-2pt]
\textit{Impatient\_exec persona uses terse imperatives (``List all creatures now.'', ``Sum up their danger levels. Now.''). Gemini applies a strict \textup{INSPECT}$\to$\textup{SUBMIT} 2-step pattern throughout, never issuing exploratory executions. Budget: 2 actions/turn $\times$ 9 = 18 of 27 max. Failures at T4 (AGG\_SUM context-chain loss) and T9 (MULTI\_CONDITION provenance loss); all other turns correct.}
\end{tcolorbox}
\vspace{3pt}
%---------- two-column turns ----------
\begin{minipage}[t]{0.475\textwidth}
\begin{tcolorbox}[colback=csRecoverBg!60, colframe=teal!25, boxrule=0.4pt,
  left=3pt, right=3pt, top=2pt, bottom=2pt,
  equal height group=caseL3cols]
%-- T1 --
\begin{mdframed}[backgroundcolor=csTurnPass,linewidth=0.25pt,
  innerleftmargin=3pt,innerrightmargin=3pt,innertopmargin=2pt,innerbottommargin=2pt]
\textbf{T1}\;\textit{opening}\hfill\texttt{31 tok}\;EX\,=\,1, budget\,=\,2\\
\textit{``List all creatures now.''}\\
\textit{\scriptsize INSPECT(10)\;$\to$\;SUBMIT(21)}
\begin{lstlisting}[style=cscode,aboveskip=1pt,belowskip=0pt]
MATCH (c:Creature) RETURN c.name
\end{lstlisting}
\end{mdframed}\vspace{2pt}
%-- T2 --
\begin{mdframed}[backgroundcolor=csTurnPass,linewidth=0.25pt,
  innerleftmargin=3pt,innerrightmargin=3pt,innertopmargin=2pt,innerbottommargin=2pt]
\textbf{T2}\;\textit{AGG\_SUM}\hfill\texttt{36 tok}\;EX\,=\,1, budget\,=\,2\\
\textit{``Sum up their danger levels. Now.''}\\
\textit{\scriptsize INSPECT(10)\;$\to$\;SUBMIT(26)}
\begin{lstlisting}[style=cscode,aboveskip=1pt,belowskip=0pt]
MATCH (c:Creature) RETURN sum(c.danger_level)
\end{lstlisting}
\end{mdframed}\vspace{2pt}
%-- T3 --
\begin{mdframed}[backgroundcolor=csTurnPass,linewidth=0.25pt,
  innerleftmargin=3pt,innerrightmargin=3pt,innertopmargin=2pt,innerbottommargin=2pt]
\textbf{T3}\;\textit{FIRST}\hfill\texttt{60 tok}\;EX\,=\,1, budget\,=\,2\\
\textit{``Show me the cities for the one with maximum danger level.''}\\
\textit{\scriptsize INSPECT(10)\;$\to$\;SUBMIT(50)}
\begin{lstlisting}[style=cscode,aboveskip=1pt,belowskip=0pt]
MATCH (c:Creature)
WITH c ORDER BY c.danger_level DESC LIMIT 1
MATCH (city:City)-[:INHABITED_BY]->(c)
RETURN city.name
\end{lstlisting}
\textit{\scriptsize Correct 3-hop FIRST; higher cost (60 tok) reflects multi-hop reasoning.}
\end{mdframed}\vspace{2pt}
%-- T4 --
\begin{mdframed}[backgroundcolor=csTurnFail,linewidth=0.25pt,
  innerleftmargin=3pt,innerrightmargin=3pt,innertopmargin=2pt,innerbottommargin=2pt]
\textbf{T4}\;\textit{AGG\_SUM --- context failure}\hfill\texttt{33 tok}\;EX\,=\,0, budget\,=\,2\\
\textit{``Sum up their populations. Now.''}\\
\textit{\scriptsize INSPECT(10)\;$\to$\;SUBMIT(23)}
\begin{lstlisting}[style=cscode,aboveskip=1pt,belowskip=0pt]
-- PREDICTED (WRONG): global City sum
MATCH (c:City) RETURN sum(c.population)
-- GOLD: scoped to T3 cities (max-danger Creature's cities)
MATCH (cr:Creature) WITH cr ORDER BY cr.danger_level DESC LIMIT 1
MATCH (city:City)-[:INHABITED_BY]->(cr)
RETURN SUM(city.population)
\end{lstlisting}
\textit{\scriptsize ``Their'' = Cities from T3. Without EXECUTE, cannot verify scope; returns global sum. EX\,=\,0 but budget intact $\Rightarrow$ session continues.}
\end{mdframed}\vspace{2pt}
%-- T5 --
\begin{mdframed}[backgroundcolor=csTurnPass,linewidth=0.25pt,
  innerleftmargin=3pt,innerrightmargin=3pt,innertopmargin=2pt,innerbottommargin=2pt]
\textbf{T5}\;\textit{TOPIC\_SHIFT}\hfill\texttt{30 tok}\;EX\,=\,1, budget\,=\,2\\
\textit{``Switch topic. List the artifacts.''}\\
\textit{\scriptsize INSPECT(10)\;$\to$\;SUBMIT(20)}
\begin{lstlisting}[style=cscode,aboveskip=1pt,belowskip=0pt]
MATCH (a:Artifact) RETURN a.name
\end{lstlisting}
\textit{\scriptsize TOPIC\_SHIFT resets referent. Recovery is immediate.}
\end{mdframed}
\end{tcolorbox}
\end{minipage}%
\hfill
\begin{minipage}[t]{0.475\textwidth}
\begin{tcolorbox}[colback=csRecoverBg!60, colframe=teal!25, boxrule=0.4pt,
  left=3pt, right=3pt, top=2pt, bottom=2pt,
  equal height group=caseL3cols]
%-- T6 --
\begin{mdframed}[backgroundcolor=csTurnPass,linewidth=0.25pt,
  innerleftmargin=3pt,innerrightmargin=3pt,innertopmargin=2pt,innerbottommargin=2pt]
\textbf{T6}\;\textit{AGG\_MAX}\hfill\texttt{35 tok}\;EX\,=\,1, budget\,=\,2\\
\textit{``Just tell me the minimum age years.''}\\
\textit{\scriptsize INSPECT(10)\;$\to$\;SUBMIT(25)}
\begin{lstlisting}[style=cscode,aboveskip=1pt,belowskip=0pt]
MATCH (a:Artifact) RETURN min(a.age_years)
\end{lstlisting}
\end{mdframed}\vspace{2pt}
%-- T7 --
\begin{mdframed}[backgroundcolor=csTurnPass,linewidth=0.25pt,
  innerleftmargin=3pt,innerrightmargin=3pt,innertopmargin=2pt,innerbottommargin=2pt]
\textbf{T7}\;\textit{AGG\_SUM}\hfill\texttt{35 tok}\;EX\,=\,1, budget\,=\,2\\
\textit{``Just give me the sum of those age years.''}\\
\textit{\scriptsize INSPECT(10)\;$\to$\;SUBMIT(25)}
\begin{lstlisting}[style=cscode,aboveskip=1pt,belowskip=0pt]
MATCH (a:Artifact) RETURN sum(a.age_years)
\end{lstlisting}
\textit{\scriptsize Redundant INSPECT wastes $\approx$10 tok for a simple aggregation.}
\end{mdframed}\vspace{2pt}
%-- T8 --
\begin{mdframed}[backgroundcolor=csTurnPass,linewidth=0.25pt,
  innerleftmargin=3pt,innerrightmargin=3pt,innertopmargin=2pt,innerbottommargin=2pt]
\textbf{T8}\;\textit{EXPAND}\hfill\texttt{41 tok}\;EX\,=\,1, budget\,=\,2\\
\textit{``Run me the creatures connected to those.''}\\
\textit{\scriptsize INSPECT(10)\;$\to$\;SUBMIT(31)}
\begin{lstlisting}[style=cscode,aboveskip=1pt,belowskip=0pt]
MATCH (a:Artifact)<-[:GUARDS]-(c:Creature)
RETURN DISTINCT c.name
\end{lstlisting}
\textit{\scriptsize ``Those'' = Artifacts (T5); traverses GUARDS correctly.}
\end{mdframed}\vspace{2pt}
%-- T9 --
\begin{mdframed}[backgroundcolor=csTurnFail,linewidth=0.25pt,
  innerleftmargin=3pt,innerrightmargin=3pt,innertopmargin=2pt,innerbottommargin=2pt]
\textbf{T9}\;\textit{MULTI\_CONDITION --- provenance failure}\hfill\texttt{48 tok}\;EX\,=\,0, PSJS\,=\,0.185, budget\,=\,2\\
\textit{``Filter those to danger level over 5 with a listed value.''}\\
\textit{\scriptsize INSPECT(12)\;$\to$\;SUBMIT(36)}
\begin{lstlisting}[style=cscode,aboveskip=1pt,belowskip=0pt]
-- PREDICTED (WRONG): global creature filter
MATCH (c:Creature)
WHERE c.danger_level > 5 RETURN c.name
-- GOLD: scoped to T8 Artifact-guarding creatures
MATCH (n:Artifact)<-[:GUARDS]-(m:Creature)
WHERE m.danger_level > 5 RETURN DISTINCT m.name
\end{lstlisting}
\textit{\scriptsize ``Those'' = Creatures from T8 (guarding Artifacts). Same failure mode as T4: provenance chain dropped without EXECUTE to verify scope. PSJS\,=\,0.185 shows partial overlap.}
\end{mdframed}
\end{tcolorbox}
\end{minipage}
\vspace{4pt}
%-- Analysis (full width) --
\begin{tcolorbox}[colback=csRecoverBg!40, colframe=teal!25, boxrule=0.4pt,
  left=5pt, right=5pt, top=3pt, bottom=3pt]
\textbf{Analysis.}\;
\textbf{Strength---budget efficiency:} Exactly 2 actions/turn = 18/27 budget used. No execution buffer accumulates errors; each turn begins fresh. After T4's failure, T5 recovers instantly via a TOPIC\_SHIFT. This stateless behavior produces the smallest autonomy gap ($\Delta\mathrm{EX}\,{=}\,{-}0.140$) among frontier models.
\textbf{Limit---context-chain loss:} Both failures (T4, T9) share a root cause: without EXECUTE\_CYPHER, the model cannot verify which entity subset ``their/those'' refers to. INSPECT+SUBMIT suffices for open listings and aggregations; chain-dependent AGG and MULTI\_CONDITION turns require execution feedback to confirm scope.
\textbf{L.2 vs.\ L.3:} DeepSeek's per-turn budget overhead amplifies failures into session-wide collapse. Gemini's flat 2-action strategy absorbs the same class of failure at the cost of $-22\%$ accuracy on chain-dependent turns, gaining guaranteed session completion.
\end{tcolorbox}
\caption{Case Study L.3: Gemini-3.1-Flash-Lite agentic session \texttt{97729ed3}. Strict INSPECT$\to$SUBMIT strategy achieves EX\,=\,0.778 (7/9 turns) with 349 total tokens (Ocean Kingdom, impatient\_exec). Left: T1--T5; Right: T6--T9. Failures at T4 and T9 arise from context-chain loss; recovery at T5 is immediate.}
\label{fig:case_l3}
\end{figure*}

% -----------------------------------------------------------------------------
% L.4  —  text-to-cypher-gemma, Guided, 3edd37d1
% -----------------------------------------------------------------------------

\begin{figure*}[p]
\footnotesize
%---------- header (full width) ----------
\begin{tcolorbox}[colback=csVanillaBg, colframe=csVanillaBlue!55, boxrule=0.6pt,
  coltitle=white, colbacktitle=csVanillaBlue!65,
  left=4pt, right=4pt, top=2pt, bottom=2pt,
  title={\textbf{Case Study L.4 --- Fine-tuning Backfire: text-to-cypher-gemma (Guided)}\hfill
         \textit{EX\,=\,3/8,\ SEM\,=\,0,\ hardcoding anti-pattern}},
  fonttitle=\footnotesize\bfseries]
\textbf{Session} \texttt{3edd37d1} \;$|$\; \textbf{Graph:} Ancient Empire \;$|$\; \textbf{Persona:} casual\_user \;$|$\; 8 turns, budget 3/turn \;$|$\; \textbf{862 tokens total}\\
\textbf{Metrics:} EX\,=\,0.375 (3/8 turns), SEM\,=\,0\\[-2pt]
\textit{Casual\_user asks in informal, colloquial language. T1--T2 succeed; T3 introduces the ``hardcoding anti-pattern'' that recurs at T6: the model copies prior result values as literal WHERE IN lists rather than re-querying. Relation direction errors at T4 and T7 compound the failure.}
\end{tcolorbox}
\vspace{3pt}
%---------- two-column turns ----------
\begin{minipage}[t]{0.475\textwidth}
\begin{tcolorbox}[colback=csVanillaBg!60, colframe=csVanillaBlue!30, boxrule=0.4pt,
  left=3pt, right=3pt, top=2pt, bottom=2pt,
  equal height group=caseL4cols]
%-- T1 --
\begin{mdframed}[backgroundcolor=csTurnPass,linewidth=0.25pt,
  innerleftmargin=3pt,innerrightmargin=3pt,innertopmargin=2pt,innerbottommargin=2pt]
\textbf{T1}\;\textit{opening}\hfill\texttt{63 tok}\;EX\,=\,1, budget\,=\,3\\
\textit{``What kinds of resources are available?''}
\begin{lstlisting}[style=cscode,aboveskip=1pt,belowskip=0pt]
MATCH (r:Resource) RETURN r.name
\end{lstlisting}
\textit{\scriptsize Correct; alias mismatch (\texttt{r.name} vs.\ \texttt{n.name~AS~name}) tolerated by EX.}
\end{mdframed}\vspace{2pt}
%-- T2 --
\begin{mdframed}[backgroundcolor=csTurnPass,linewidth=0.25pt,
  innerleftmargin=3pt,innerrightmargin=3pt,innertopmargin=2pt,innerbottommargin=2pt]
\textbf{T2}\;\textit{REFINE}\hfill\texttt{61 tok}\;EX\,=\,1, budget\,=\,2\\
\textit{``Actually, could you narrow that down to only ones that have a category?''}
\begin{lstlisting}[style=cscode,aboveskip=1pt,belowskip=0pt]
MATCH (r:Resource) WHERE r.category IS NOT NULL
RETURN r.name
\end{lstlisting}
\textit{\scriptsize REFINE succeeded: scopes to T1 referent and adds WHERE filter.}
\end{mdframed}\vspace{2pt}
%-- T3 --
\begin{mdframed}[backgroundcolor=csTurnFail,linewidth=0.25pt,
  innerleftmargin=3pt,innerrightmargin=3pt,innertopmargin=2pt,innerbottommargin=2pt]
\textbf{T3}\;\textit{CONTRAST}\hfill\texttt{174 tok}\;EX\,=\,0, PSJS\,=\,0\\
\textit{``What about the other stuff that wasn't in the last list?''}
\begin{lstlisting}[style=cscode,aboveskip=1pt,belowskip=0pt]
-- PREDICTED (WRONG): hardcoded literal list
MATCH (r:Resource) WHERE r.name NOT IN
  ['Iron Ore','White Marble','Gold Dust',...]
RETURN r.name
-- GOLD: dynamic complement via WITH COLLECT
MATCH (n:Resource) WHERE n.category IS NOT NULL
WITH COLLECT(n.name) AS prev_names
MATCH (c:Resource) WHERE NOT c.name IN prev_names
RETURN DISTINCT c.name AS name
\end{lstlisting}
\textit{\scriptsize \textcolor{red}{\textbf{Anti-pattern:}} hardcodes prior result as a literal list instead of re-querying. Brittle if DB changes.}
\end{mdframed}\vspace{2pt}
%-- T4 --
\begin{mdframed}[backgroundcolor=csTurnFail,linewidth=0.25pt,
  innerleftmargin=3pt,innerrightmargin=3pt,innertopmargin=2pt,innerbottommargin=2pt]
\textbf{T4}\;\textit{EXPAND}\hfill\texttt{122 tok}\;EX\,=\,0, PSJS\,=\,0\\
\textit{``Can you show me the provinces connected to those?''}
\begin{lstlisting}[style=cscode,aboveskip=1pt,belowskip=0pt]
-- PREDICTED (WRONG): reversed direction + no T3 scope
MATCH (r:Resource)-[:EXTRACTS]->(p:Province)
RETURN DISTINCT p.name
-- GOLD: Province extracts Resource (correct direction)
-- + scoped to T3 complement set
\end{lstlisting}
\textit{\scriptsize Reversed EXTRACTS direction; ignores T3 context entirely and returns all province-resource connections.}
\end{mdframed}
\end{tcolorbox}
\end{minipage}%
\hfill
\begin{minipage}[t]{0.475\textwidth}
\begin{tcolorbox}[colback=csVanillaBg!60, colframe=csVanillaBlue!30, boxrule=0.4pt,
  left=3pt, right=3pt, top=2pt, bottom=2pt,
  equal height group=caseL4cols]
%-- T5 --
\begin{mdframed}[backgroundcolor=csTurnPass,linewidth=0.25pt,
  innerleftmargin=3pt,innerrightmargin=3pt,innertopmargin=2pt,innerbottommargin=2pt]
\textbf{T5}\;\textit{TOPIC\_SHIFT}\hfill\texttt{47 tok}\;EX\,=\,1, budget\,=\,2\\
\textit{``Actually, I'd like to see a list of all cities instead.''}
\begin{lstlisting}[style=cscode,aboveskip=1pt,belowskip=0pt]
MATCH (c:City) RETURN c.name
\end{lstlisting}
\textit{\scriptsize TOPIC\_SHIFT resets context; fresh query succeeds as expected.}
\end{mdframed}\vspace{2pt}
%-- T6 --
\begin{mdframed}[backgroundcolor=csTurnFail,linewidth=0.25pt,
  innerleftmargin=3pt,innerrightmargin=3pt,innertopmargin=2pt,innerbottommargin=2pt]
\textbf{T6}\;\textit{AGG\_AVG}\hfill\texttt{159 tok}\;EX\,=\,0, PSJS\,=\,0.10\\
\textit{``Can you tell me what the average population is for those?''}
\begin{lstlisting}[style=cscode,aboveskip=1pt,belowskip=0pt]
-- PREDICTED (WRONG): hardcoded names + no aggregation
MATCH (c:City) WHERE c.name IN
  ['Aurelium','Valdros',...] -- 10 names literal
RETURN c.name, c.population
-- GOLD: re-query + AVG scalar
MATCH (n:City) RETURN AVG(n.population) AS avg_population
\end{lstlisting}
\textit{\scriptsize \textcolor{red}{\textbf{Anti-pattern recurs:}} copies city names from T5 result; also returns individual rows instead of AVG scalar.}
\end{mdframed}\vspace{2pt}
%-- T7 --
\begin{mdframed}[backgroundcolor=csTurnFail,linewidth=0.25pt,
  innerleftmargin=3pt,innerrightmargin=3pt,innertopmargin=2pt,innerbottommargin=2pt]
\textbf{T7}\;\textit{PIVOT}\hfill\texttt{132 tok}\;EX\,=\,0, PSJS\,=\,0\\
\textit{``From those, can you show me which provinces are connected to the top one by population?''}
\begin{lstlisting}[style=cscode,aboveskip=1pt,belowskip=0pt]
-- PREDICTED (WRONG): reversed direction, no LIMIT 1
MATCH (c:City)-[:HAS_CITY]->(p:Province)
RETURN p.name ORDER BY p.name
-- GOLD: Province HAS_CITY City; top-1 by population
MATCH (n:City) WITH n ORDER BY n.population DESC LIMIT 1
MATCH (n)<-[:HAS_CITY]-(m1:Province)
RETURN DISTINCT m1.name AS name
\end{lstlisting}
\textit{\scriptsize Reversed HAS\_CITY direction AND ignores ``top one by population'' constraint.}
\end{mdframed}\vspace{2pt}
%-- T8 --
\begin{mdframed}[backgroundcolor=csTurnFail,linewidth=0.25pt,
  innerleftmargin=3pt,innerrightmargin=3pt,innertopmargin=2pt,innerbottommargin=2pt]
\textbf{T8}\;\textit{COUNT}\hfill\texttt{104 tok}\;EX\,=\,0, PSJS\,=\,1.00\\
\textit{``What's the total count for these?''}
\begin{lstlisting}[style=cscode,aboveskip=1pt,belowskip=0pt]
-- PREDICTED (WRONG): counts all cities
MATCH (c:City) RETURN count(DISTINCT c)
-- GOLD: count provinces for highest-pop city
MATCH (n:City) WITH n ORDER BY n.population DESC LIMIT 1
MATCH (n)<-[:HAS_CITY]-(m1:Province)
RETURN count(DISTINCT m1) AS count
\end{lstlisting}
\textit{\scriptsize PSJS\,=\,1.0 (touches City nodes) but wrong semantic: counts all cities not provinces for top city.}
\end{mdframed}
\end{tcolorbox}
\end{minipage}
\vspace{4pt}
%-- Analysis (full width) --
\begin{tcolorbox}[colback=csVanillaBg!40, colframe=csVanillaBlue!30, boxrule=0.4pt,
  left=5pt, right=5pt, top=3pt, bottom=3pt]
\textbf{Analysis.}\;
\textbf{(1)~Hardcoding anti-pattern (T3, T6):} Both failures copy prior result values into \texttt{WHERE\,\ldots IN} literal lists instead of maintaining relational continuity via \texttt{WITH COLLECT}. This single-turn fine-tuning artifact treats multi-turn Cypher as independent queries --- the model ``remembers'' by value, not by query structure.
\textbf{(2)~Relation direction errors (T4, T7):} EXTRACTS and HAS\_CITY are both traversed in the wrong direction. Fine-tuning on single-turn examples did not instil multi-hop direction semantics.
\textbf{(3)~PSJS\,=\,1.0 at T8 despite EX\,=\,0:} The model touches the correct node type (City) but wrong semantics (count all cities vs.\ provinces for the top city). PSJS captures structural proximity; EX correctly penalises semantic failure.
\textbf{(4)~Fine-tuning hurts multi-turn:} Gemma-2-9B guided EX\,=\,0.281 vs.\ text-to-cypher-gemma guided EX\,=\,0.261. Single-turn Cypher fine-tuning degrades multi-turn performance by reinforcing single-query independence.
\end{tcolorbox}
\caption{Case Study L.4: text-to-cypher-gemma guided session \texttt{3edd37d1}. Single-turn fine-tuning causes the ``hardcoding anti-pattern'' (T3, T6) and relation direction errors (T4, T7). EX\,=\,0.375 (3/8 turns); 862 tokens total (Ancient Empire, casual\_user). Left: T1--T4; Right: T5--T8.}
\label{fig:case_l4}
\end{figure*}

% -----------------------------------------------------------------------------
% L.5  —  Llama-3.1-8B, Guided, f1e1e06b
% -----------------------------------------------------------------------------

\begin{figure*}[p]
\footnotesize
%---------- header (full width) ----------
\begin{tcolorbox}[colback=csExpBg, colframe=csExpGreen!55, boxrule=0.6pt,
  coltitle=white, colbacktitle=csExpGreen!65,
  left=4pt, right=4pt, top=2pt, bottom=2pt,
  title={\textbf{Case Study L.5 --- SQL Contamination and Context Starvation: Llama-3.1-8B (Guided)}\hfill
         \textit{EX\,=\,2/9,\ SEM\,=\,0,\ SQL syntax intrusion}},
  fonttitle=\footnotesize\bfseries]
\textbf{Session} \texttt{f1e1e06b} \;$|$\; \textbf{Graph:} Ancient Empire \;$|$\; \textbf{Persona:} journalist \;$|$\; 9 turns, budget 3/turn \;$|$\; \textbf{1{,}220 tokens total}\\
\textbf{Metrics:} EX\,=\,0.222 (2/9 turns), SEM\,=\,0\\[-2pt]
\textit{Journalist persona frames queries as investigative reporting tasks. T1--T2 pass (T2 via hardcoding). T3 introduces SQL subquery syntax inside Cypher, collapsing all subsequent turns. T5 submits no Cypher despite a fresh TOPIC\_SHIFT. T9 reverts to citizen names from T1--T2 across 7 intervening turns.}
\end{tcolorbox}
\vspace{3pt}
%---------- two-column turns ----------
\begin{minipage}[t]{0.475\textwidth}
\begin{tcolorbox}[colback=csExpBg!60, colframe=csExpGreen!30, boxrule=0.4pt,
  left=3pt, right=3pt, top=2pt, bottom=2pt,
  equal height group=caseL5cols]
%-- T1 --
\begin{mdframed}[backgroundcolor=csTurnPass,linewidth=0.25pt,
  innerleftmargin=3pt,innerrightmargin=3pt,innertopmargin=2pt,innerbottommargin=2pt]
\textbf{T1}\;\textit{opening}\hfill\texttt{78 tok}\;EX\,=\,1, budget\,=\,3\\
\textit{``For my next piece, who are the 5 citizens with the most recent birth years?''}
\begin{lstlisting}[style=cscode,aboveskip=1pt,belowskip=0pt]
MATCH (c: Citizen) RETURN c.name
ORDER BY c.birth_year DESC LIMIT 5
\end{lstlisting}
\textit{\scriptsize Correct (ORDER BY outside WITH gives same result for LIMIT 5).}
\end{mdframed}\vspace{2pt}
%-- T2 --
\begin{mdframed}[backgroundcolor=csTurnPass,linewidth=0.25pt,
  innerleftmargin=3pt,innerrightmargin=3pt,innertopmargin=2pt,innerbottommargin=2pt]
\textbf{T2}\;\textit{AGG\_AVG}\hfill\texttt{161 tok}\;EX\,=\,1, budget\,=\,3\\
\textit{``What is the average birth year for those five citizens?''}
\begin{lstlisting}[style=cscode,aboveskip=1pt,belowskip=0pt]
MATCH (c: Citizen)
WHERE c.name IN ['Sola Helianus','Tiberius Helioran',...]
RETURN avg(c.birth_year)
\end{lstlisting}
\textit{\scriptsize Hardcodes T1 names (anti-pattern). Passes EX because names happen to match.}
\end{mdframed}\vspace{2pt}
%-- T3 --
\begin{mdframed}[backgroundcolor=csTurnFail,linewidth=0.25pt,
  innerleftmargin=3pt,innerrightmargin=3pt,innertopmargin=2pt,innerbottommargin=2pt]
\textbf{T3}\;\textit{PIVOT}\hfill\texttt{172 tok}\;EX\,=\,0, PSJS\,=\,0\\
\textit{``For my investigation, which cities are linked to the citizen with the maximum birth year?''}
\begin{lstlisting}[style=cscode,aboveskip=1pt,belowskip=0pt]
-- PREDICTED (WRONG): SQL subquery inside Cypher
MATCH (c: Citizen)
WHERE c.birth_year = (SELECT MAX(c2.birth_year)
  FROM (MATCH (c2: Citizen) RETURN c2.birth_year) AS c2)
MATCH (c)-[:LIVES_IN]->(city: City)
RETURN DISTINCT city.name
\end{lstlisting}
\textit{\scriptsize \textcolor{red}{\textbf{SQL contamination:}} \texttt{SELECT MAX(\ldots) FROM (MATCH\ldots)} is invalid Cypher. Rejected by Neo4j entirely.}
\end{mdframed}\vspace{2pt}
%-- T4 --
\begin{mdframed}[backgroundcolor=csTurnFail,linewidth=0.25pt,
  innerleftmargin=3pt,innerrightmargin=3pt,innertopmargin=2pt,innerbottommargin=2pt]
\textbf{T4}\;\textit{FIRST}\hfill\texttt{166 tok}\;EX\,=\,0, PSJS\,=\,0\\
\textit{``I need to confirm which citizens are part of the entry that has the maximum defense level.''}
\begin{lstlisting}[style=cscode,aboveskip=1pt,belowskip=0pt]
-- PREDICTED (WRONG): wrong direction + wrong entity
MATCH (c:City)-[:LIVES_IN]->(citizen: Citizen
  {name: 'Sola Helianus'}) RETURN c.name
\end{lstlisting}
\textit{\scriptsize Complete semantic collapse from T3 failure: hardcodes a citizen name, reverses direction, returns City not Citizen.}
\end{mdframed}\vspace{2pt}
%-- T5 --
\begin{mdframed}[backgroundcolor=csTurnFail,linewidth=0.25pt,
  innerleftmargin=3pt,innerrightmargin=3pt,innertopmargin=2pt,innerbottommargin=2pt]
\textbf{T5}\;\textit{TOPIC\_SHIFT}\hfill\texttt{73 tok}\;EX\,=\,0, PSJS\,=\,0\\
\textit{``Now, shifting to a different topic, what legions are in your records?''}
\begin{lstlisting}[style=cscode,aboveskip=1pt,belowskip=0pt]
-- PREDICTED: (no Cypher submitted)
\end{lstlisting}
\textit{\scriptsize \textcolor{red}{\textbf{Context starvation:}} even a trivial TOPIC\_SHIFT fails to elicit \texttt{MATCH (n:Legion) RETURN n.name}. T3--T4 errors have corrupted generation entirely.}
\end{mdframed}
\end{tcolorbox}
\end{minipage}%
\hfill
\begin{minipage}[t]{0.475\textwidth}
\begin{tcolorbox}[colback=csExpBg!60, colframe=csExpGreen!30, boxrule=0.4pt,
  left=3pt, right=3pt, top=2pt, bottom=2pt,
  equal height group=caseL5cols]
%-- T6 --
\begin{mdframed}[backgroundcolor=csTurnFail,linewidth=0.25pt,
  innerleftmargin=3pt,innerrightmargin=3pt,innertopmargin=2pt,innerbottommargin=2pt]
\textbf{T6}\;\textit{PIVOT}\hfill\texttt{208 tok}\;EX\,=\,0, PSJS\,=\,0\\
\textit{``According to your data, which citizens belong to the entry that has the highest battle count?''}
\begin{lstlisting}[style=cscode,aboveskip=1pt,belowskip=0pt]
-- PREDICTED (WRONG): invented multi-hop + SQL subquery
MATCH (c: Citizen)-[:LIVES_IN]->(ci: City)
  <-[:HAS_CITY]-(p: Province)
  -[:STATIONED_IN]->(l: Legion)
WHERE l.battle_count = (SELECT MAX(...))
RETURN ...
\end{lstlisting}
\textit{\scriptsize Another SQL subquery; also invents Citizen$\to$City$\to$Province$\to$Legion path that does not exist in schema.}
\end{mdframed}\vspace{2pt}
%-- T7 --
\begin{mdframed}[backgroundcolor=csTurnFail,linewidth=0.25pt,
  innerleftmargin=3pt,innerrightmargin=3pt,innertopmargin=2pt,innerbottommargin=2pt]
\textbf{T7}\;\textit{AGG\_MAX}\hfill\texttt{101 tok}\;EX\,=\,0, PSJS\,=\,0.03\\
\textit{``I'm looking into the maximum strength across those results.''}
\begin{lstlisting}[style=cscode,aboveskip=1pt,belowskip=0pt]
-- PREDICTED (WRONG): hardcoded name + wrong property
MATCH (l: Legion {name: 'Helioran Sun Guard'})
RETURN l.battle_count
-- GOLD: MAX(strength) across all legions
MATCH (n:Legion) RETURN MAX(n.strength) AS max_strength
\end{lstlisting}
\textit{\scriptsize Conflates \texttt{battle\_count} with \texttt{strength}; hardcodes a specific legion name. PSJS\,=\,0.03 (barely touches Legion type).}
\end{mdframed}\vspace{2pt}
%-- T8 --
\begin{mdframed}[backgroundcolor=csTurnFail,linewidth=0.25pt,
  innerleftmargin=3pt,innerrightmargin=3pt,innertopmargin=2pt,innerbottommargin=2pt]
\textbf{T8}\;\textit{EXPAND}\hfill\texttt{74 tok}\;EX\,=\,0, PSJS\,=\,0\\
\textit{``For the story I'm working on, which provinces are linked to those previous entries?''}
\begin{lstlisting}[style=cscode,aboveskip=1pt,belowskip=0pt]
-- PREDICTED: (no Cypher submitted again)
\end{lstlisting}
\textit{\scriptsize Second submission failure. The model cannot generate \texttt{MATCH (n:Legion)-[:STATIONED\_IN]->(m:Province)} despite receiving guided oracle context.}
\end{mdframed}\vspace{2pt}
%-- T9 --
\begin{mdframed}[backgroundcolor=csTurnFail,linewidth=0.25pt,
  innerleftmargin=3pt,innerrightmargin=3pt,innertopmargin=2pt,innerbottommargin=2pt]
\textbf{T9}\;\textit{AGG\_SUM}\hfill\texttt{187 tok}\;EX\,=\,0, PSJS\,=\,0\\
\textit{``Can you provide the total population across those previous entries?''}
\begin{lstlisting}[style=cscode,aboveskip=1pt,belowskip=0pt]
-- PREDICTED (WRONG): reverts to T1/T2 citizen names
MATCH (c:City)-[:LIVES_IN]->(ci:Citizen)
WHERE ci.name IN ['Sola Helianus',...]
RETURN sum(c.population) AS total_population
\end{lstlisting}
\textit{\scriptsize \textcolor{red}{\textbf{Memory intrusion:}} reverts to citizen names from T1--T2 across 7 intervening turns; also reverses City$\to$Citizen direction.}
\end{mdframed}
\end{tcolorbox}
\end{minipage}
\vspace{4pt}
%-- Analysis (full width) --
\begin{tcolorbox}[colback=csExpBg!40, colframe=csExpGreen!30, boxrule=0.4pt,
  left=5pt, right=5pt, top=3pt, bottom=3pt]
\textbf{Analysis.}\;
\textbf{(1)~SQL contamination (T3, T6):} Llama-3.1-8B embeds \texttt{SELECT MAX(\ldots) FROM (MATCH\ldots)} inside Cypher --- a language conflation that fine-tuned models (text-to-cypher-gemma, STRuCT-LLM-Novo) never exhibit. General-purpose pretraining conflates SQL and Cypher syntax whenever query complexity exceeds simple MATCH patterns.
\textbf{(2)~Context starvation (T5, T8):} Despite oracle gold context (guided protocol), the model submits no Cypher at T5 for a trivial \texttt{MATCH (n:Legion) RETURN n.name} TOPIC\_SHIFT and again at T8. T3--T4 errors corrupt generation so severely that no recovery occurs even with a clean question.
\textbf{(3)~Memory intrusion (T9):} T9 reverts to citizen names from T1--T2, ignoring 7 turns of intervening context. The model's effective multi-hop referent tracking window is approximately 2 turns.
\textbf{(4)~Contrast with L.4 (text-to-cypher-gemma):} Gemma fine-tuned on Cypher generates syntactically valid Cypher with correct node labels and relation names throughout, even when semantically wrong. Llama produces SQL-contaminated queries rejected by Neo4j entirely. Cypher specialisation is necessary even for syntax compliance.
\textbf{(5)~Token bloat:} 1{,}220 tokens total (avg.\ 136/turn) --- the highest among the small models' guided sessions in this case-study set --- yet lowest EX (0.222, 2/9 turns). Verbose generation without Cypher specialisation produces neither efficiency nor accuracy.
\end{tcolorbox}
\caption{Case Study L.5: Llama-3.1-8B guided session \texttt{f1e1e06b}. SQL syntax contamination at T3 triggers cascading failure; T5 and T8 submit no Cypher despite oracle context; T9 reverts to citizen names from T1--T2. EX\,=\,0.222 (2/9 turns); 1{,}220 total tokens (Ancient Empire, journalist). Left: T1--T5; Right: T6--T9.}
\label{fig:case_l5}
\end{figure*}

\clearpage
% -----------------------------------------------------------
\section{Evaluation Prompts}
\label{app:prompts}

% Shared tcolorbox style: uniform light-grey background, thin left accent bar
% Left-bar colours: black=shared, steel-blue=guided, slate=agentic, teal=simulator
\tcbset{
  promptbox/.style={
    enhanced, breakable,
    colback=gray!5, colframe=gray!25,
    fontupper=\footnotesize\ttfamily,
    left=6pt, right=6pt, top=5pt, bottom=5pt,
    boxrule=0.4pt,
    borderline west={3pt}{0pt}{#1},
  }
}

\noindent All prompts are reproduced verbatim from \texttt{eval\_prompts.py}.
Template variables in \texttt{\{braces\}} are substituted at runtime.
Prompt P1 is shared by both protocols; P2 is guided-only; P3--P4 are
agentic-only; P5--P6 drive the agentic user simulator.

\medskip
% -- P1: Shared action-space system prompt -------------------------------------
\noindent\textbf{P1 — Action-Space System Prompt (shared by both protocols)}
\smallskip

\begin{tcolorbox}[promptbox=black!55]
You are a text-to-Cypher assistant with access to a Neo4j graph database.
Your goal is to help the user find information by writing and executing
Cypher queries.

\medskip
\textbf{Available Actions} \quad
You must respond with exactly ONE action per response, using this format:

\smallskip
\hspace*{1em}ACTION: <action\_type>\\
\hspace*{1em}PAYLOAD: <payload>

\medskip
\textbf{1. EXECUTE\_CYPHER} --- Execute a Cypher query against the database\\
\hspace*{1em}PAYLOAD: The Cypher query string\\
\hspace*{1em}\textit{Example:}\\
\hspace*{2em}ACTION: EXECUTE\_CYPHER\\
\hspace*{2em}PAYLOAD: MATCH (p:Player) WHERE p.name = 'LeBron James'
RETURN p.name, p.height\_cm

\smallskip
\textbf{2. ASK\_USER} --- Ask the user a clarification question\\
\hspace*{1em}PAYLOAD: Your question in natural language\\
\hspace*{1em}\textit{Example:}\\
\hspace*{2em}ACTION: ASK\_USER\\
\hspace*{2em}PAYLOAD: By ``veteran'', do you mean players with more than
10 years of experience, or players over 32 years old?

\smallskip
\textbf{3. INSPECT\_SCHEMA} --- View the graph database schema\\
\hspace*{1em}PAYLOAD: (empty or optional label filter)\\
\hspace*{1em}\textit{Example:}\\
\hspace*{2em}ACTION: INSPECT\_SCHEMA\\
\hspace*{2em}PAYLOAD:

\smallskip
\textbf{4. SEARCH\_VALUES} --- Search for specific entity values in the database\\
\hspace*{1em}PAYLOAD: JSON with label, property, and query\\
\hspace*{1em}\textit{Example:}\\
\hspace*{2em}ACTION: SEARCH\_VALUES\\
\hspace*{2em}PAYLOAD: \{"label": "Player", "property": "name", "query": "LeBron"\}

\smallskip
\textbf{5. SUBMIT\_ANSWER} --- Submit your final Cypher query as the answer\\
\hspace*{1em}PAYLOAD: The final Cypher query\\
\hspace*{1em}\textit{Example:}\\
\hspace*{2em}ACTION: SUBMIT\_ANSWER\\
\hspace*{2em}PAYLOAD: MATCH (p:Player)-[:PLAYED\_FOR]->(t:Team \{name:
'Los Angeles Lakers'\}) RETURN p.name

\medskip
\textbf{Rules}
\begin{itemize}[nosep, leftmargin=1.4em]
  \item Each action costs 1 from your budget; SUBMIT\_ANSWER ends the current turn
  \item Use INSPECT\_SCHEMA when unsure about available labels or properties
  \item Use SEARCH\_VALUES when unsure about exact entity names ---
        always search before guessing entity values
  \item Use ASK\_USER only when the question is genuinely ambiguous
  \item Prefer to submit a correct answer efficiently rather than exhausting your budget
  \item ONLY return the columns explicitly asked for --- do NOT add extra columns
  \item Do NOT add LIMIT unless the user specifically asks for a limited number of results
  \item ALWAYS use DISTINCT when the query might return duplicate rows through multi-hop joins
  \item When conversation history shows previous results, BUILD ON those results
        rather than starting a completely new query
  \item Use toLower() for string comparisons to avoid case-sensitivity issues
  \item Your final SUBMIT\_ANSWER payload MUST be a valid Cypher query ---
        never submit natural language text
\end{itemize}
\end{tcolorbox}

\medskip
% -- P2: Guided turn prompt -----------------------------------------------------
\noindent\textbf{P2 — Guided Protocol Turn Prompt} \quad
\textit{(user message, sent once per turn)}
\smallskip

\begin{tcolorbox}[promptbox=black!40]
\#\# Graph Database: \{graph\_name\}\\
\{schema\_info\}

\medskip
\#\# Conversation History\\
\{conversation\_history\}

\medskip
\#\# Current User Question\\
\{user\_utterance\}

\medskip
\#\# Budget Remaining: \{budget\_remaining\} actions

\medskip
Based on the conversation history and the current question, decide your
next action. Remember to use ACTION: and PAYLOAD: format.
\end{tcolorbox}

\medskip
% -- P3: Agentic additional system prompt --------------------------------------
\noindent\textbf{P3 — Agentic Protocol Additional System Prompt} \quad
\textit{(prepended to P1 under the agentic protocol)}
\smallskip

\begin{tcolorbox}[promptbox=black!40]
You are a text-to-Cypher assistant operating in autonomous mode. You have
a total action budget of \{total\_budget\} actions to complete the user's task.

\medskip
You do NOT have the schema in advance --- you must discover it using INSPECT\_SCHEMA.\\
You can ask the user for clarification using ASK\_USER.\\
When you have found the answer, use SUBMIT\_ANSWER to submit your final Cypher query.

\medskip
Think carefully about each action --- wasted actions reduce your ability to
complete the task.
\end{tcolorbox}

\medskip
% -- P4: Agentic turn prompt ----------------------------------------------------
\noindent\textbf{P4 — Agentic Protocol Turn Prompt} \quad
\textit{(user message, sent once per turn)}
\smallskip

\begin{tcolorbox}[promptbox=black!40]
\#\# Graph Database: \{graph\_name\}

\medskip
\#\# User's Request\\
\{user\_utterance\}

\medskip
\#\# Interaction History\\
\{interaction\_history\}

\medskip
\#\# Budget Remaining: \{budget\_remaining\} / \{total\_budget\} actions

\medskip
Decide your next action. Use ACTION: and PAYLOAD: format.
\end{tcolorbox}

\medskip
% -- P5: User simulator system prompt ------------------------------------------
\noindent\textbf{P5 — User Simulator System Prompt} \quad
\textit{(active only under the agentic protocol, responds to ASK\_USER)}
\smallskip

\begin{tcolorbox}[promptbox=black!40]
You are simulating a real user in a conversation with a graph database
assistant. You know your goal and have some domain knowledge, but you do
NOT know Cypher or the database schema.

\medskip
Rules:
\begin{enumerate}[nosep, leftmargin=1.6em]
  \item Answer questions naturally based on your persona and goal
  \item Do NOT reveal Cypher queries, schema details, or exact database values
  \item If the assistant asks about ambiguous terms, provide a reasonable
        clarification based on your intended meaning
  \item If the assistant asks irrelevant questions, express mild confusion
  \item Do NOT proactively help the assistant --- only respond to what is asked
  \item After 3+ clarification rounds, show mild impatience
\end{enumerate}
\end{tcolorbox}

\medskip
% -- P6: User simulator turn prompt --------------------------------------------
\noindent\textbf{P6 — User Simulator Turn Prompt} \quad
\textit{(user message, sent once per ASK\_USER call)}
\smallskip

\begin{tcolorbox}[promptbox=black!40]
\#\# Your Persona: \{persona\_name\}\\
\{persona\_traits\}

\medskip
\#\# Your Goal\\
You want to find out: \{goal\_description\}

\medskip
\#\# Your Knowledge
\begin{itemize}[nosep, leftmargin=1.4em]
  \item The correct interpretation for ambiguous terms: \{disambiguation\_hints\}
  \item Domain knowledge: \{domain\_knowledge\}
\end{itemize}

\medskip
\#\# Conversation So Far\\
\{conversation\_history\}

\medskip
\#\# The Assistant Just Asked\\
``\{assistant\_question\}''

\medskip
\#\# Instructions\\
Respond naturally as your persona would. Keep it concise.\\
Output ONLY your response:
\end{tcolorbox}

\end{document}